\documentclass[10pt,twocolumn,letterpaper]{article}

\newcommand{\final}{1}

\usepackage{titlesec}

\usepackage[pagebackref=true,breaklinks=true,letterpaper=true,colorlinks,bookmarks=false]{hyperref}
\usepackage{floatrow}
\usepackage{graphicx}
\usepackage{amsmath}
\usepackage{amssymb}
\usepackage{amsthm}
\usepackage{mathtools}

\usepackage{color}
\usepackage{ifthen}
\usepackage{subfig}

\usepackage{algorithm}
\usepackage{algpseudocode}
\usepackage{cleveref}
\usepackage{xcolor}
\usepackage{enumitem}

\usepackage{newtxtext}
\usepackage{newtxmath}
\usepackage{adjustbox}

\usepackage{csquotes}

\newfloatcommand{capbtabbox}{table}[][\FBwidth]

\usepackage{iccv}
\definecolor{SithColor}{rgb}{0.7,0,0} %
\newcommand{\liyi}[1]{{\color{SithColor} Li-Yi: #1 $\qed$}}
\definecolor{ConsularColor}{rgb}{0,0.4,0} %
\definecolor{GuardianColor}{rgb}{0,0,0.8} %
\definecolor{WinduColor}{rgb}{0.56,0.34,0.62} %
\definecolor{YiColor}{rgb}{0.44,0.66,0.38} %
\newcommand{\yichao}[1]{{\color{GuardianColor} Yichao: #1 $\qed$}}
\newcommand{\qisun}[1]{{\color{ConsularColor} Qi: #1 $\qed$}}
\newcommand{\hqi}[1]{{\color{WinduColor} Haozhi: #1 $\qed$}}
\newcommand{\simon}[1]{\textcolor{violet}{Simon: #1 $\qed$}}
\newcommand{\zc}[1]{\textcolor{teal}{Zhili: #1 $\qed$}}
\newcommand{\yi}[1]{{\color{YiColor} Yi: #1 $\qed$}} %
\newcommand{\N}[1]{{#1}} %

\newcommand{\warning}[1]{{\it\color{red} #1}}
\definecolor{NoteColor}{rgb}{0.8,0.5,0.2}
\newcommand{\note}[1]{{\it\color{NoteColor} #1}}
\newcommand{\nothing}[1]{}

\ifthenelse{\equal{\final}{1}}
{
\renewcommand{\liyi}[1]{}
\renewcommand{\yichao}[1]{}
\renewcommand{\qisun}[1]{}
\renewcommand{\hqi}[1]{}
\renewcommand{\simon}[1]{}
\renewcommand{\zc}[1]{}
\renewcommand{\yi}[1]{}

\renewcommand{\warning}[1]{}
\renewcommand{\note}[1]{}
}

\DeclareMathOperator*{\argmax}{argmax}
\DeclareMathOperator*{\argmin}{argmin}

\LetLtxMacro\oldcaption\caption
\renewcommand{\caption}[2][]{\oldcaption[#1]{{#1} {#2}}}

\renewcommand{\mathbf}{\boldsymbol}

\newcommand{\wireframe}{\mathsf{W}}
\newcommand{\image}{I}
\newcommand{\vertex}{\mathbf{v}}
\newcommand{\vertexA}{\mathbf{u}}
\newcommand{\vertexB}{\mathbf{w}}
\newcommand{\edge}{e}
\newcommand{\vertexSet}{{\mathsf{V}}}
\newcommand{\edgeSet}{\mathsf{E}}
\newcommand{\location}{\mathbf{p}}

\newcommand{\depth}{z}

\newcommand{\jtype}{J}
\newcommand{\jtypeT}{T}
\newcommand{\jtypeC}{C}
\newcommand{\jtypeInstance}{t}

\newcommand{\intrinsic}{K}

\newcommand{\loss}{L}

\newcommand{\x}{x}
\newcommand{\y}{y}

\newcommand{\nn}{F}
\newcommand{\featureMap}{Y}
\newcommand{\junctionMap}{J}
\newcommand{\offsetMap}{\mathbf{O}}
\newcommand{\edgeMap}{E}

\newcommand{\edgeDirMap}{\Theta}
\newcommand{\depthMap}{D}
\newcommand{\juncDepthMap}{\mathcal{D}}
\newcommand{\vanishPoint}{\mathbf{V}}
\newcommand{\assignment}{\mathsf{A}}

\newcommand{\threshold}{\vartheta}

\newcommand{\distance}{\mathrm{dist}}

\newcommand{\junctionLoss}{\loss_{\junctionMap}}
\newcommand{\offsetLoss}{\loss_{\offsetMap}}
\newcommand{\edgeLoss}{\loss_{\edgeMap}}

\newcommand{\edgeDirLoss}{\loss_{\edgeDirMap}}
\newcommand{\depthLoss}{\loss_{\depthMap}}
\newcommand{\juncDepthLoss}{\loss_{\juncDepthMap}}
\newcommand{\vanishPointLoss}{\loss_{\vanishPoint}}

\newcommand{\CE}{\mathrm{CrossEntropy}}

\titlespacing*{\subsubsection}{0pt}{3pt}{2pt}
\titlespacing*{\paragraph}{0pt}{0.3ex}{1em}
\setitemize{itemsep=2pt,topsep=0pt,parsep=0pt,partopsep=0pt}
\begin{document}

\allowdisplaybreaks

\title{Reconstructing Manhattan Worlds from a Single View} %
\title{Learning to Reconstruct Manhattan Scenes from Images} %
\title{Learning to Reconstruct 3D Wireframes from Single Images of Man-Made Scenes} %
\title{Learning to Reconstruct Manhattan Wireframes from Single Images} %
\title{Learning to Reconstruct 3D Manhattan Wireframes from a Single Image} %

\author{
Yichao Zhou${}^{1,2,}$\thanks{This work was done when Y. Zhou was an intern at Adobe Research.}
\hspace{2.5mm}
Haozhi Qi${}^1$
\hspace{1.5mm}
Yuexiang Zhai${}^{1}$
\hspace{1.5mm}
Qi Sun${}^2$
\hspace{1.5mm}
Zhili Chen${}^{2}$
\hspace{1.5mm}
Li-Yi Wei${}^2$
\hspace{1.5mm}
Yi Ma${}^{1}$
\\
\\
${}^1$UC Berkeley \qquad\qquad\qquad ${}^2$Adobe Research
}

\iccvfinalcopy
\def\iccvPaperID{1585} %
\def\httilde{\mbox{\tt\raisebox{-.5ex}{\symbol{126}}}}

\ificcvfinal\pagestyle{empty}\fi

\maketitle

\begin{abstract}

In this paper, we propose a method to obtain a compact and accurate 3D wireframe representation from a single image by effectively exploiting global structural regularities.
Our method trains a convolutional neural network to simultaneously detect salient junctions and straight lines, as well as predict their 3D depths and vanishing points.
Compared with the state-of-the-art learning-based wireframe detection methods, our network is simpler and more unified, leading to better 2D wireframe detection.
With global structural priors from parallelism, our method further reconstructs a full 3D wireframe model, a compact vector representation suitable for a variety of high-level vision tasks such as AR and CAD.
We conduct extensive evaluations on a large synthetic dataset of urban scenes as well as real images.
Our code and datasets have been made public at \url{https://github.com/zhou13/shapeunity}.

\end{abstract}

\section{Introduction}
\label{sec:introduction}

Recovering 3D geometry of a scene from RGB images is one of the most fundamental and yet challenging problems in computer vision.
Most existing off-the-shelf commercial solutions to obtain 3D geometry still requires {\em active} depth sensors such as structured lights (e.g., Apple ARKit and Microsoft Mixed Realty Toolkit) or LIDARs (popular in autonomous driving).
Although these systems can meet the needs of specific purposes, they are limited by the cost, range, and working conditions (indoor or outdoor) of the sensors.
The representations of final outputs are typically dense point clouds, which are not only memory and computation intense, but also may contain noises and errors due to transparency, occlusions, reflections, etc.

\nothing{\liyi{(March 15, 2019) Seems redundant given the mentioning of \cite{Huang:2017:TCS} below.}%
Higher level structural priors can help reconstructing cleaner geometry from such RGBD data\nothing{ for large scale indoor and outdoor environments} \cite{Kelly:2017:BLS,Huang:2017:TCS}\new{, but still require active sensing}.
}%

\begin{figure}[tb]
  \centering
  \subfloat[Input image]{
    \label{fig:teaser:input}
    \begin{minipage}[t]{0.32\linewidth}
    \frame{\includegraphics[width=\linewidth]{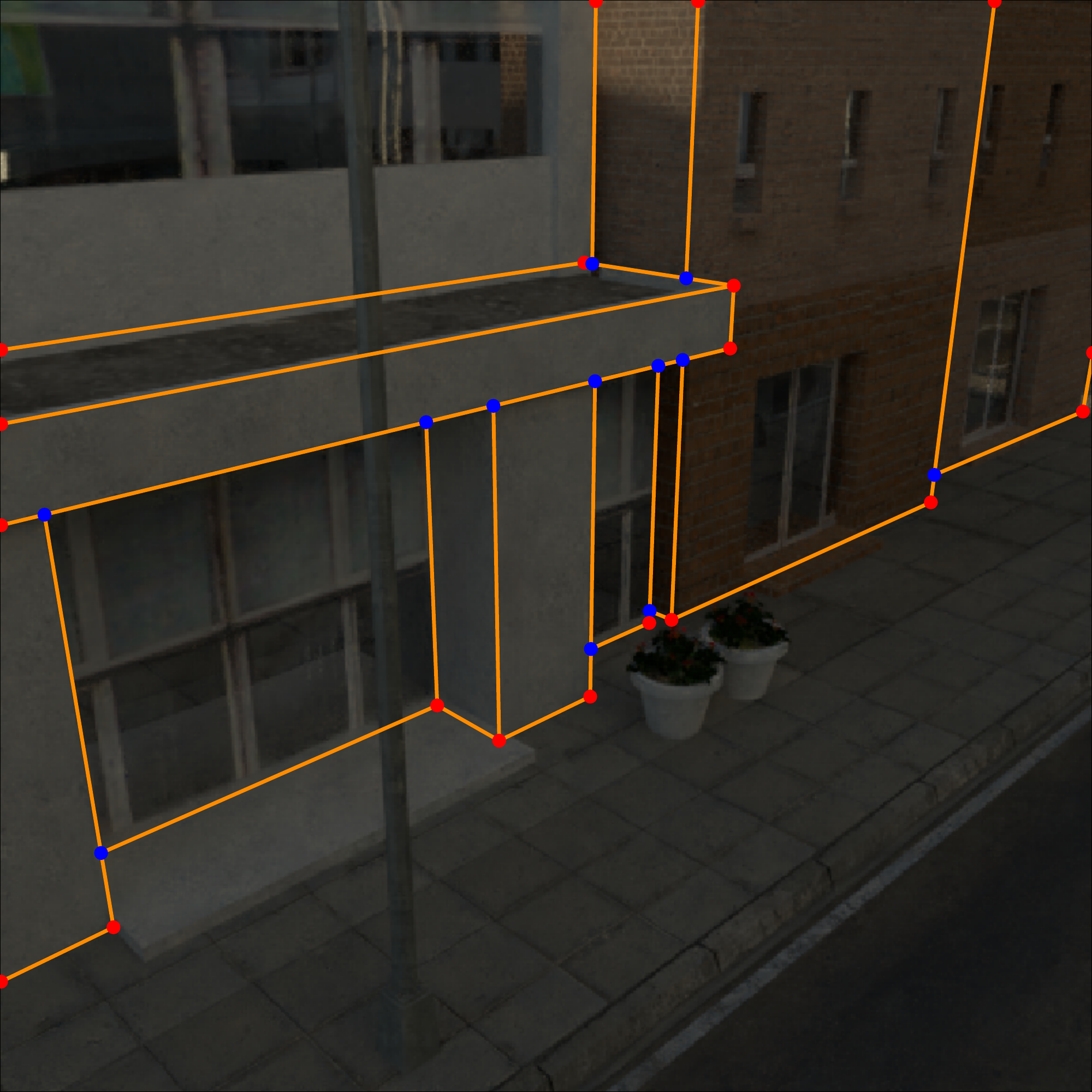}}
    
    \vspace{1ex}
    
    \frame{\includegraphics[width=\linewidth]{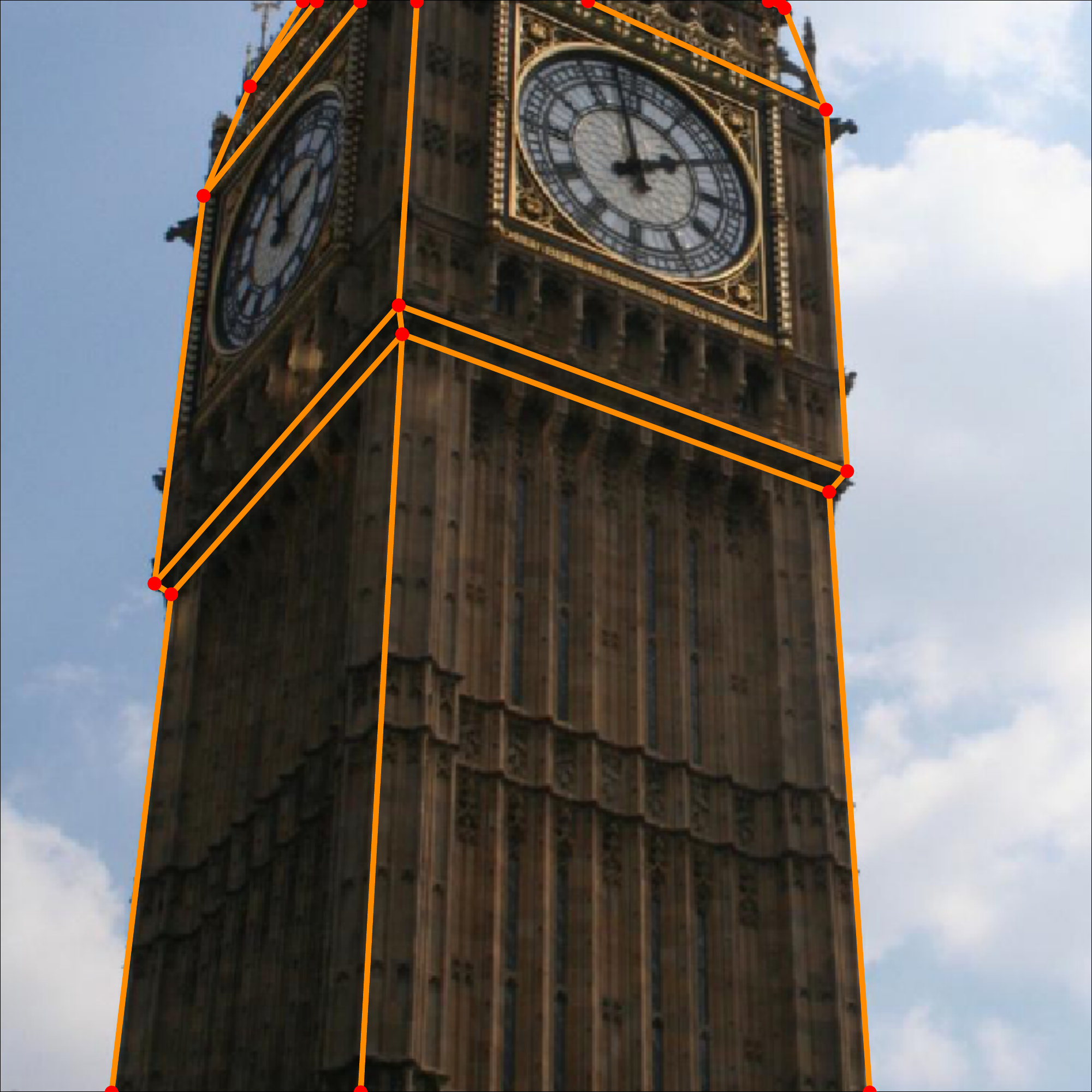}}
    \end{minipage}
  }%
  \subfloat[3D wireframe]{
    \label{fig:teaser:our}
    \begin{minipage}[t]{0.32\linewidth}
    \frame{\includegraphics[width=\linewidth]{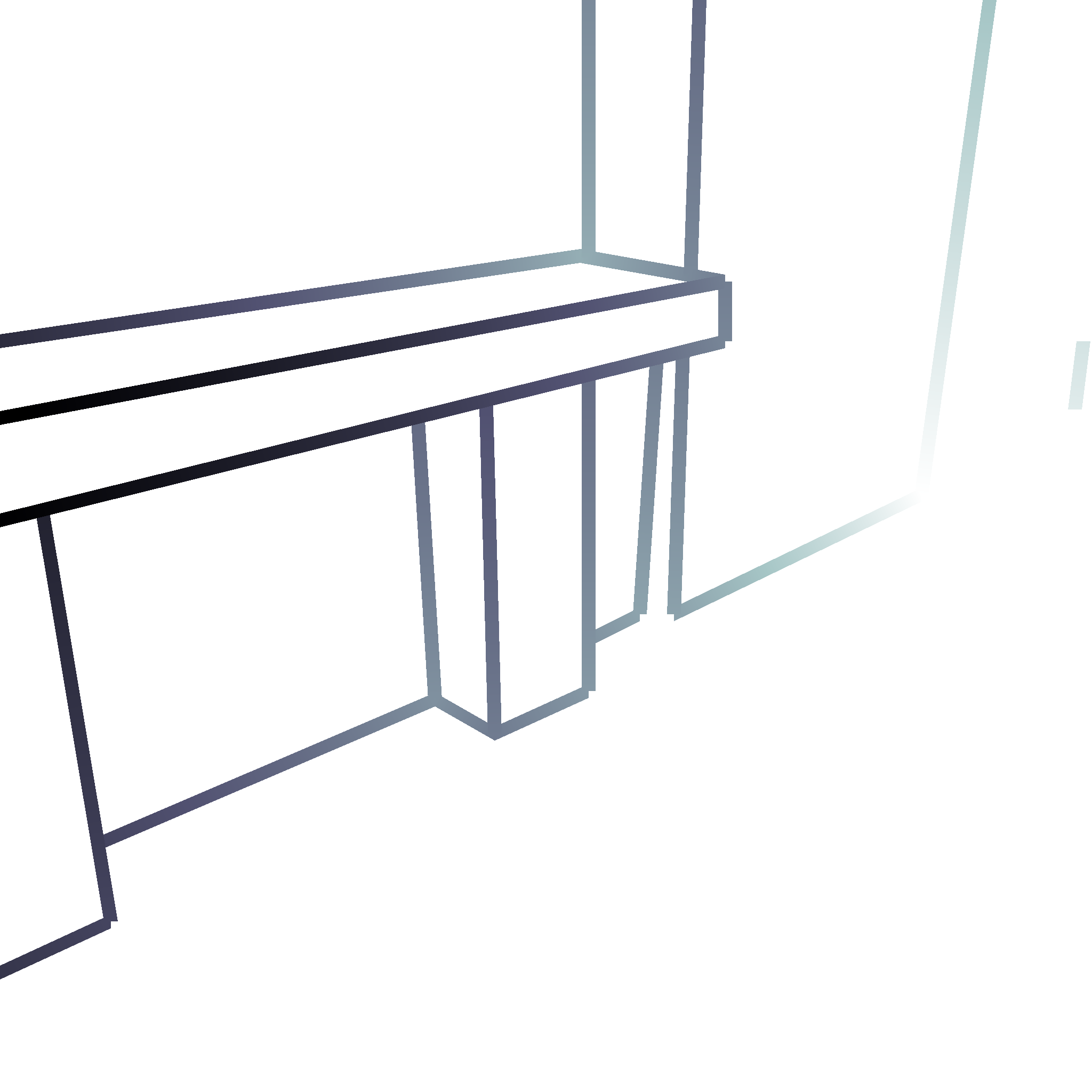}}
    
    \vspace{1ex}
    
    \frame{\includegraphics[width=\linewidth]{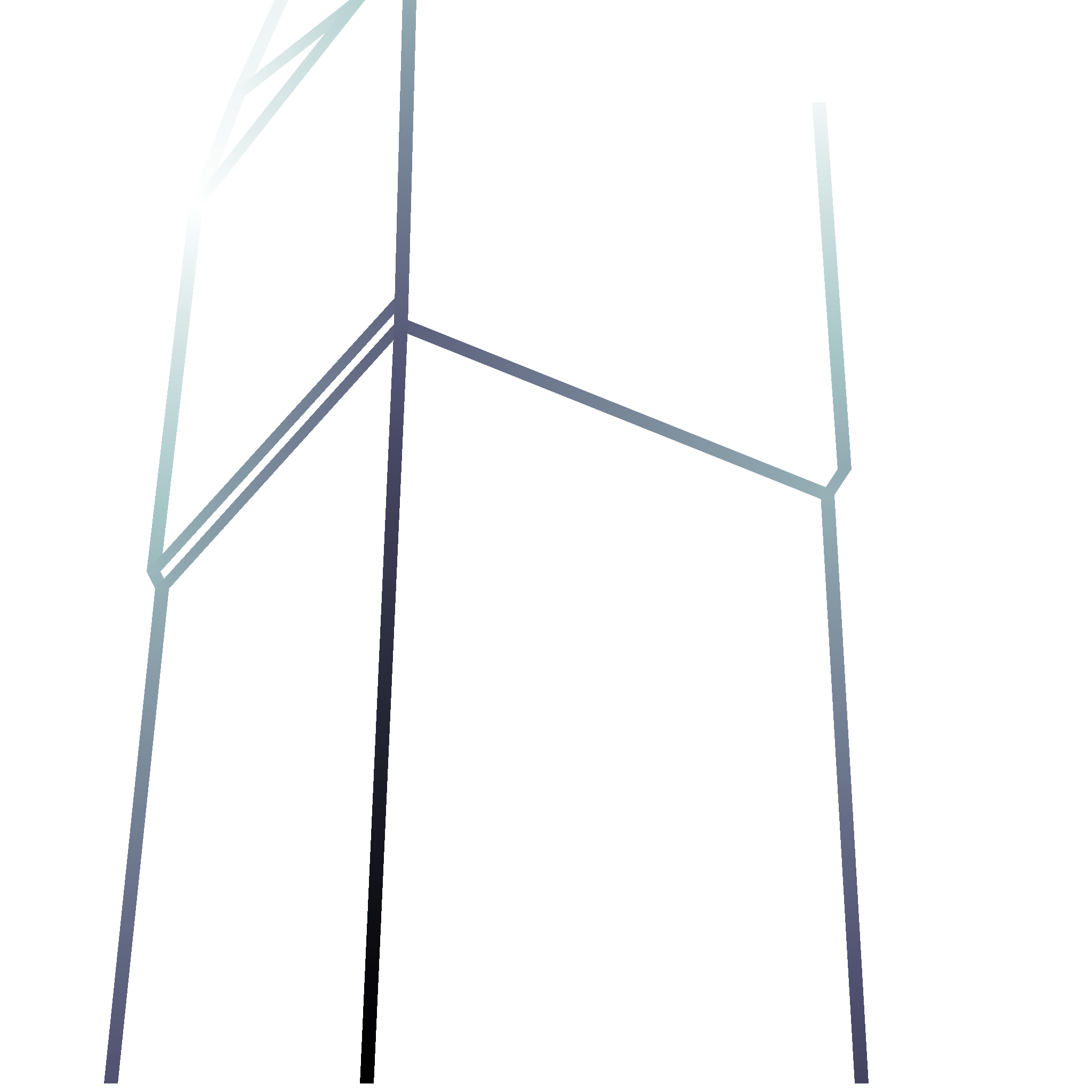}}
    \end{minipage}
  }%
  \subfloat[Novel view]{
    \label{fig:teaser:novel_view}
    \begin{minipage}[t]{0.32\linewidth}
    \frame{\includegraphics[width=\linewidth]{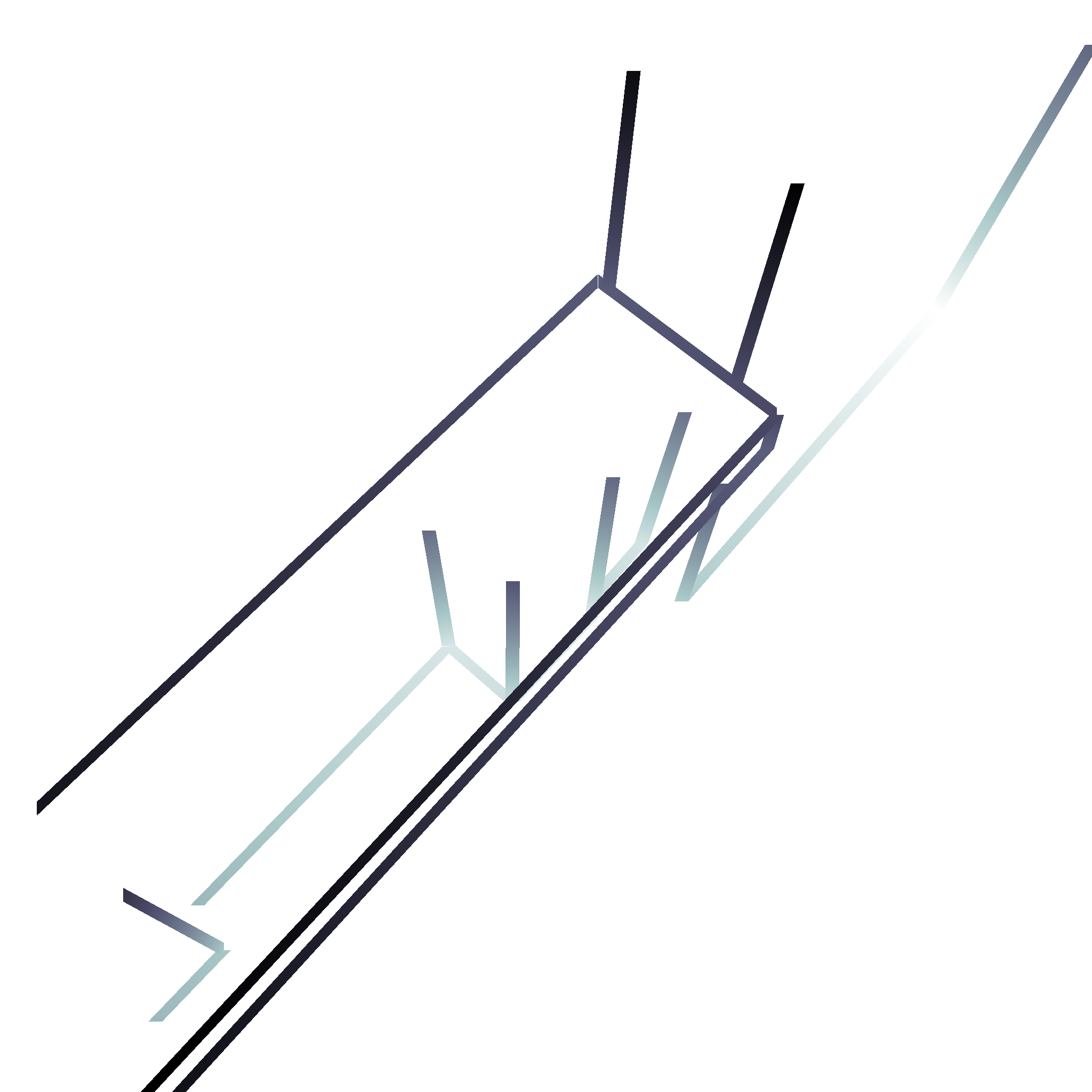}}
    
    \vspace{1ex}
    
    \frame{\includegraphics[width=\linewidth]{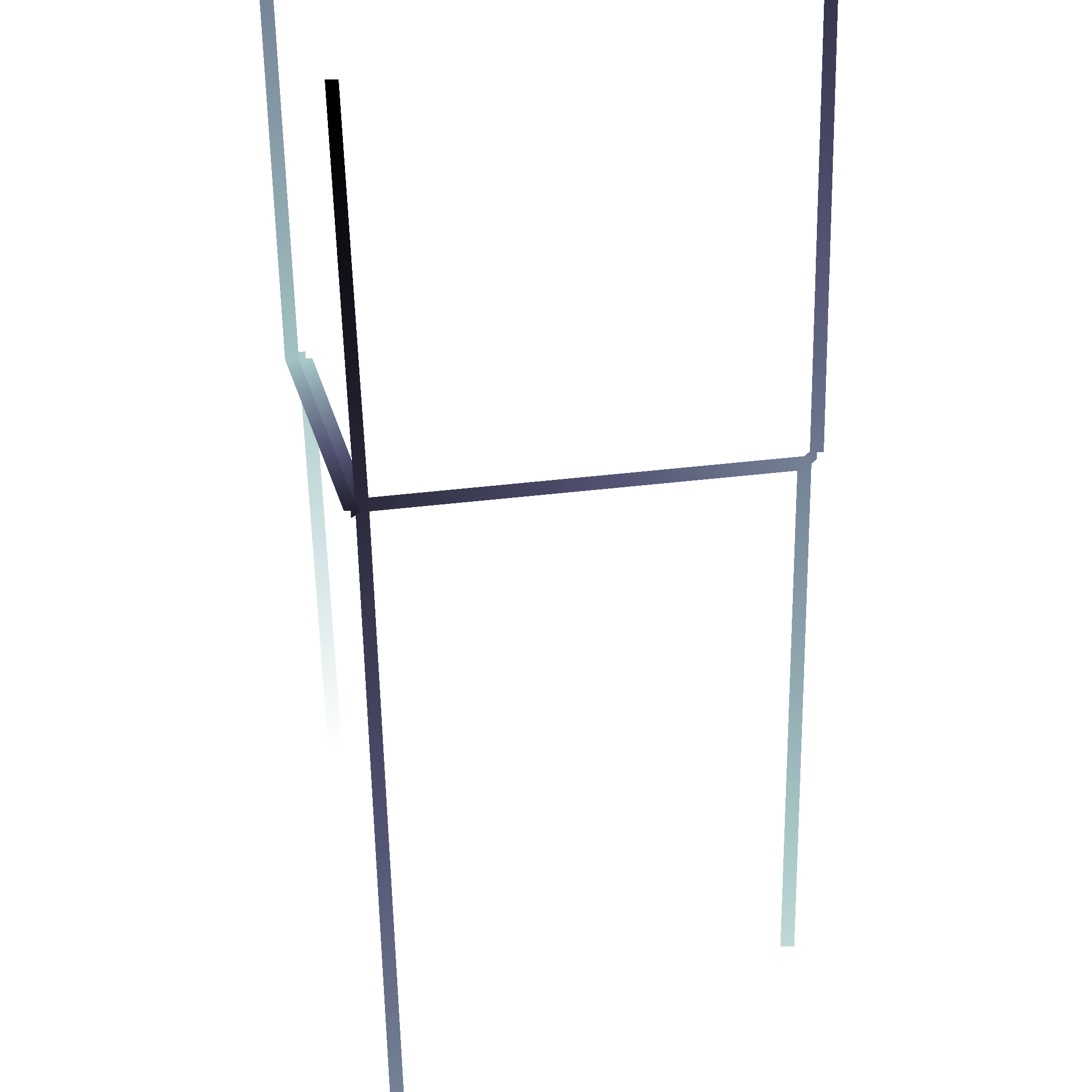}}
    \end{minipage}
  }%
  \caption[Results of our method tested on a single synthetic image (top row) and a real image (bottom row).]
  {%
  \N{Column \subref{fig:teaser:input} shows the input images overlaid with the groundtruth wireframes, in which the red and blue dots represent the C- and T-type junctions, respectively}.
Column \subref{fig:teaser:our} shows the predicted 3D wireframe from our system, with grayscale visualizing depth.
Column \subref{fig:teaser:novel_view} shows alternative views of \subref{fig:teaser:our}.
\N{Note that our system recovers geometrically salient wireframes, without being affected by the textural lines, e.g., the vertical textural patterns on the Big Ben facade.}
}%

  \label{fig:teaser}
\end{figure}

On the other hand, traditional image-based 3D reconstruction methods, such as Structure from Motion (SfM) and visual SLAM, often rely on local features.  Although the efficiency and reliability have been improving (e.g., Microsoft Hololens, Magic Leap), they often need multiple cameras with depth sensors \cite{izadi2011kinectfusion} for better accuracy.  The final scene representation remains quasi-dense point clouds, which are typically incomplete, noisy, and cumbersome to store and share.  Consequently, complex post-processing techniques such as plane-fitting \cite{Huang:2017:TCS} and mesh refinement \cite{kazhdan2006poisson,lorensen1987marching} are required. Such traditional representations can hardly meet the increasing  demand for high-level 3D modeling, content editing, and model sharing from hand-held cameras, mobile phones, and even drones.

Unlike conventional 3D geometry capturing systems, the human visual system does not perceive the world as uniformly distributed points. Instead, humans are remarkably effective, efficient, and robust in utilizing geometrically salient \emph{global} structures such as lines, contours, planes, and smooth surfaces to perceive 3D scenes \cite{Bertamini:2013:VSP}.  However, it remains challenging for vision algorithms to detect and utilize such global structures from local image features, until recent advances in deep learning which makes learning high-level features possible from labeled data.  The examples include detecting planes \cite{Yang:2018:ECCV,Liu:2018:Planenet}, surfaces \cite{Groueix:2018:AN}, 2D wireframes \cite{Huang:2018:LPW}, room layouts \cite{Zou:2018:LNR}, key points for mesh fitting \cite{zhang2016joint,wu2016single}, and sparse scene representations from multiple images \cite{Eslami:2018:NSR}.

In this work, we infer global 3D scene layouts from learned line and junction features, as opposed to local corner-like features such as SIFT \cite{Furukawa:2009:MWS}, ORB \cite{Mur-Artal-2015}, or line segments \cite{Hofer:2017:Efficient,Fleet:2014:Lsd,Ramalingam:2013:lifting} used in conventional SfM or visual SLAM systems.
Our algorithm learns to detect a special type of wireframes that consist of junctions and lines representing the corners and edges of buildings.
We call our representation the \emph{geometric wireframe} and demonstrate that together with certain global priors (such as globally or locally Manhattan \cite{Coughlan:1999:MWC,Furukawa:2009:MWS,Ramalingam:2013:lifting}), the wireframe representation allows effective and accurate recovery of the scene's 3D geometry, even from a single input image.
Our method trains a neural network to estimate global lines and two types of junctions with depths, and constructs full 3D wireframes using the estimated depths and geometric constraints.

\begin{figure*}[thb]
    \centering
    \includegraphics[width=0.99\linewidth]{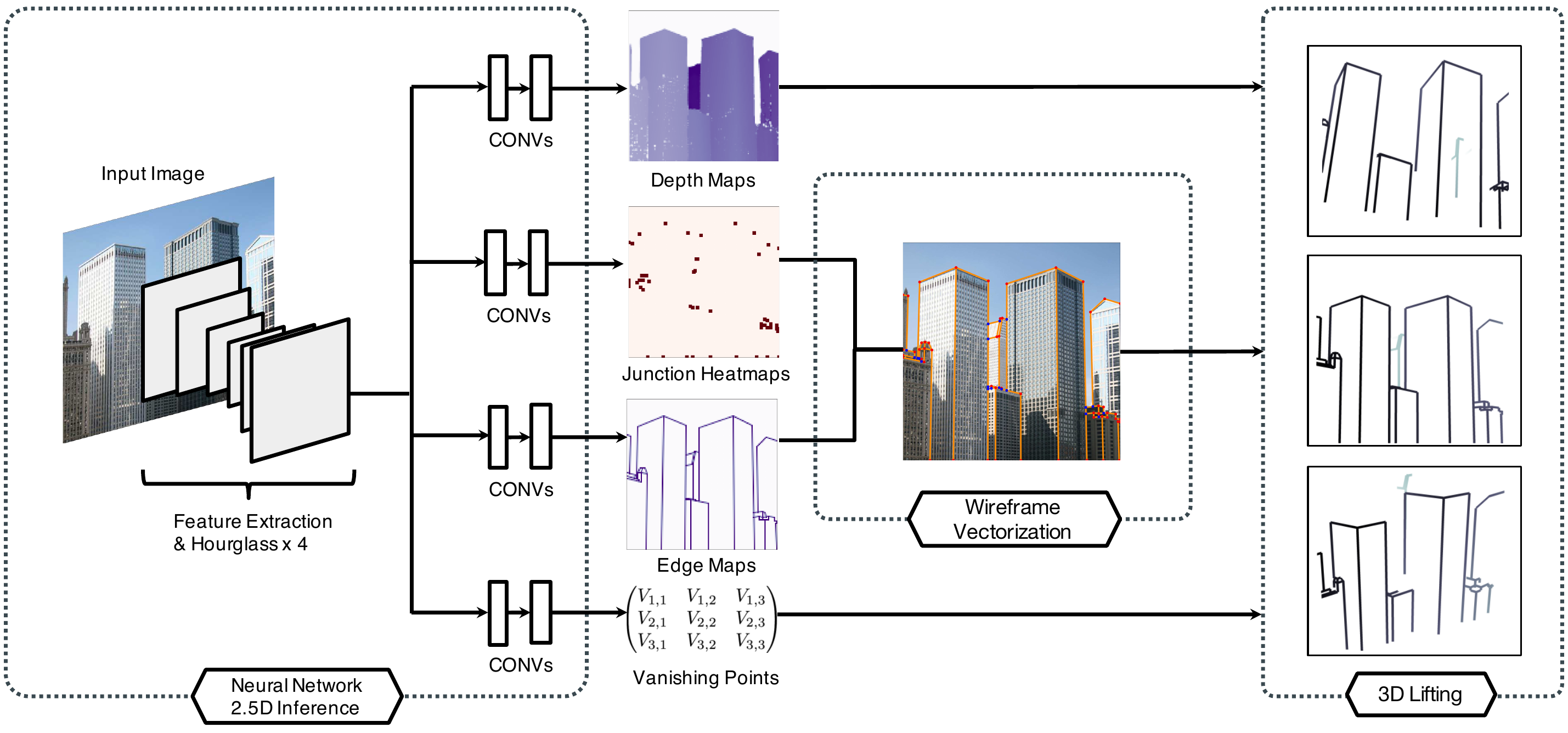}    
    \caption{Overall pipeline of the proposed method.}
    \label{fig:pipeline}
\end{figure*}

Previously, there have been efforts trying to understand the indoor scenes with the help of the 3D synthetic datasets such as the SUNCG \cite{Song:2017:Semantic,Zhang:2017:PRI}. Our work aims at natural urban environments with a variety of geometries and textures.
To this end, we build two new datasets containing both synthetic and natural urban scenes. \Cref{fig:teaser} shows the sampled results of the reconstruction and \Cref{fig:pipeline} shows the full pipeline of our system.

\paragraph{Contributions of this paper.}
Comparing to existing wireframe detection algorithms such as \cite{Huang:2018:LPW}, our method
\begin{itemize}
    \item jointly detects junctions, lines, depth, and vanishing points with a single neural network, exploiting the tight relationship among those geometric structures;
    \item learns to differentiate two types of junctions: the physical intersections of lines and planes ``C-junctions'', and the occluding ``T-junctions'';
    \item recovers a full 3D wireframe of the scene from the lines and junctions detected in a single RGB image.
\end{itemize}

\section{Methods}
\label{sec:method}

As depicted in \Cref{fig:pipeline}, our system starts with a neural network that takes a single image {as} input and jointly predicts multiple 2D heatmaps\nothing{ for potential lines and junctions}, from which we vectorize lines and junctions as well as estimate their initial depth values and vanishing points.
We call this intermediate result a 2.5D wireframe.
Using both the depth values and vanishing points estimated from the same network as the prior, we then lift the wireframe from the 2.5D \emph{image-space} into the full 3D \emph{world-space}.

\subsection{Geometric Representation}
\label{sec:representation}
In a geometric wireframe $\wireframe=(\vertexSet,\edgeSet)$ of the scene, 
$\vertexSet$ and $\edgeSet \subseteq \vertexSet \times \vertexSet$ are the junctions and lines.
Specifically, $\edgeSet$ represents lines from physical intersections of two planes while $\vertexSet$ represents (physical or projective) intersections of lines {among $\edgeSet$}.
Unlike \cite{Hofer:2017:Efficient,Huang:2018:LPW}, our $\edgeSet$ totally excludes planar textural lines, such as the vertical textures of Big Ben in \Cref{fig:teaser}. The so-defined $\wireframe$ aims to capture global scene geometry instead of local textural details.\footnote{In urban scenes, lines from regular textures (such as windows on a facade) do encode accurate scene geometry \cite{zhang2012tilt}.  The neural network can still use them for inferring the wireframe but only not to keep them in the final output, which is designed to give a compact representation of the geometry only.}
\N{By ruling out planar textural lines, we can group the junctions into two categories. Let $\jtype_{\vertex} \in \{\jtypeC,\jtypeT\}$ be the junction type of $\vertex$, in which each junction can either be a \emph{C-junction} ($\jtype_{\vertex} = \jtypeC$) or a \emph{T-junction} ($\jtype_{\vertex} = \jtypeT$).} Corner C-junctions are actual intersections of physical planes or edges, while T-junctions are generated by occlusion.  Examples of T-junctions (in blue) and C-junctions (in red) can be found in \Cref{fig:teaser}.  We denote them as two disjoint sets $\vertexSet = \vertexSet_C \cup \vertexSet_T$, in which $\vertexSet_{\jtypeC} = \{\vertex \in \vertexSet \mid \jtype_{\vertex}= \jtypeC\}$ and $\vertexSet_{\jtypeT} = \{\vertex \in \vertexSet \mid \jtype_{\vertex}= \jtypeT\}$.  \N{We note that the number of lines incident to a T-junction in $\edgeSet$ is always 1 rather than 3 because a T-junction do not connect to the two foreground vertices in 3D.}  Junction types are important for inferring 3D wireframe geometry, as different 3D priors will be applied to each type.\footnote{There is another type of junctions which are caused by lines intersecting with the image boundary. We treat them as C-junctions for simplicity.}  For each C-junction $\vertex_c \in \vertexSet_C$, define $\depth_{\vertex_c}$ as the depth of vertex $\vertex_c$, i.e., the $z$ coordinate of $\vertex_c$ in the \emph{camera space}. For each occlusional T-junction $\vertex_t \in \vertexSet_T$, we define $\depth_{\vertex_t}$ as the depth on the occluded line in the background because the foreground line depth can always be recovered from other junctions. With depth information, 3D wireframes that are made of C-junctions, T-junctions, and lines give a compact representation of the scene geometry. Reconstructing such 3D wireframes from a single image is our goal.

\subsection{From a Single Image to 2.5D Representation}

Our first step is to train a neural network that learns the desired junctions, lines, depth, and vanishing points from our labeled datasets. 
We first briefly describe the desired outputs from the network and the architecture of the network. The associated loss functions for training the network will be specified in detail in the next sections.

Given the image $I$ of a scene, the pixel-wise outputs of our neural network consist of five outputs {$-$ junction probability $\junctionMap$, junction offset $\offsetMap$, edge probability $\edgeMap$\nothing{, depth $\depthMap$}, junction depth $\juncDepthMap$, and vanishing points $\vanishPoint$}:
 \begin{align}
  \featureMap \doteq (\junctionMap, \offsetMap, \edgeMap, %
  \nothing{\depthMap,} \juncDepthMap, \vanishPoint), \;\;
  \hat \featureMap \doteq (\hat \junctionMap, \hat \offsetMap, \hat \edgeMap, %
  \nothing{\hat \depthMap,} \hat \juncDepthMap, \hat \vanishPoint),
\end{align}
where symbols with and without hats represent the ground truth and the prediction from the neural network, respectively.  The meaning of each symbol is detailed in \Cref{sec:intermediate}.

\subsubsection{Network Design}

Our network structure is based on the stacked hourglass network \cite{Newell:2016:Stacked}.
The input images are cropped and re-scaled to $512 \times 512$ before entering the network.  The feature-extracting module, the first part of the network, includes strided convolution layers and one max pooling layer to downsample the feature map to  $128 \times 128$.
The following part consists of $S$ hourglass modules. Each module will gradually downsample then upsample the feature map. The stacked hourglass network will gradually refine the output map to match the supervision from the training data. 
Let the output of the $j$th hourglass module given the $i$th image be $F_j(I_i)$.  During the training stage, the total loss to minimize is:
\begin{equation*}
  \loss^{\text{total}}  \doteq \sum_{i=1}^{N}\sum_{j=1}^{S} \loss(\featureMap_i^{(j)}, \hat \featureMap_{i})
  = \sum_{i=1}^{N}\sum_{j=1}^{S} \loss(\nn_j(\image_i), \hat \featureMap_{i}),
\end{equation*}
where
$i$ represents the index of images in the training dataset; $j$ \N{represents} the index of the
hourglass modules; $N$ represents the number of \nothing{total }training images in a batch; $S$ represents the number of stacks used in the neural network; $L(\cdot, \cdot)$ represents the loss of an individual image; $Y_i^{(j)}$ represents the predicted intermediate representation of image $I_i$ from the $j$th hourglass module, and $\hat Y_i$ represents the ground truth intermediate representation of image $I_i$.

The loss of an individual image is a superposition of the loss functions {$\loss_{k}$ specified in the next section}:
\begin{equation*}
  \loss \doteq
  \sum_k \lambda_{k}\loss_{k},\ k\in\{\junctionMap,\offsetMap,\edgeMap,\juncDepthMap,\vanishPoint\}.
\nothing{
  \lambda_{\junctionMap}\junctionLoss +
  \lambda_{\offsetMap}\offsetLoss +
  \lambda_{\edgeMap}\edgeLoss +
  \lambda_{\edgeDirMap}\edgeDirLoss +
  \lambda_{\juncDepthMap}\juncDepthLoss +
  \lambda_{\vanishPoint}\vanishPointLoss.
}%
\label{eq:loss}
\end{equation*}
The hyper-parameters $\lambda_k$ represents the weight of each sub-loss.  During experiments, we set $\lambda$ so that $\lambda_kL_k$ are of similar scales.

\subsubsection{Output Maps and Loss Functions}\label{sec:intermediate}
\paragraph{Junction Map $J$ and Loss $\junctionLoss$.}
The ground truth junction map $\hat \junctionMap$ is a down-sampled heatmap for the input image, whose value represents whether there exists a junction in that pixel. \N{For each junction type $\jtypeInstance \in \{\jtypeC, \jtypeT\}$, we estimate its junction heatmap}
\begin{equation*}
  \hat\junctionMap_\jtypeInstance(\location) = 
  \begin{cases}
 1  & \exists \vertex \in \vertexSet_\jtypeInstance: \location = \lfloor \frac{\vertex}{4} \rfloor \\
 0 & \text{otherwise} \\
  \end{cases},\; \jtypeInstance \in \{\jtypeC, \jtypeT\}.
\end{equation*}
where $\location$ is the integer coordinate on the heatmap and $\vertex$ is the coordinate of a junction with type $\jtypeInstance$ in the image space.  Following \cite{Newell:2016:Stacked}, the resolution of the junction heatmap is $4$ times less than the resolution of the input image.%

Because some pixels may contain two types of junctions, we treat the junction prediction as two per-pixel binary classification problems. We use the classic softmax cross entropy loss to predict the junction maps:
\begin{equation*}
  \junctionLoss(\junctionMap, \hat \junctionMap) \doteq \frac{1}{n}
  \sum_{\mathclap{\jtypeInstance\in \{C, T\}\;\;}}
  \sum_{\mathclap{\location}}
  \CE\left(
  \junctionMap_\jtypeInstance(\location), \hat \junctionMap_\jtypeInstance(\location)
  \right),
\end{equation*}
where $n$ is the number of pixels of the heatmap.
The resulting $\junctionMap_\jtypeInstance(\x,\y) \in (0, 1)$ represents the probability whether there exists a junction with type $\jtypeInstance$ at $[4\x, 4\x+4) \times [4\y, 4\y+4)$ in the input image.

\paragraph{Offset Map $\offsetMap$ and Loss $\offsetLoss$.}
Comparing to the input image, the lower resolution of $\junctionMap$ might affect the precision of junction positions. Inspired by \cite{wang2017point}, we use an offset map to store
the difference vector from \N{$\hat \junctionMap$} to its original position \N{with sub-pixel accuracy}:
\begin{equation*}
  \hat\offsetMap_\jtypeInstance(\location) = \begin{cases}
  \frac{\vertex}{4} - \location & \exists \vertex \in \vertexSet_\jtypeInstance:
  \location = \lfloor \frac{\vertex}{4} \rfloor \\
  0 & \text{otherwise} \\
  \end{cases},\; \jtypeInstance \in \{\jtypeC, \jtypeT\}.
\end{equation*}
\nothing{  In some rare cases, it is possible that there are multiple junctions $\vertex$ inside a single pixel of the heatmap. If that happens, we break the tie arbitrarily. }

We use the $\ell_2$-loss for the offset map and use the heatmap as a mask to compute the loss only near the actual junctions. Mathematically, the loss function is written as
\begin{equation*}
  \offsetLoss(\offsetMap, \hat \offsetMap) \doteq
  \sum_{\mathclap{\jtypeInstance\in \{C, T\}\;\;}}
  \frac{
  \sum_{\location}
  \hat \junctionMap_\jtypeInstance(\location) \left|\left|
  \offsetMap_\jtypeInstance(\location) - \hat \offsetMap_\jtypeInstance(\location)
  \right|\right|_2^2
  }{\sum_{\location}\hat \junctionMap_\jtypeInstance(\location)},
\end{equation*}
where $\offsetMap_\jtypeInstance(\location)$ is computed by applying a sigmoid and constant translation function to the last layer
of the offset branch in the neural network to enforce $\offsetMap_\jtypeInstance(\location) \in [0, 1)^2$.
We normalize $\offsetLoss$ by the number of junctions of each type.

\paragraph{Edge Map $\edgeMap$ and Loss $\edgeLoss$.}
To estimate line positions, we represent them in an edge heatmap.
For the ground truth lines, we draw them on the edge map using an anti-aliasing technique \cite{Zingl:2012:Rasterizing} for better accuracy. Let $\distance(\location, \edge)$ be the shortest distance between a pixel $\location$ and the nearest line segment $\edge$. We define the edge map to be
\begin{equation*}
  \hat \edgeMap(\location) = \begin{cases}
    \max_\edge 1-\distance(\location,\edge) & \exists \edge \in \edgeSet: \distance(\location,\edge) < 1, \\
    0 & \text{otherwise}.
  \end{cases}
\end{equation*}
Intuitively, $\edgeMap(\location) \in [0, 1]$ represents \N{the probability of a line close to point $\location$}.  Because the range of the edge map is always between $0$ and $1$, we can treat it as a
probability distribution and use the sigmoid cross entropy loss on the $\edgeMap$ and $\hat \edgeMap$\N{:}
\begin{equation*}
  \edgeLoss(\edgeMap, \hat \edgeMap) \doteq \frac 1 n
  \sum_{\location} 
  \CE\left(
  \edgeMap(\location), \hat \edgeMap(\location)
  \right).
\end{equation*}

\paragraph{Junction Depth Maps $\juncDepthMap$ and Loss $\juncDepthLoss$.}
To estimate the depth $\depth_{\vertex}$ for each junction $\vertex$, we define the junction-wise depth map as
\begin{equation*}
  \hat\juncDepthMap_\jtypeInstance(\location) = 
  \begin{cases}
    \depth_{\vertex}  & \exists \vertex \in \vertexSet_t: \location = \lfloor \frac{\vertex}{4} \rfloor \\
    0 & \text{otherwise} \\
  \end{cases},\; \jtypeInstance \in \{\jtypeC, \jtypeT\}.
\end{equation*}

In many datasets with unknown depth units and camera intrinsic matrix $\intrinsic$, $\depth_{\vertex}$ remains a relative scale instead of absolute depth.
To remove the ambiguity from global scaling, we use scale-invariant loss (SILog) which has been introduced in the single image depth estimation literature \cite{Eigen:2014:Depth}.
It removes the influence of the global scale by summing the log difference between each pixel pair.
\begin{align*}
  \depthLoss(\juncDepthMap, \hat\juncDepthMap) &\doteq
  \sum_{\jtypeInstance}
  \frac{1}{n_\jtypeInstance}
  \sum_{\mathclap{\location \in \vertexSet_t}}
  \big(\log \juncDepthMap_\jtypeInstance(\location)-\log \hat\juncDepthMap_\jtypeInstance(\location)\big)^2 \nonumber \\ &-
  \sum_{\jtypeInstance}\frac{1}{n_\jtypeInstance^2}
  \big(\sum_{\mathclap{\location \in \vertexSet_t}}
  \log \juncDepthMap_\jtypeInstance(\location) - \log\hat\juncDepthMap_\jtypeInstance(\location)\big)^2. \label{eq:d1}
\end{align*}
\nothing{
We use SILog whenever the scale of the depth map or the camera intrinsic matrix $\intrinsic$ is unknown for the dataset.  Otherwise, we can use the $\ell_2$ loss to \N{preserve} the scale \N{of the depth}:
}
\nothing{
\N{For other datasets containing known depth scale or $\intrinsic$, we use the $\ell_2$ loss to \N{preserve} the depth scale}
\begin{equation*}
  \depthLoss'(\juncDepthMap, \hat\juncDepthMap)\doteq\frac 1 n
  \sum_{\jtypeInstance}
  \sum_{\mathclap{\location\in\vertexSet_t}}
  \left\|\juncDepthMap_\jtypeInstance(\location)-\hat\juncDepthMap_\jtypeInstance(\location)\right\|_2^2. 
\label{eq:d2}
\end{equation*}
}

\paragraph{Vanishing Point Map $\vanishPoint$ and Loss $\vanishPointLoss$.}
Lines in man-made outdoor scenes often \N{cluster around} the three mutually orthogonal directions.  Let $i \in \{1,2,3\}$ represent
these three directions. In perspective geometry, parallel lines in direction $i$ will intersect at the same \N{vanishing} point $(V_{i,x}, V_{i,y})$ in the image space, possibly at infinity. To prevent $V_{i,x}$ or $V_{i,y}$ from becoming too large, we normalize the vector so that
\begin{equation}
    \vanishPoint_i = \frac{1}{V_{i,x}^2 + V_{i,y}^2 + 1} 
    \begin{bmatrix}
    V_{i,x}, V_{i,y}, 1
    \end{bmatrix}^T.
\end{equation}
Because the two horizontal vanishing points $\vanishPoint_1$ and $\vanishPoint_2$ are order agnostic from a single
RGB image\N{, we use} the Chamfer $\ell_2$-loss for $\vanishPoint_1$ and $\vanishPoint_2$, and the
$\ell_2$-loss for $\vanishPoint_3$ (the vertical vanishing point):
\begin{multline*}
    \vanishPointLoss(\vanishPoint, \hat\vanishPoint) \doteq 
    \min(\|\vanishPoint_1 - \hat\vanishPoint_1\|, \|\vanishPoint_2 - \hat\vanishPoint_1\|) \\+
    \min(\|\vanishPoint_1 - \hat\vanishPoint_2\|, \|\vanishPoint_2 - \hat\vanishPoint_2\|) +
    \|\vanishPoint_3 - \hat\vanishPoint_3\|_2^2.
\end{multline*}

\subsection{Heatmap Vectorization} \label{sec:vec}
As seen from \Cref{fig:pipeline}, the outputs of the neural network are essentially image-space 2.5D heatmaps of the desired wireframe. Vecterization is needed to obtain a compact wireframe representation.

\paragraph{Junction Vectorization.}
Recovering the junctions $\vertexSet$ from the junction heatmaps $\junctionMap$ is straightforward.
Let $\threshold_C$ and $\threshold_T$ be the thresholds for $\junctionMap_C$ and $\junctionMap_T$.  The junction candidate sets can be estimated as
\begin{equation}
\vertexSet_t \gets \{\location + \offsetMap_t(\location) \mid \junctionMap_t(\location) \ge \threshold_t \},\ t\in\{C,T\}. \\
\end{equation}

\nothing{
To detect the junctions from the heat maps, we use the classical non-maximal suppression algorithm.
The algorithm is shown in Algorithm \ref{alg:junction-detection}.  Here, $\mathcal{N}_4(x, y)$
represents the 4-near pixels around $(x, y)$. In addition, we introduce a hyper parameter $\gamma$
to control the softness of the non-maximal suppression.  This is because when the resolution of the
heat map is not that high, there are often two junctions occupying the two nearby pixels.

\begin{algorithm}
  \begin{algorithmic}[1]
    \caption{Junction Vectorization Algorithm} \label{alg:junction-detection}
    \Require Junction map $J$ and hyper-parameter $\gamma$.
    \Ensure C-junction set $J_C$, T-junction set $J_T$.
    \State $J_C \gets \Call{ComputeJunction}{J(0, \cdot)}$
    \State $J_T \gets \Call{ComputeJunction}{J(1, \cdot)}$
    \Procedure{ComputeJunction}{$J$}
      \ForAll{$(x,y) \in \textsc{ArgSort}(J)$}
        \ForAll{$(x', y') \in \mathcal{N}_4(x, y)$}
          \If{$\gamma J(x,y) > J(x', y')$}
            \State $J(x', y') \gets 0$
          \EndIf
        \EndFor
      \EndFor
      \State \Return $\{(x, y)\mid H(x, y) > 0\}$.
    \EndProcedure
  \end{algorithmic}
\end{algorithm}
}%

\paragraph{Line Vectorization.}
Line vectorization has two stages. In the first stage, we detect and construct the line
candidates from all the corner C-junctions. This can be done by enumerating all the pairs of junctions $\vertexA, \vertexB \in \vertexSet_C$, connecting them, and testing if their line confidence score is greater than a threshold $c(\vertexA, \vertexB) \ge \threshold_E$. The confidence score of a line with two endpoints $\vertexA$ and $\vertexB$ is given as
$c(\vertexA, \vertexB) =
  \frac{1}{|\vertexA\vertexB|}
  \sum_{{\location \in {P}(\vertexA, \vertexB)}}
  \edgeMap(\location)$
where ${P}(\vertexA,\vertexB)$ represents the set of pixels in the rasterized line $\vec{\vertexA\vertexB}$, and $|\vec{\vertexA\vertexB}|$ represents the number of pixels in that line. %

In the second stage, we \N{construct} all \N{the} lines
\N{between ``T-T'' and ``T-C'' junction pairs.}
We repeatedly add a T-junction to the wireframe if it is tested to be close to a detected line. Unlike corner C-junctions, the degree of a T-junction is always one. So for each T-junction, we find the best edge associated with it. This process is repeated until no more lines could be \N{added}.
Finally, we run a post-processing procedure to remove lines that are too close or cross each other. By handling C-junctions and T-junctions separately, our line vectorization algorithm is both efficient and robust for scenes with hundreds of lines. A more detailed description is discussed in the supplementary material.

\subsection{\N{Image-Space} 2.5D to \N{World-Space} 3D}
\label{sec:lift}
So far, we have obtained vectorized junctions and lines in 2.5D image space with depth in a relative scale.  However, in scenarios such as AR and 3D design, absolute depth values are necessary for 6DoF manipulation of the 3D wireframe. In this section, we present the steps to estimate them with our network predicted vanishing points.

\subsubsection{Calibration from Vanishing Points}
In datasets such as MegaDepth \cite{Zhengqi:2018:MegaDepth}, the camera calibration matrix
$K \in \mathbb{R}^{3\times3}$ of each image is unknown, although it is critical for a full 3D wireframe reconstruction.  Fortunately, calibration matrices can be inferred from three mutually orthogonal vanishing points if the scenes are mostly Manhattan.
According to \cite{Ma:2003:IVI}, if we transform the orthogonal vanishing points $\vanishPoint_i$ to the calibrated coordinates
$\bar{\vanishPoint}_i \doteq K^{-1}\vanishPoint_i$,
then $\bar{\vanishPoint}_i$ should be mutually orthogonal, i.e.,
\begin{equation*}
\vanishPoint_{i}K^{-T}K^{-1}
\vanishPoint_{j} =0,\quad \forall i,j\in\{1,2,3\},\  i\neq j. 
\end{equation*}
These equations impose three linearly independent constraints on $K^{-T}K^{-1}$ and would enable solving up to three unknown parameters in the calibration matrix, such as the optical center and the focal length.

\subsubsection{Depth Refinement with Vanishing Points}
\label{sec:lifting}

Due to the estimation error, the predicted depth map may not be consistent with the detected vanishing points $\vanishPoint_i$.  In practice, we find the neural network performs better on estimating the vanishing points than predicting the 2.5D depth map. This is probably because there are more geometric cues for the vanishing points, while estimating depth requires priors from data.  Furthermore, the unit of the depth map might be unknown due to the dataset (e.g., MegaDepth) and the usage of SILog loss. Therefore, we use the vanishing points to refine the junction depth and determine its absolute value.  Let $\tilde \depth_{\vertex} \doteq \juncDepthMap_{\jtype_{\vertex}}(\vertex)$ be the predicted depth for junction $\vertex$ from our neural network.  We design the following convex objective function:
\vspace{-0.5mm}
\begin{align}
  \min_{z,\alpha} \quad& \sum_{i=1}^{3} \sum_{(\vertexA, \vertex) \in \assignment_i}
  \left\|\left(\depth_{\vertexA} \bar\vertexA - \depth_{\vertex} \bar\vertex
  \right) \times \bar\vanishPoint_i\right\|_2 \notag \\
  &\qquad \quad + \lambda_R \sum_{\vertex \in \vertexSet} 
  \left\|\depth_{\vertex} - \alpha\tilde\depth_{\vertex}\right\|_2^2 \label{eq:lift-objective} \\
  \text{subject to} \quad &  \depth_{\vertex} \ge 1, \quad \forall \vertex \in \vertexSet, \label{eq:lift-scale-constraint}  \\
  & \lambda{\depth_{\vertexA}} + (1-\lambda){\depth_{\vertex}} \le {\depth_{\vertexB}},
  \:   \label{eq:lift-depth-constraint} \\ &
  \forall \vertexB \in \vertexSet_T, (\vertexA,\vertex) \in \edgeSet: \vertexB=\lambda\vertexA+(1-\lambda)\vertex,  \notag
\end{align}\normalsize
where $\assignment_i$ represents the set of lines corresponding to vanishing point $i$; $\alpha$ resolves the scale ambiguity in the depth dimension; $\bar\vertexA \doteq \intrinsic^{-1} [u_x \: u_y \: 1]^T$ is the vertex position in the calibrated coordinate.
The goal of the first term in \Cref{eq:lift-objective}
is to encourage the line $(\depth_{\vertexA} \bar\vertexA,\depth_{\vertexB} \bar\vertexB)$
parallel to vanishing point $\bar\vanishPoint_i$ by penalizing over the parallelogram area spanned by those two vectors.  The second term regularizes $\depth_{\vertex}$ so that it is close to the network's estimation $\tilde \depth_{\vertex}$ up to a scale.
\Cref{eq:lift-scale-constraint} prevents the degenerating solution $\mathbf{\depth} = \mathbf{0}$. \Cref{eq:lift-depth-constraint} is a convex relaxation of $\frac{\lambda}{\depth_{\vertexA}} + \frac{1-\lambda}{\depth_{\vertexB}} \ge \frac{1}{\depth_{\vertex}}$, the depth constraint for T-junctions.

\section{Datasets and Annotation}\label{sec:data}
One of the bottlenecks of supervised learning is inadequate dataset for training and testing. Previously, \cite{Huang:2018:LPW} develops a dataset for 2D wireframe detection.
However, their dataset does not contain the 3D depth or the type of junctions.
To the best of our knowledge, there is no public image dataset that has both wireframe and 3D information.
To validate our approach, we create a hybrid dataset with a larger number of synthetic images of city scenes and smaller number of real images. 
The former has accurate 3D geometry and automatically annotated ground truth 3D wireframes from mesh edges, while the latter is manually labelled with less accurate 3D information.

\paragraph{SceneCity Urban 3D Dataset (SU3).}
To obtain a large \N{number of images} with accurate geometrical wireframes, we use a progressively\liyi{what does this mean?} generated 3D mesh repository, SceneCity\footnote{https://www.cgchan.com/}.
The dataset \N{is made up of} simple polygons \N{with} artist-tuned materials and textures.
We extract the C-junctions from the \N{vertices} of the mesh and compute T-junctions using computational geometry algorithms \N{and OpenGL.} Our dataset \N{includes} 230 \N{cities}, \N{each containing $8 \times 8$ city blocks}.  The cities have different building arrangements and lighting conditions by varying the sky maps.
We randomly generate 100 viewpoints for each city based on criteria such as the number of captured buildings to simulate hand-held and drone cameras.
The synthetic outdoor images are then rendered through global illumination by Blender\nothing{ modeler}, which provides $23,000$ images in total.
We use the images of the first 227 cities for training and the rest 3 cities for validation.

\paragraph{Realistic Landmark Dataset.}
The MegaDepth v1 dataset \cite{Li:2018:MDL} contains real images \N{of} 196 landmarks in the world. It also contains the depth maps of these images via structure from motion.
We select about 200 images that approximately meet the assumptions of our method, manually label their wireframes, and \N{register} them with the rough 3D depth.

In our experiments, we pretrain our network on the SU3 dataset, and then use 2/3 of the real images to finetune the model.
The remaining 1/3 is for testing.

\section{Experiments}
\label{sec:result}

We conduct extensive experiments to evaluate our method and validate the design of our pipeline with ablation studies.  In addition, we compare our method with the state-of-the-art 2D wireframe extraction approaches. We then evaluate the performance of our vanishing point estimation and depth refinement steps.  Finally, we demonstrate the examples of our 3D wireframe reconstruction.

\subsection{Implementation Details} \label{sec:detail}
Our backbone is a two-stack hourglass network \cite{Newell:2016:Stacked}. Each stack consists of 6 stride-2 residual blocks and 6 nearest neighbour upsamplers. After the stacked hourglass feature extractor, we insert different ``head'' modules for each map.  Each head contains a $3\times3$ convolutional layer to reduce the number of channels followed by a $1\times1$ convolutional layer to compute the corresponding map. For vanishing point regression, we use a different head with two consecutive stride-2 convolution layers followed by a global average pooling layer and a fully-connected layer to regress the position of the vanishing points.

During the training, the ADAM \cite{Kingma:2014:Adam} optimizer is used. The learning rate and weight decay are set to $8\times 10^{-4}$ and $1\times 10^{-5}$. All the experiments are conducted on four NVIDIA GTX 1080Ti GPUs, with each GPU holding $12$ mini-batches. For the SceneCity Urban 3D dataset, we train our network for 25 epochs. The loss weights are set as $\lambda_{\junctionMap} = 2.0$,  $\lambda_{\offsetMap}=0.25$ $\lambda_{\edgeMap} = 3.0$, and $\lambda_{\juncDepthMap}=0.1$ so that all the loss terms are roughly equal. For the real-world dataset, we initialize the network with the one trained on the SU3 dataset and use a $10^{-4}$ learning rate to train for 5 epochs. We horizontally flip the input image as data-augmentation.  Unless otherwise stated, the input images are cropped to $512 \times 512$. The final output is of stride $4$, i.e., with size $128 \times 128$.
During heatmap vectorization, we use the hyper-parameter $\threshold_C=0.2$, $\threshold_T=0.3$, and $\threshold_E=0.65$.

\subsection{Evaluation Metrics}
We use the standard AP (average precision) from object detection \cite{everingham2010pascal} to evaluate our junction prediction results. Our algorithm produces a set of junctions and their associated scores. The prediction is considered correct if its $\ell^2$ distance to the nearest ground truth is within a threshold.  By this criterion, we can draw the precision-recall curve and compute the \emph{mean AP} (mAP) as the area under this curve averaging over several different thresholds of junction distance.

In our implementation, mAP is averaged over thresholds 0.5, 1.0, and 2.0.
In practical applications, long edges between junctions are typically preferred over short ones. Therefore, we weight the mAP metric by the sum of the length of the lines connected to that junction. We use $\text{AP}^{C}$ and $\text{AP}^{T}$ to represent such weighted mAP metric for C-junctions and T-junctions, respectively.   We use the intersection over union (IoU) metric to evaluate the quality of line heatmaps.  For junction depth map, we evaluate it on the positions of the ground truth junctions with the scale invariant logarithmic error (SILog) \cite{Eigen:2014:Depth, geiger2013vision}.

\subsection{Ablation on Joint Training and Loss Functions}
We run a series of experiments to investigate how different feature designs and multi-task learning strategies affect the wireframe detection accuracy. \Cref{tab:multi-task} presents our ablation studies with different combinations of tasks to research the effects of joint training.  We also evaluate the choice of $\ell_1$- and $\ell_2$-losses for offset regression and the ordinary loss \cite{Zhengqi:2018:MegaDepth} for depth estimation.  We conclude that:
\begin{enumerate}[nosep]
    \item Regressing offset is significantly important for localizing junctions (7.4 points for $\text{AP}^{C}   $ and 3 points for $\text{AP}^{T}$), by comparing rows (a-c). In addition, $\ell_2$ loss is better than $\ell_1$ loss, probably due to its smoothness.
    \item Joint training junctions and lines improve in both tasks. Rows (c-e) show improvements with about 1.5 points in $\text{AP}^{C}$, and 0.9 point in $\text{AP}^{T}$ and line IoU. This indicates the tight relation between junctions and lines.
    \item For depth estimation, we test the ordinal loss from \cite{Zhengqi:2018:MegaDepth}. To our surprise, it does not improve the performance on our dataset (rows (f-g)). We hypothesis that this is because the relative orders of sparsely annotated junctions are harder to predict than the foreground/background relationship in \cite{Zhengqi:2018:MegaDepth}.
    \item According to rows (f) and (h), joint training with junctions and lines slightly improves the performance of depth estimation by 0.55 SILOG point.
\end{enumerate}
\begin{small}
\begin{table}
\setlength{\tabcolsep}{3pt}
\renewcommand{\arraystretch}{1.2}
\begin{center}
\begin{adjustbox}{width=\linewidth,center}
\begin{small}
\begin{tabular}{c|cccccc|c|c|c|c}
\hline
& \multicolumn{6}{c|}{supervisions} & \multicolumn{4}{c}{metrics}\\
\hline
& \multicolumn{1}{c|}{$\junctionMap$} & \multicolumn{2}{c|}{$\offsetMap$} & \multicolumn{1}{c|}{$\edgeMap$} & \multicolumn{2}{c|}{$\juncDepthMap$} & \multicolumn{2}{c|}{$\junctionMap$} & \multicolumn{1}{c|}{$\edgeMap$} &
\multicolumn{1}{c}{$\juncDepthMap$} \\
 & \multicolumn{1}{c|}{\small{CE}} & \small{$\ell_1$} & \small{$\ell_2$} & \multicolumn{1}{|c|}{\small{CE}} & \small{SILog\hspace{-3pt}} & \small{Ord} & \multicolumn{1}{c}{\small{$\text{AP}^{C}$}} & \multicolumn{1}{c|}{\small{$\text{AP}^{T}$}} & \multicolumn{1}{c|}{\small{$\text{IoU}_{\edgeMap}$}} & \multicolumn{1}{c}{\small{\hspace{-1pt}SILog\hspace{-1pt}}} \\ 
\hline
\hline
(a) & \checkmark & & & & & & 65.4 & 57.1 & / & /\\
(b) & \checkmark & \checkmark & & & & & 69.3 & 55.8 & / & / \\ 
(c) & \checkmark & & \checkmark & & & & 72.8 & 60.1 & / & / \\
\hline
(d) & & & & \checkmark & & & / & / & 73.3 & / \\
\hline
(e) & \checkmark & & \checkmark & \checkmark & & & 74.3 & 61.0 & 74.2 & / \\
\hline
(f) & & & & &\checkmark  & & / & / & / & 3.59 \\
(g) & & & & & \checkmark & \checkmark & / & / & / & 4.14 \\
\hline
(h) & \checkmark & & \checkmark & \checkmark &  \checkmark &  & \textbf{74.4} & \textbf{61.2} & \textbf{74.3} & \textbf{3.04} \\
\hline
\end{tabular}
\end{small}
\end{adjustbox}
\end{center}
\vspace{-10pt}
\caption[Ablation study of multi-task learning on 3D wireframe parsing.]
{%
The columns under ``supervisions'' indicate what losses and supervisions are used during training; the columns under ``metrics`` indicate the performance given such supervision during evaluation.  The second row shows the symbols of the feature maps; the third row shows the loss function names of the corresponding maps.  \enquote{CE} stands for the cross entropy loss, \enquote{SILog} loss is proposed by \cite{Eigen:2014:Depth}, and \enquote{Ord} represents the ordinary loss in \cite{Zhengqi:2018:MegaDepth}.  \enquote{/} indicates that the maps are not generated and thus not evaluable.
}
\label{tab:multi-task}
\end{table}
\end{small}

\nothing{
\paragraph{Depth Maps.}
As shown in Table \ref{tab:junction-depth}, supervision with the global depth map $\depthMap$ in training improves the junction depth map $\juncDepthMap$ estimate. However, the second row indicate that the estimated depth $\depthMap$ at the junction is not as good. So the use of $\juncDepthMap$ in training is also necessary.

\begin{table}
\centering
\begin{tabular}{c|cc|cc|c}
& \multicolumn{2}{c|}{training} & \multicolumn{2}{c|}{inference} & metrics \\
& $\juncDepthMap$ & $\depthMap$ & $\juncDepthMap$ & $\depthMap$ & SILog \\
\hline
\hline
(a) & \checkmark & & \checkmark & & 6.63 \\
(b) & \checkmark & \checkmark & & \checkmark & 9.47 \\
(c) & \checkmark & \checkmark & \checkmark & & \textbf{4.93} \\
\hline
\end{tabular}
\caption{Performance comparison of junction depth prediction. The columns under ``training'' indicate whether corresponding loss is enabled during the training and the columns under ``inference'' indicates where we pick the depth for each junctions.}
\label{tab:junction-depth}
\end{table}
}

\subsection{Comparison with 2D Wireframe Extraction}

One recent work related to our system is \cite{Huang:2018:LPW}, which extracts 2D wireframes from single RGB images. However, it has several fundamental differences from ours: 
1) It does not differentiate between corner C-junctions and occluding T-junctions.
2) Its outputs are only 2D wireframes while ours are 3D.
3) It trains two separated networks for detecting junctions and lines.
4) It detects texture lines while ours only detects geometric wireframes.

\begin{figure}
\centering
\includegraphics[width=0.85\linewidth]{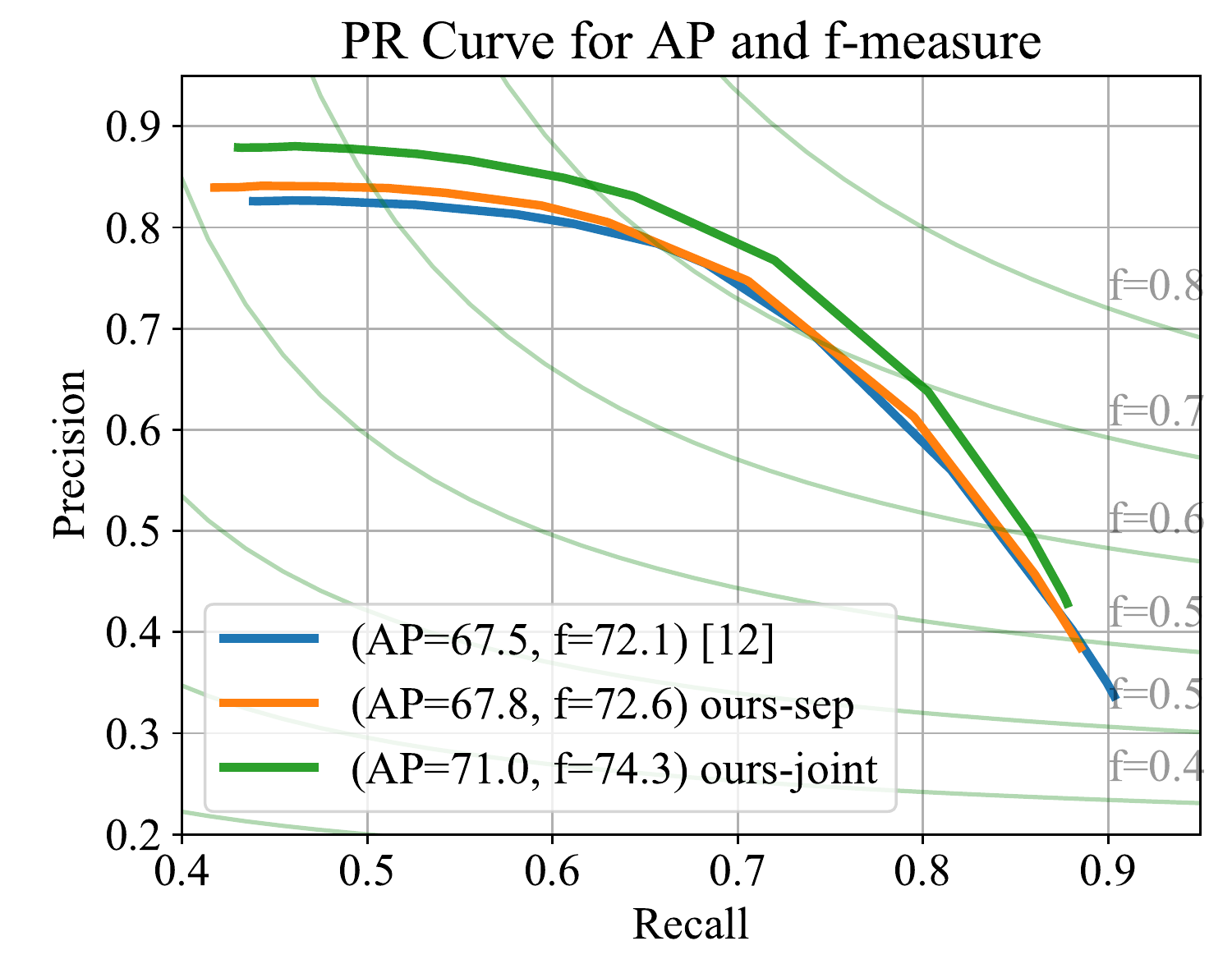}
\caption{Comparison with \cite{Huang:2018:LPW} on 2D wireframe detection. We improve the baseline method by 4 points.}
\label{fig:with_prev}
\end{figure}

In this experiment, we compare the performance with \cite{Huang:2018:LPW}. The goal of this experiment is to validate the importance of joint training. Therefore we follow the exact same training procedure and vectorization algorithms as in \cite{Huang:2018:LPW} except for the unified objective function and network structure. \Cref{fig:with_prev} shows the comparison of precision and recall curves evaluated on the test images, using the same evaluation metrics as in \cite{Huang:2018:LPW}. Note that due to different network designs, their model has about 30M parameters while ours only has 19M.  With fewer parameters, our system achieves 4-point AP improvement over \cite{Huang:2018:LPW} on the 2D wireframe detection task.

As a sanity check, we also train our network separately for lines and junctions, as shown by the green curve in \Cref{fig:with_prev}. The result is only slightly better than \cite{Huang:2018:LPW}. This experiment shows that our performance gain is from jointly trained objectives instead of neural network engineering.

\subsection{Vanishing Points and Depth Refinement}

\begin{table}[t]
\renewcommand{\arraystretch}{1.2}
\begin{adjustbox}{width=\linewidth,center}
\begin{tabular}{c|c|c|c|c|c}
\hline
& $\mathrm{avg}[E_{\mathrm{V}}]$ & $\mathrm{med}[E_{\mathrm{V}}]$ & $\mathrm{avg}[E_{\mathrm{f}}]$ & $\mathrm{med}[E_{\mathrm{f}}]$ & failures \\ \hline\hline
Ours & $\mathbf{2.69^\circ}$ & $1.55^\circ$ &  $\mathbf{4.02\%}$  & 1.38\% & $\mathbf{2.3\%}$ \\
\cite{Fleet:2014:Lsd,toldo2008robust} &  $4.65^\circ$ & $\mathbf{0.14^\circ}$ & $12.40\%$ &  $\mathbf{0.21\%}$  &  $20.0\%$  \\ \hline
\end{tabular}
\end{adjustbox}
\caption{{Performance comparison between our method and LSD/J-linkage \cite{Fleet:2014:Lsd,toldo2008robust} for vanishing point detection.  $E_{\mathrm{V}}$ represents the angular error of $\vanishPoint_i$ in degree, $E_{\mathrm{f}}$ represents the relative error of the recovered camera focal lengths, and ``failures'' represents the percentage of cases whose $E_{\mathrm{V}} > 8^\circ$.}}
\label{fig:vp}
\end{table}
In \Cref{sec:lift}, vanishing point estimation and depth refinement are used in the last stage of the 3D wireframe representation. Their robustness and precision are critical to the final quality of the system output. In this section, we conduct experiments to evaluate the performance of these methods.

For vanishing point detection, \Cref{fig:vp} shows the performance comparison between our neural network-based method and the J-Linkage clustering algorithm \cite{toldo2008robust,tardif2009non} with the LSD line detector \cite{Fleet:2014:Lsd} on the SU3 dataset.  We find that our method is more robust in term of the percentage of failures and average error, while the traditional line cluster algorithm is more accurate when it does not fail.  This is because LSD/J-linkage applies a stronger geometric prior, while the neural network learns the concept from the data.  We choose our method for its simplicity and robustness, as the focus of this project is more on the 3D wireframe representation side, but we believe the performance can be further improved by engineering a hybrid algorithm or designing a better network structure.

\begin{figure}
  \centering
  \subfloat[Ground truth]{
    \frame{\includegraphics[width=0.315\linewidth]{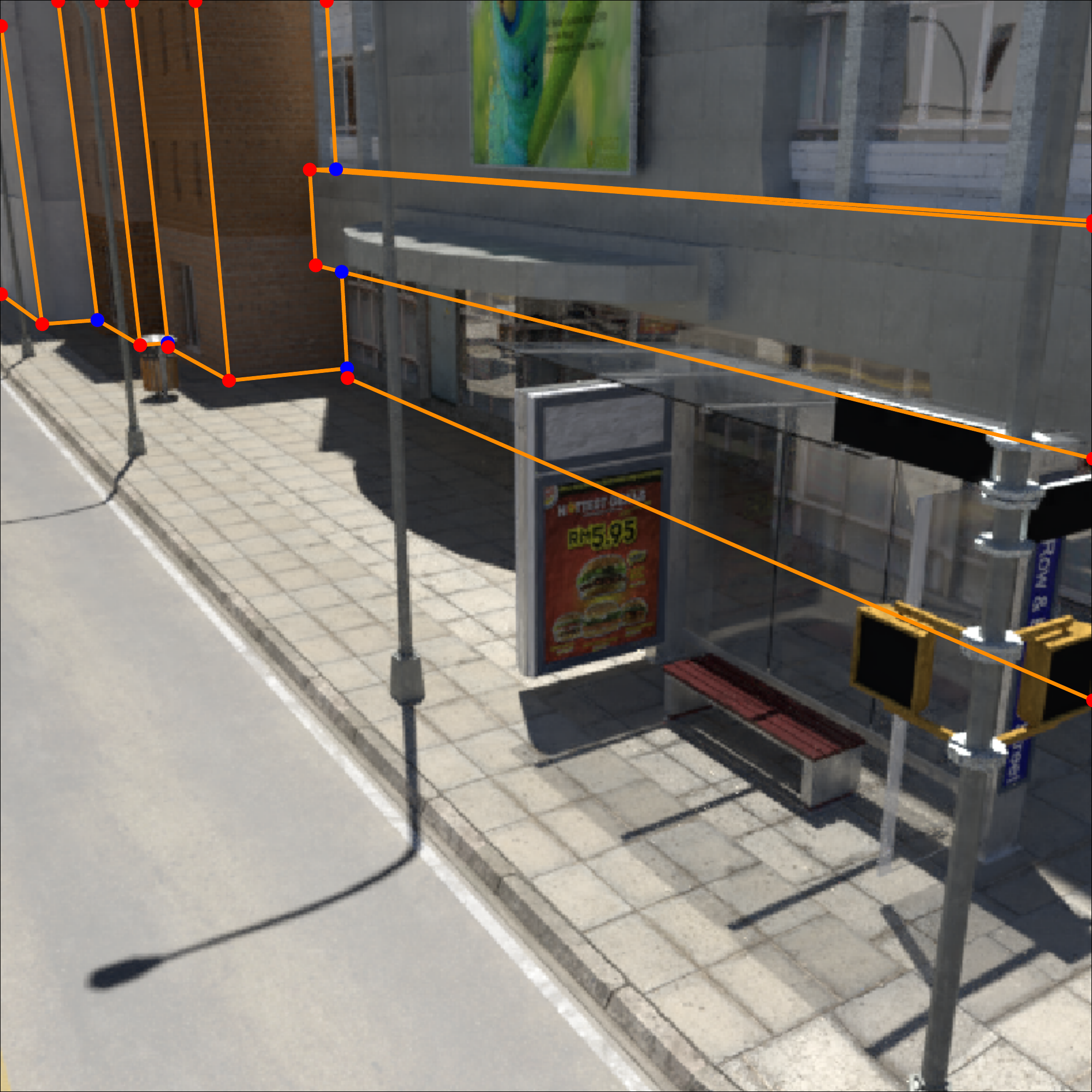}}
  }%
  \subfloat[Before refinement]{
    \frame{\includegraphics[width=0.315\linewidth]{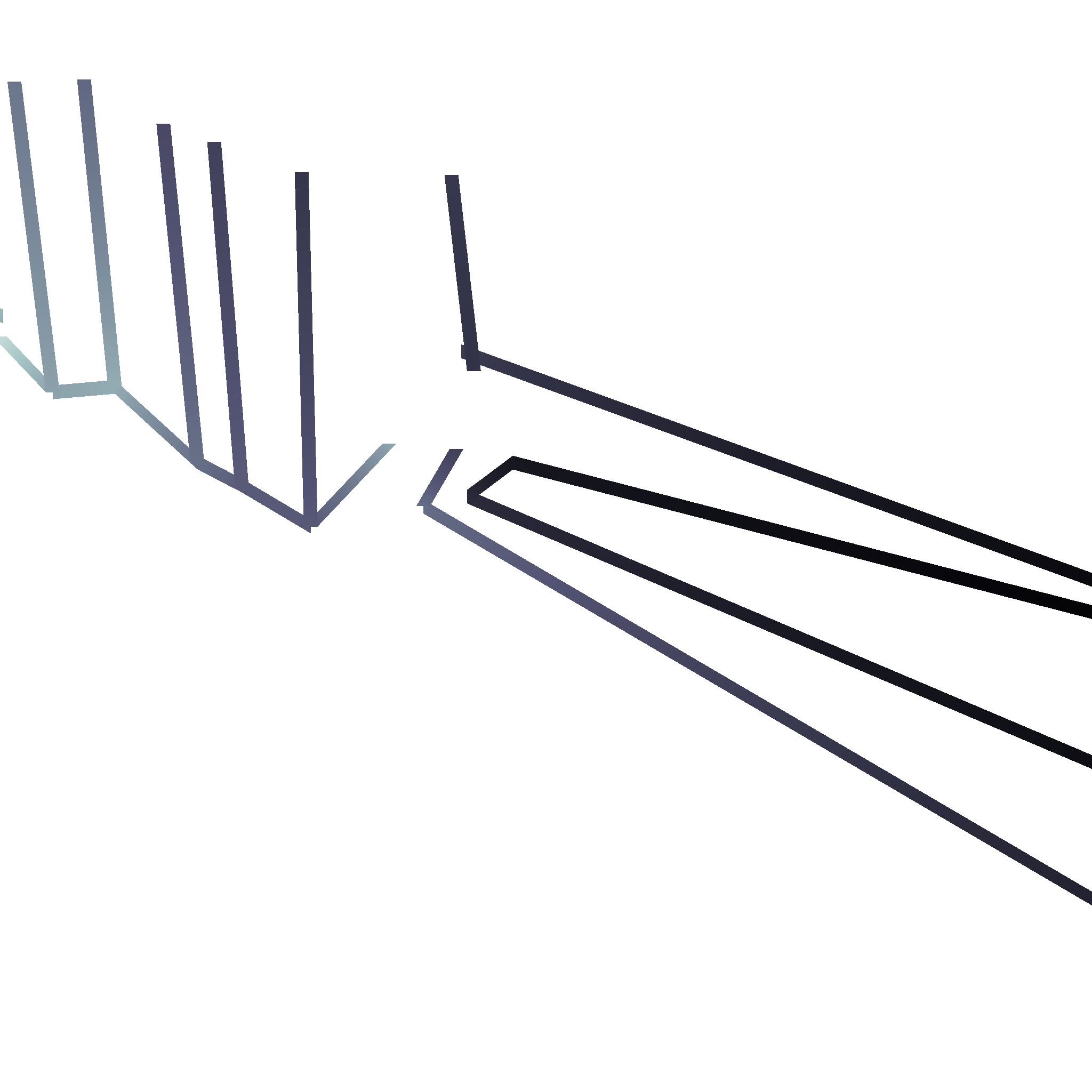}}
  }%
  \subfloat[After refinement]{
    \frame{\includegraphics[width=0.315\linewidth]{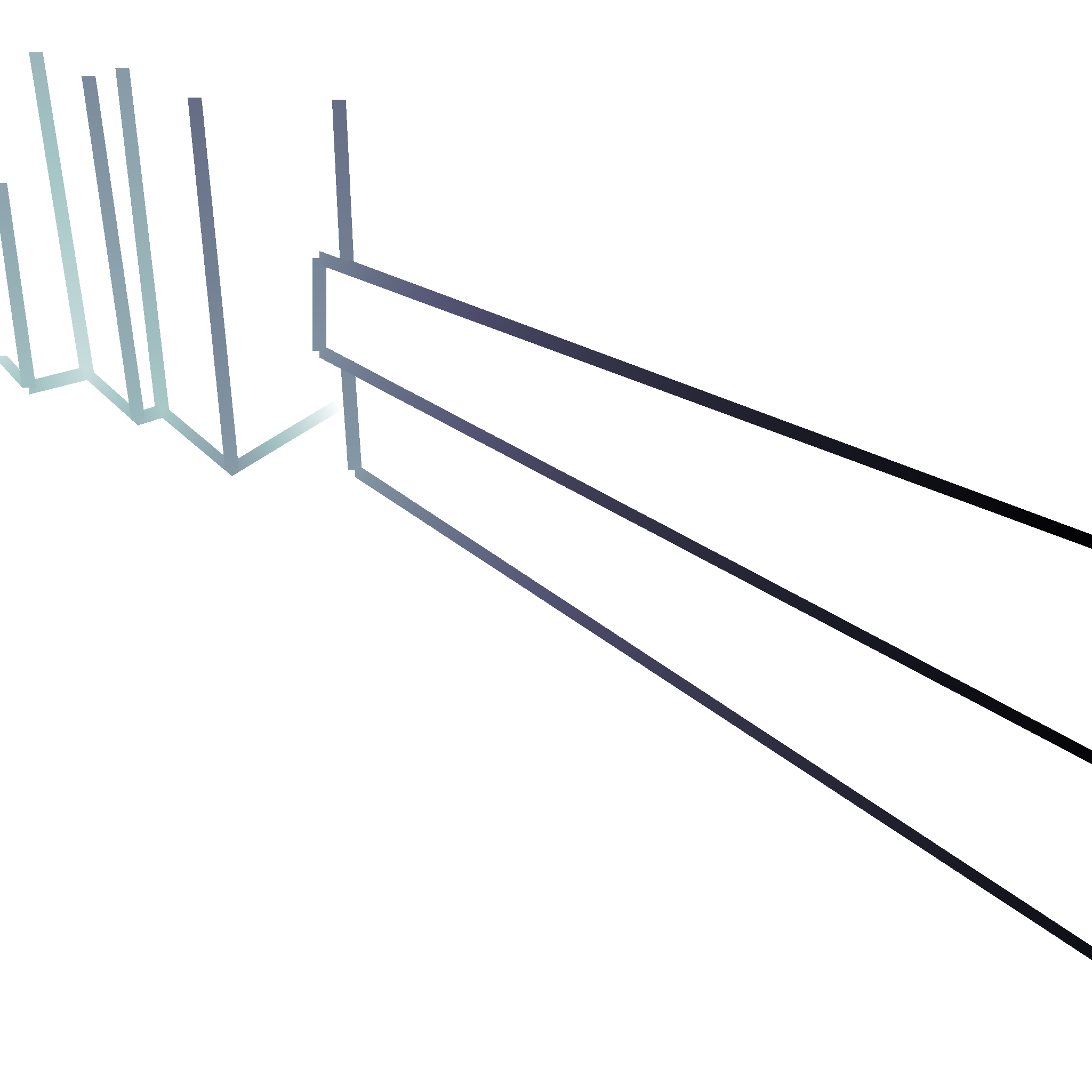}}
  }%
  \caption[Depth refinement with vanishing points.]
  {
 (b) shows a rendering of the wireframe from $\tilde \depth_{\vertex}$ from a slightly different view,
  while (c) shows the wireframe improved by the optimization in \Cref{sec:lifting}.
  }
  \label{fig:lifting}
\end{figure}

We also compare the error of the junction depth before and after depth refinement in term of SILog.  We find that on $65\%$ of the testing cases, the error is smaller after the refinement. This shows that the geometric constraints from vanishing points does help improve the accuracy of the junction depth in general. As shown in \Cref{fig:lifting}, the depth refinement also improves the visual quality of the 3D wireframe.  On the other hand, the depth refinement may not be as effective when the vanishing points are not precise enough, or the scene is too complex so that there are many erroneous lines in the wireframe.  Some failure cases can be found in the supplementary material.

\subsection{3D Wireframe Reconstruction Results}
\begin{figure*}[p]
\centering
    \begin{minipage}[t]{0.135\linewidth}
    \frame{\includegraphics[ width=\linewidth]{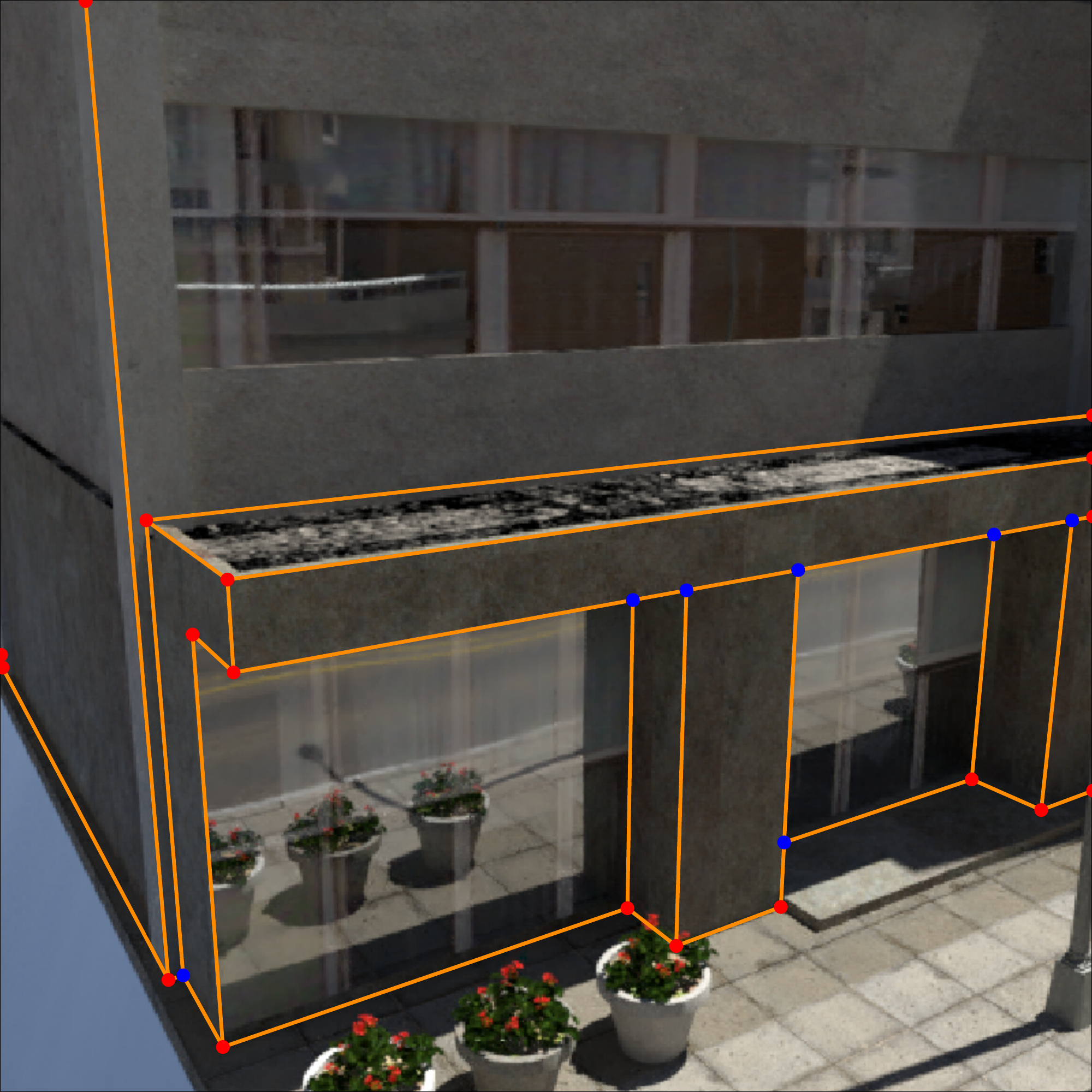}}\vspace{0.5mm}
    \frame{\includegraphics[ width=\linewidth]{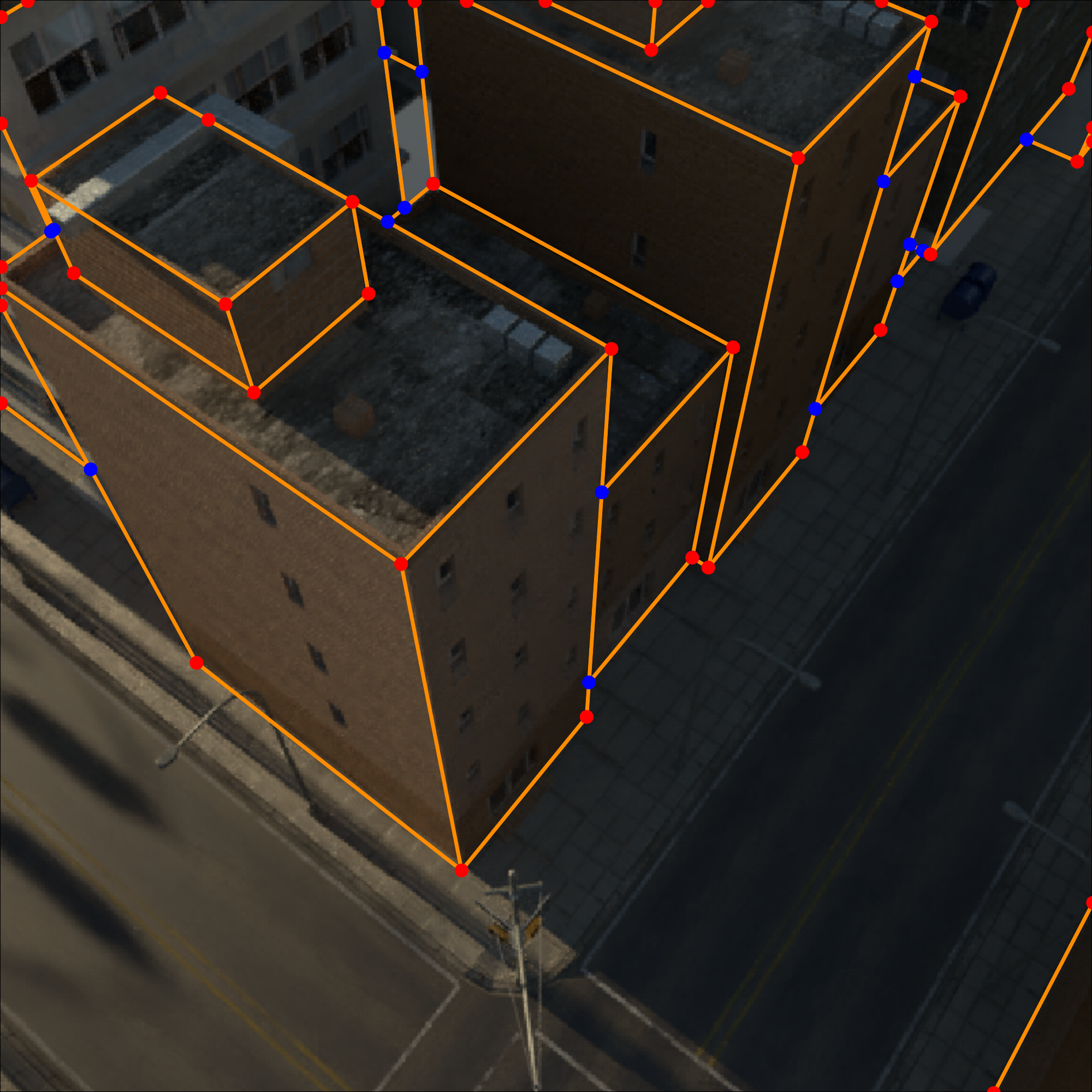}}\vspace{0.5mm}
    \frame{\includegraphics[ width=\linewidth]{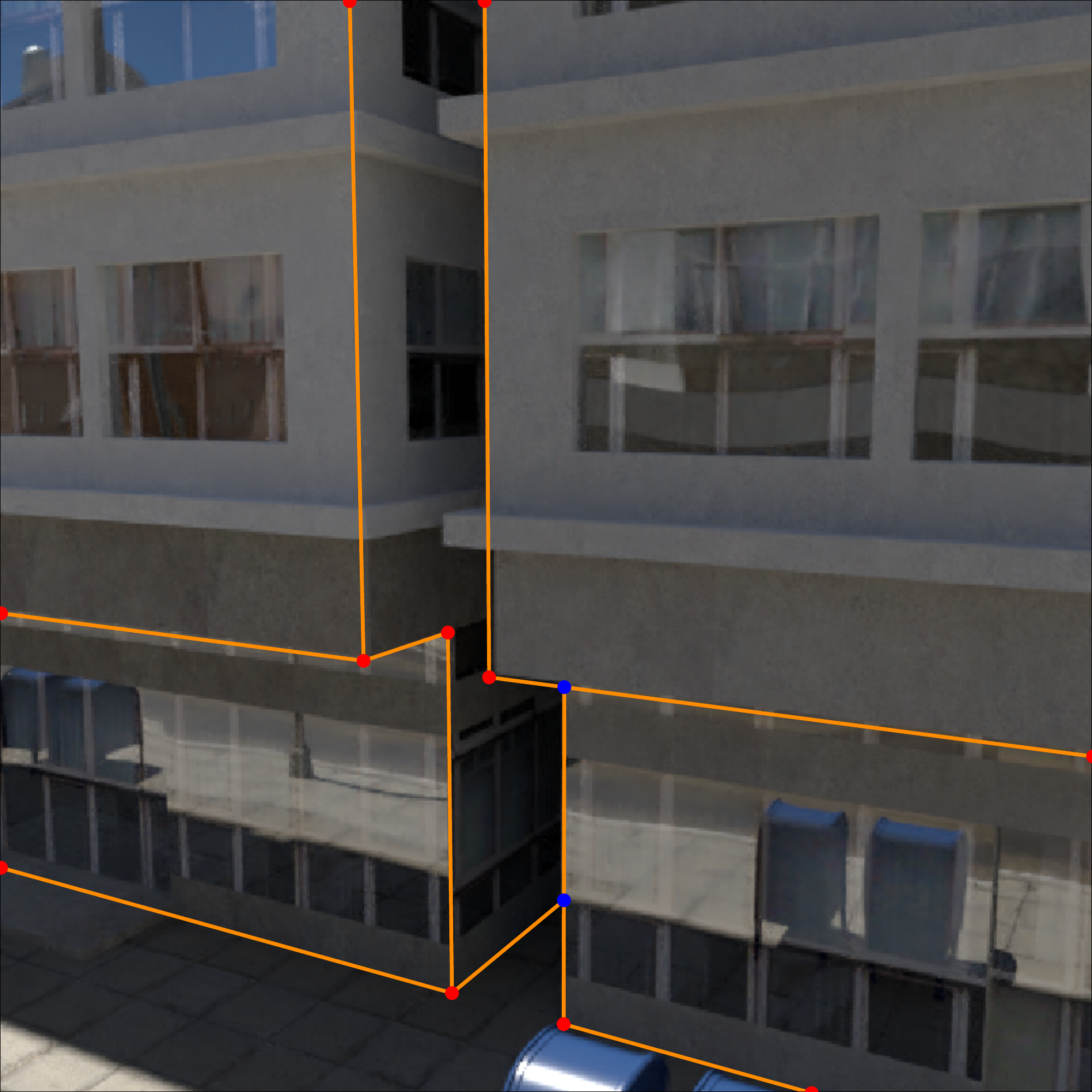}}\vspace{0.5mm}
    \frame{\includegraphics[ width=\linewidth]{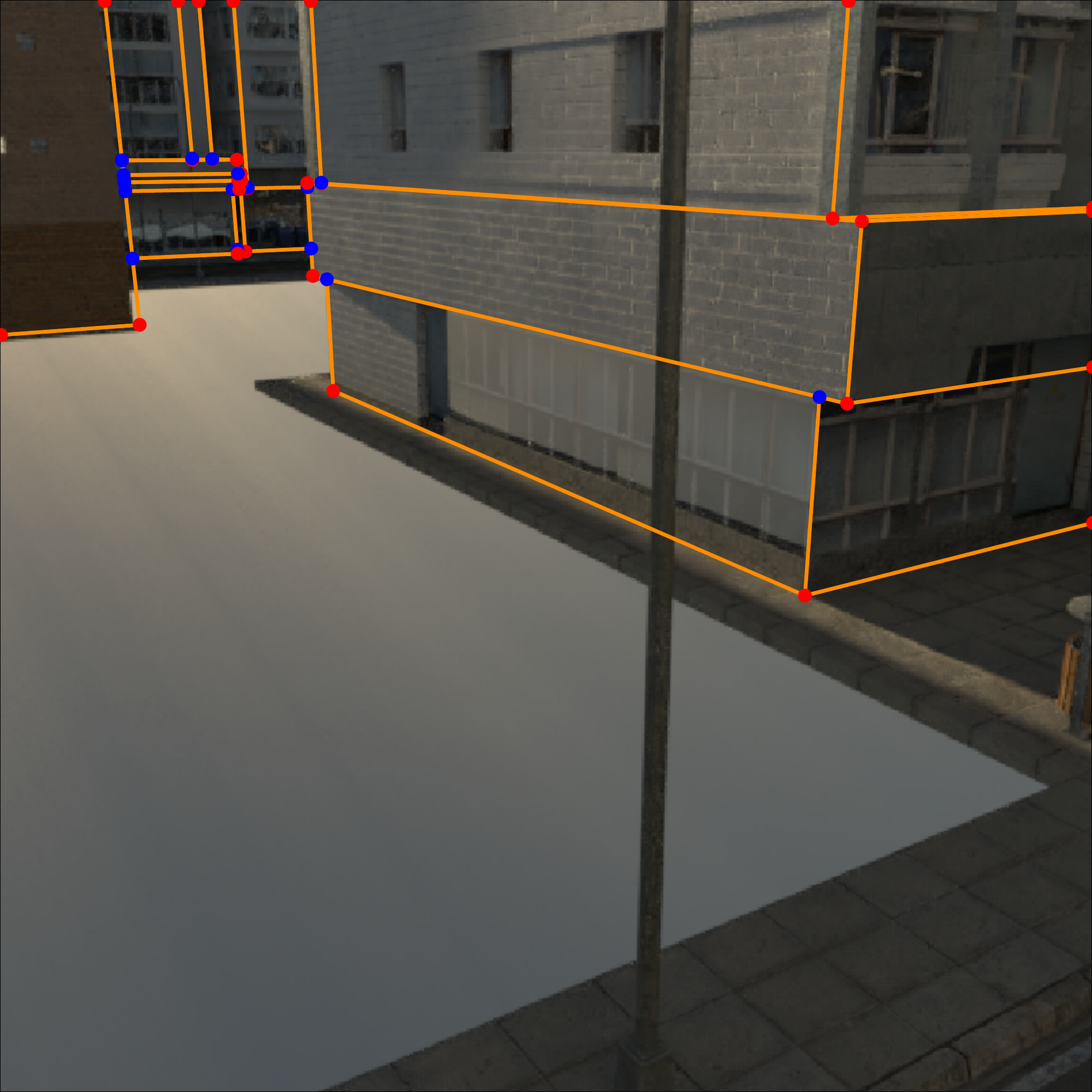}}\vspace{0.5mm}
    \frame{\includegraphics[ width=\linewidth]{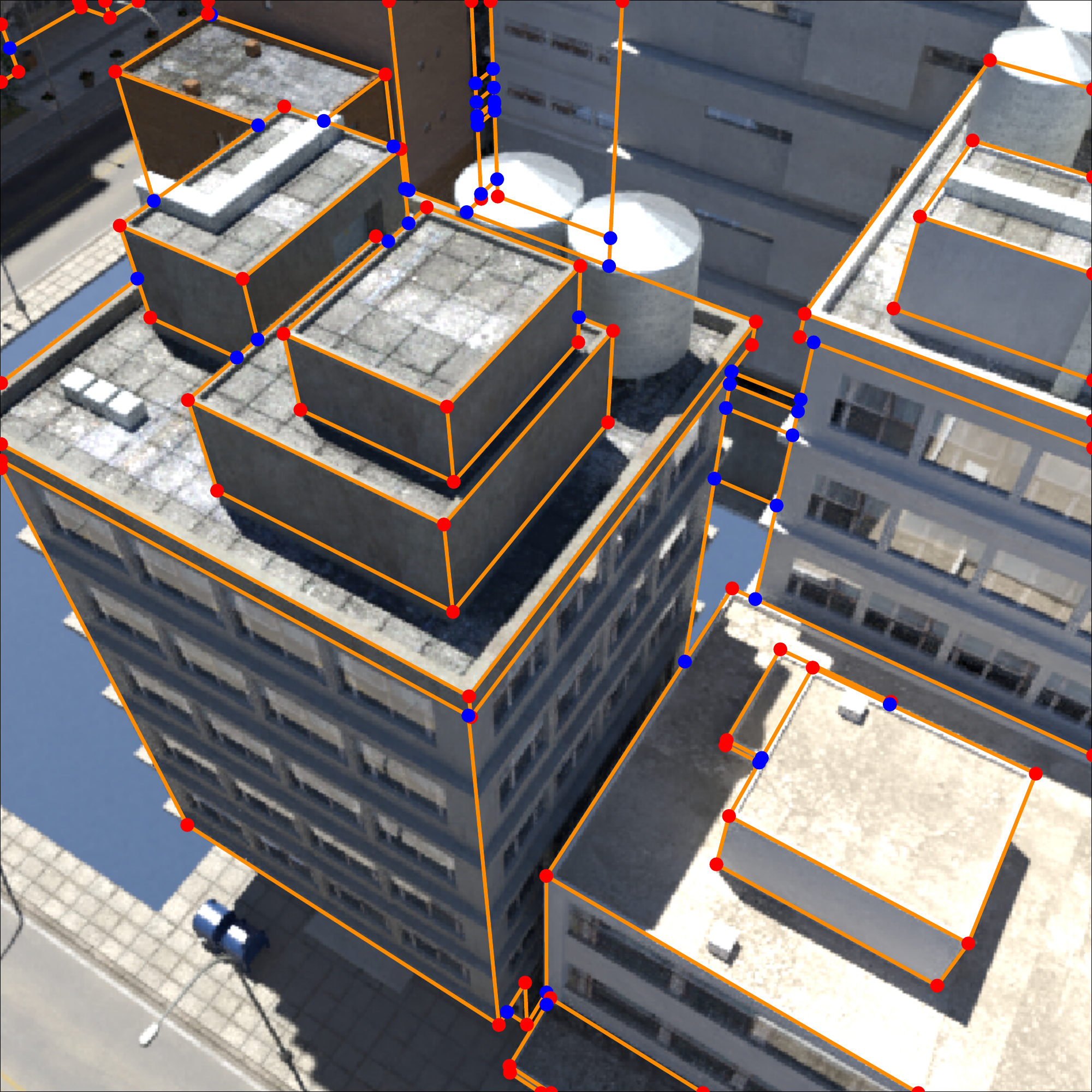}}
    \centering
    \small Ground truth 2D \nothing{  Wireframe}
    \end{minipage}
    \begin{minipage}[t]{0.135\linewidth}
    \frame{\includegraphics[ width=\linewidth]{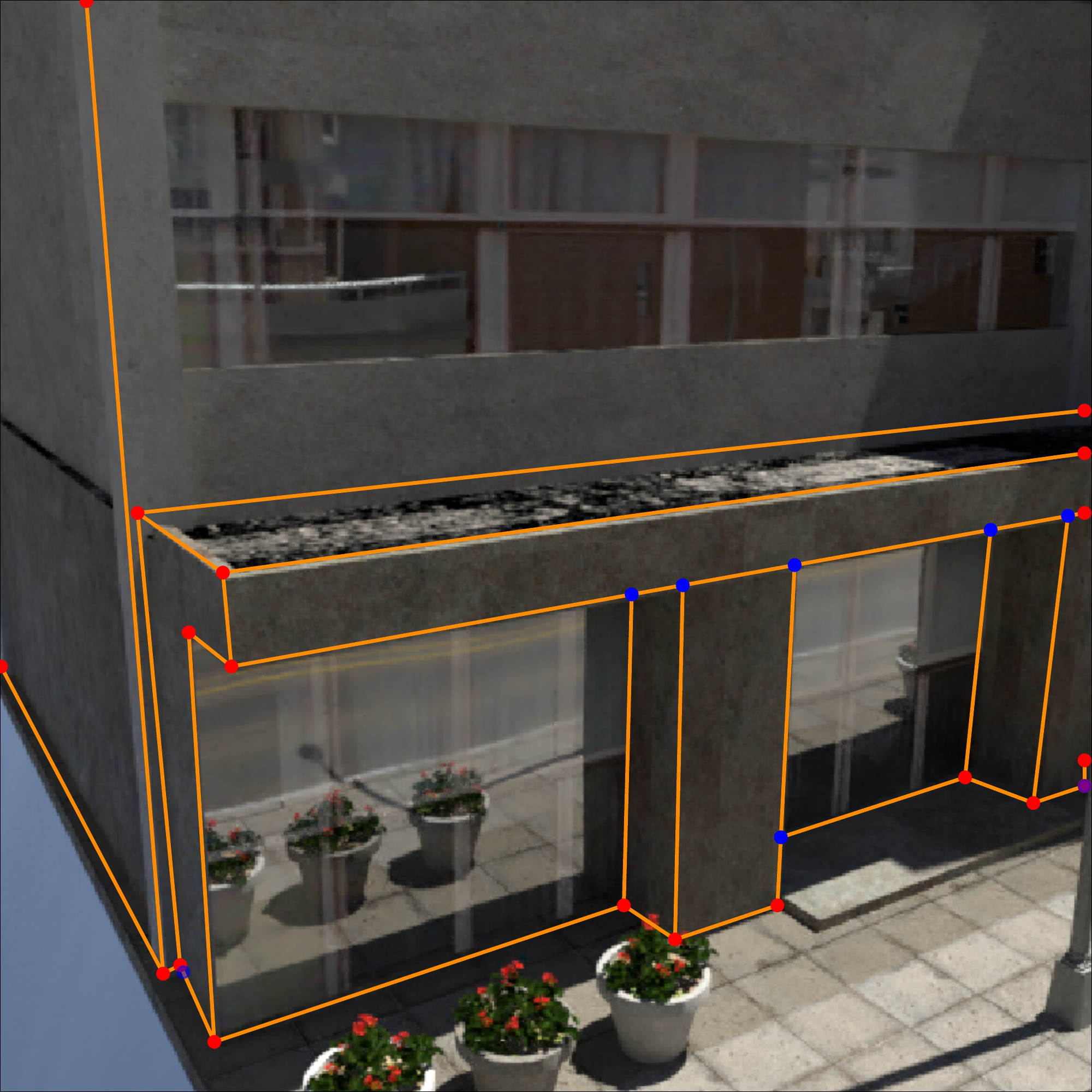}}\vspace{0.5mm}
    \frame{\includegraphics[ width=\linewidth]{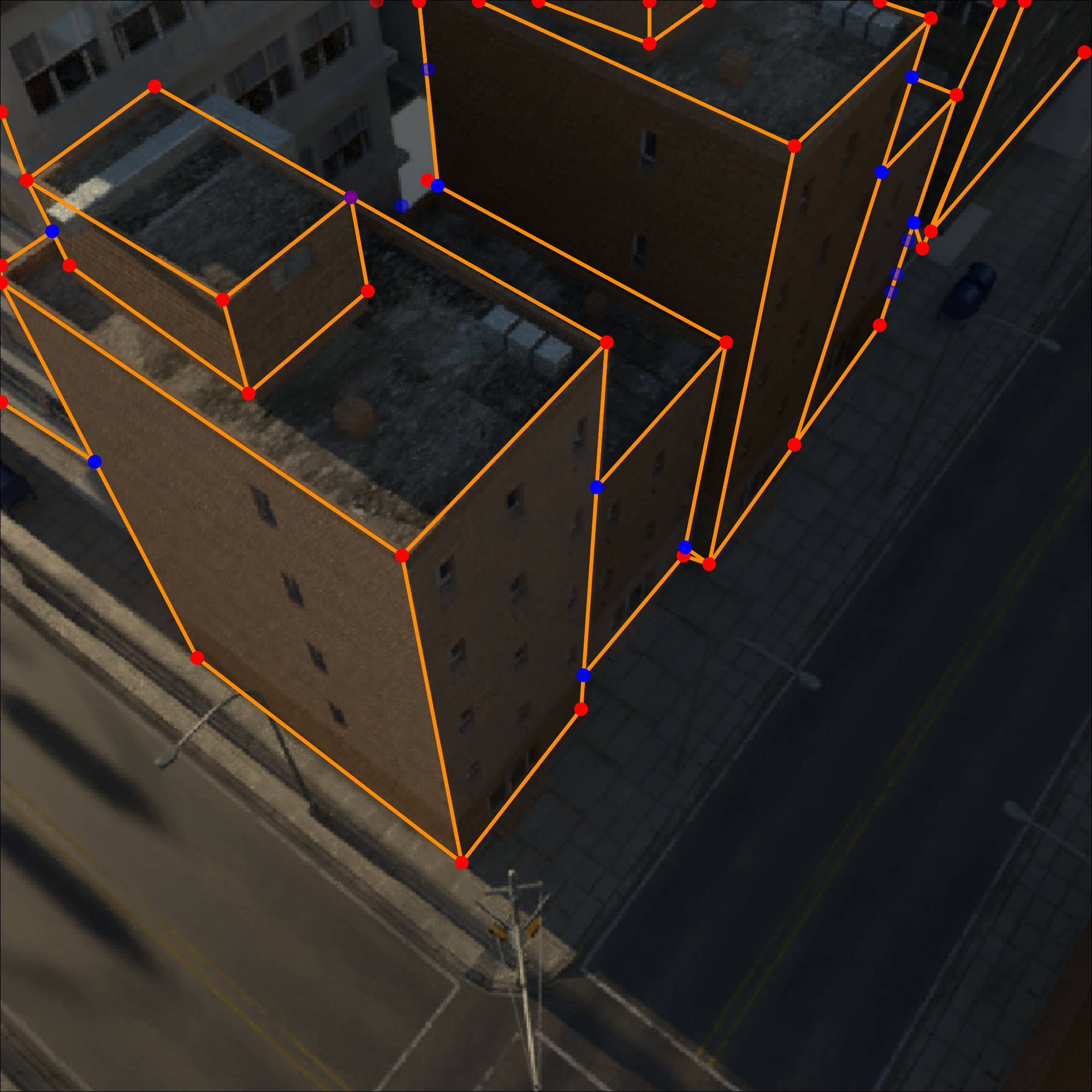}}\vspace{0.5mm}
    \frame{\includegraphics[ width=\linewidth]{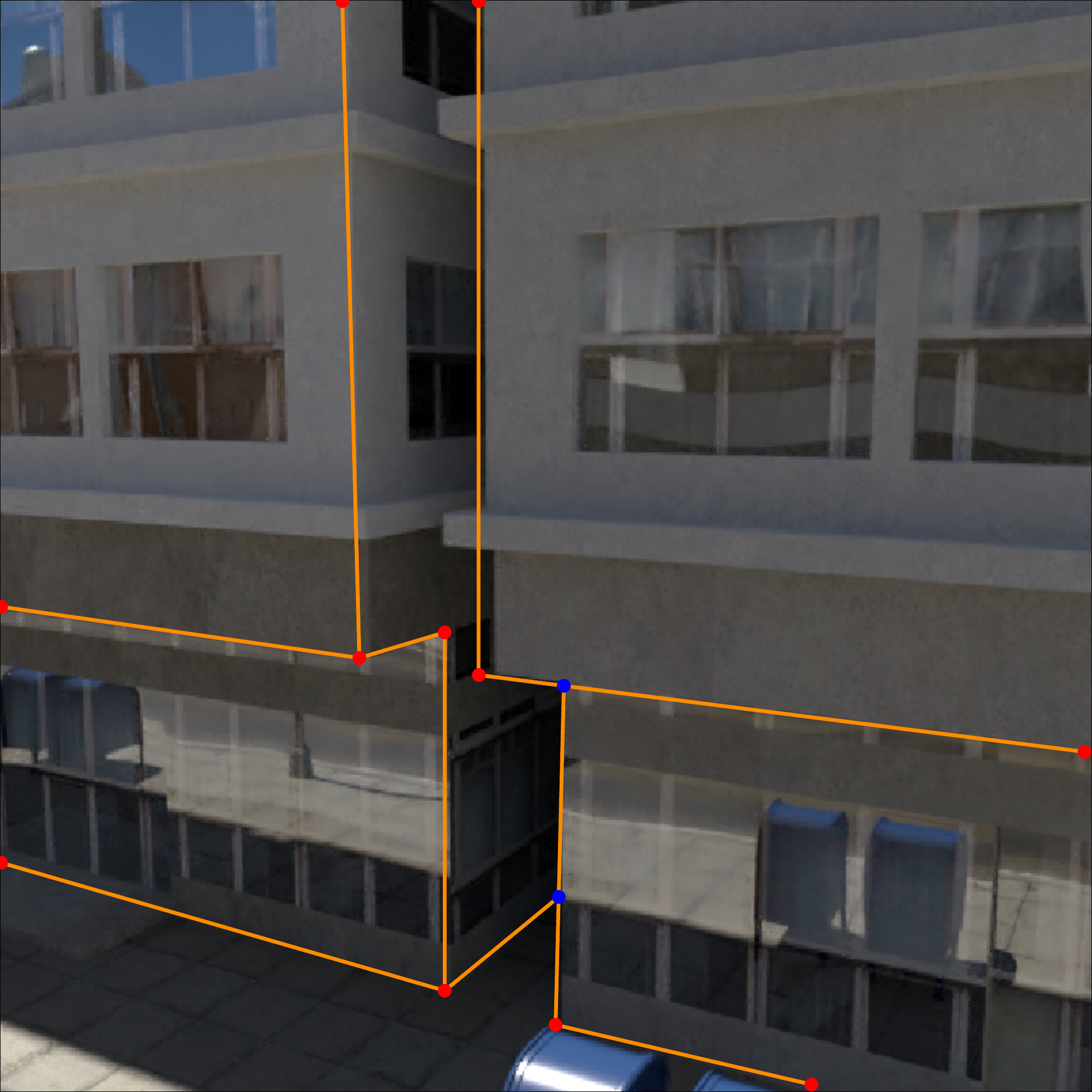}}\vspace{0.5mm}
    \frame{\includegraphics[ width=\linewidth]{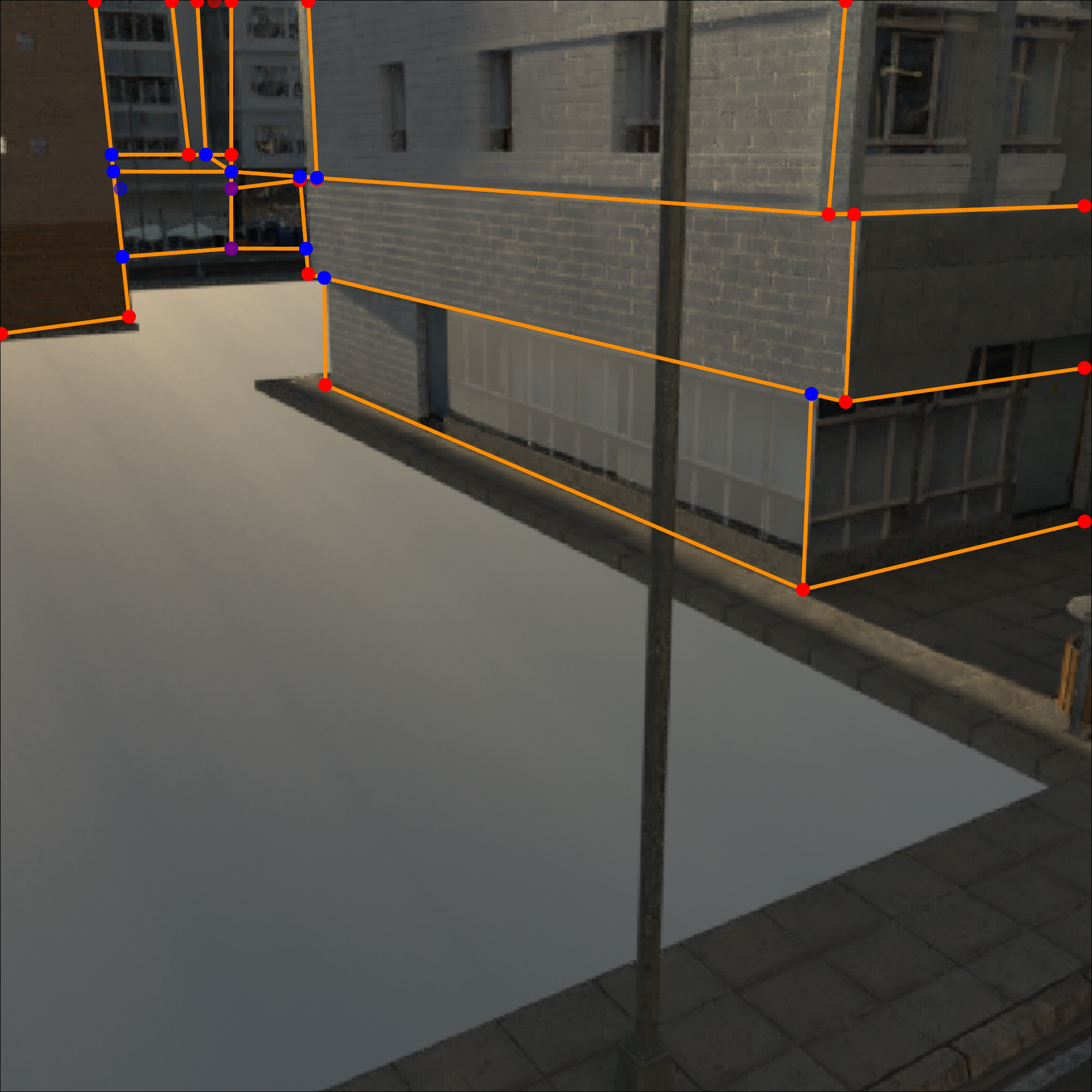}}\vspace{0.5mm}
    \frame{\includegraphics[ width=\linewidth]{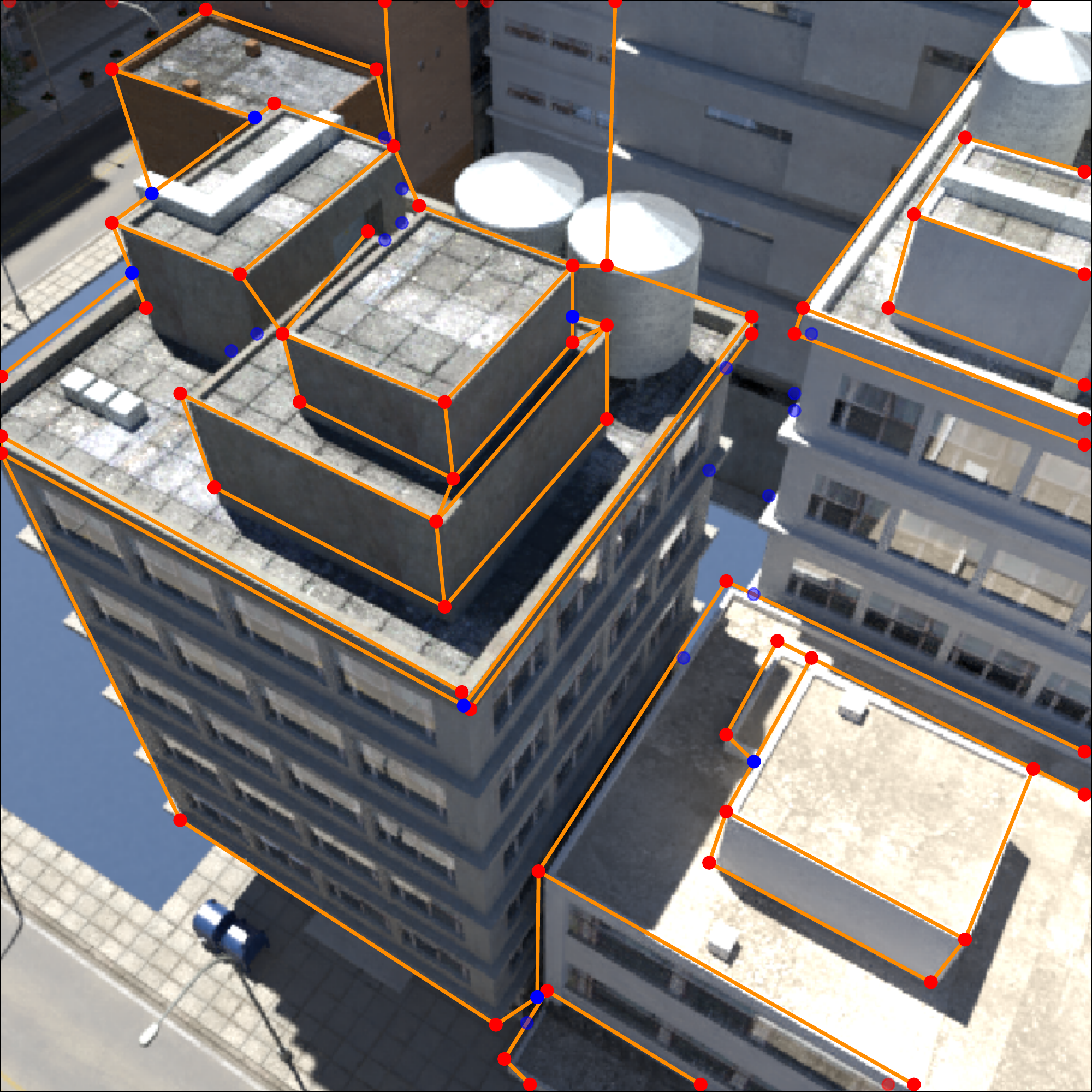}}
    \centering
    \small Our Inferred 2D
    \end{minipage}
    \begin{minipage}[t]{0.135\linewidth}
    \frame{\includegraphics[ width=\linewidth]{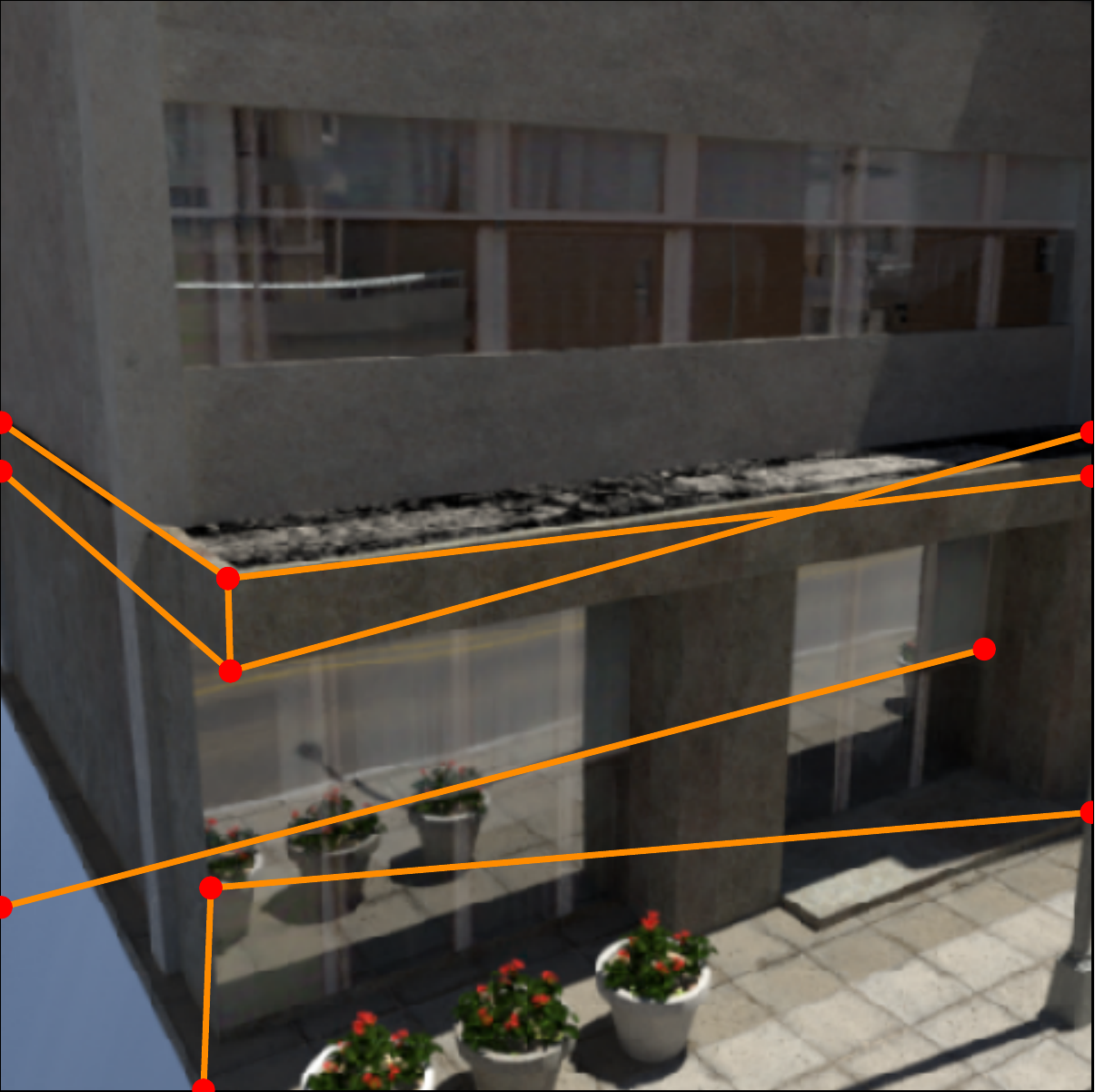}}\vspace{0.5mm}
    \frame{\includegraphics[ width=\linewidth]{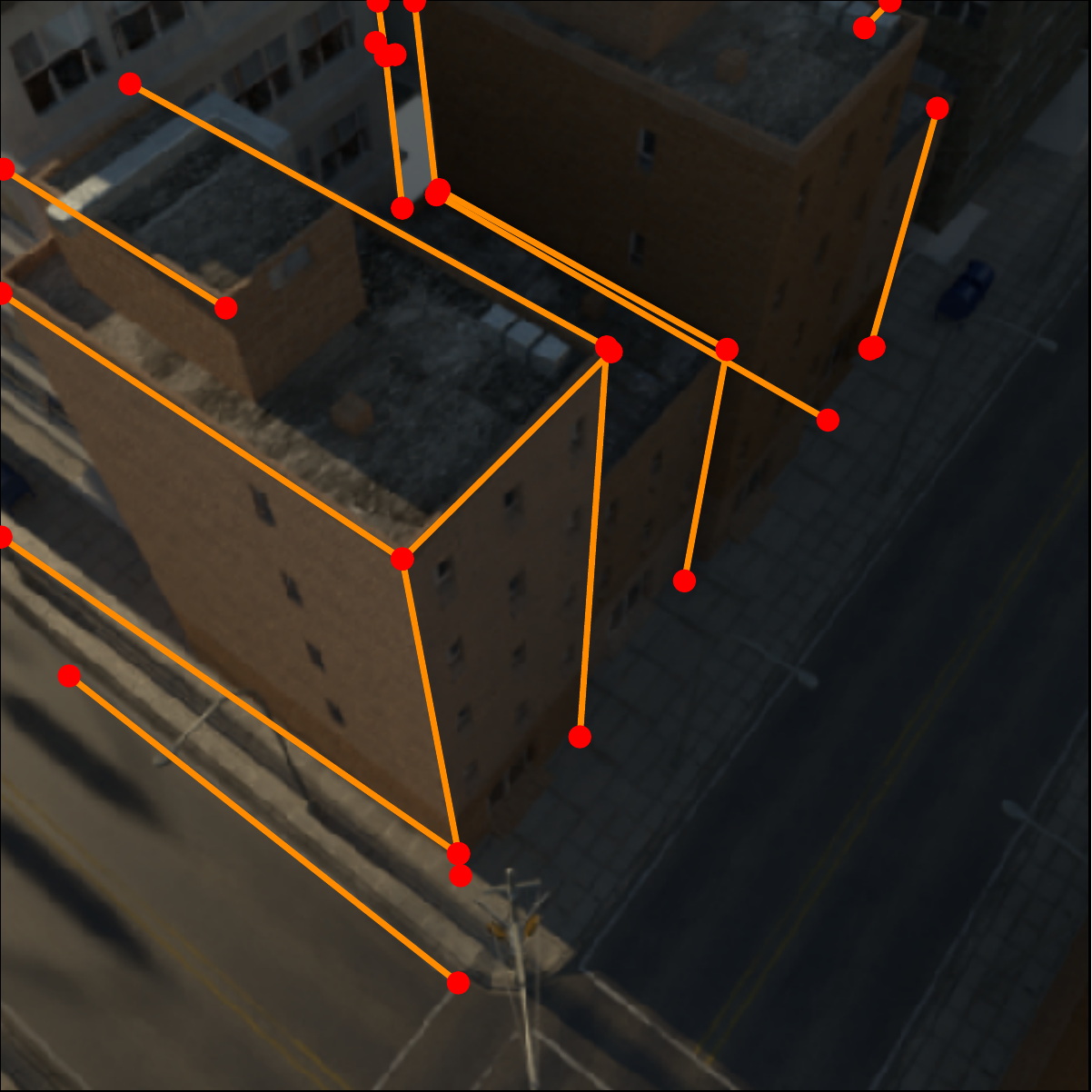}}\vspace{0.5mm}
    \frame{\includegraphics[ width=\linewidth]{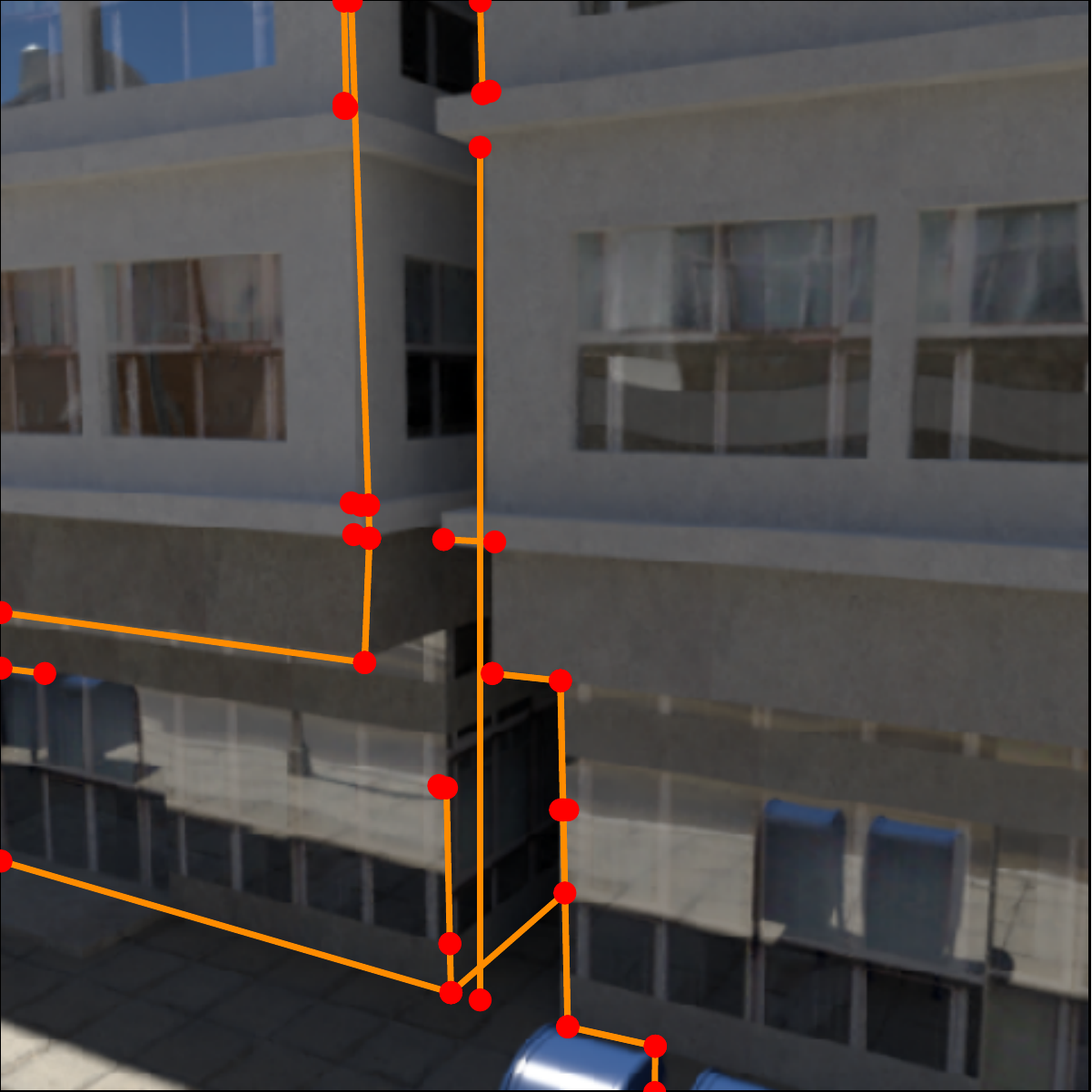}}\vspace{0.5mm}
    \frame{\includegraphics[ width=\linewidth]{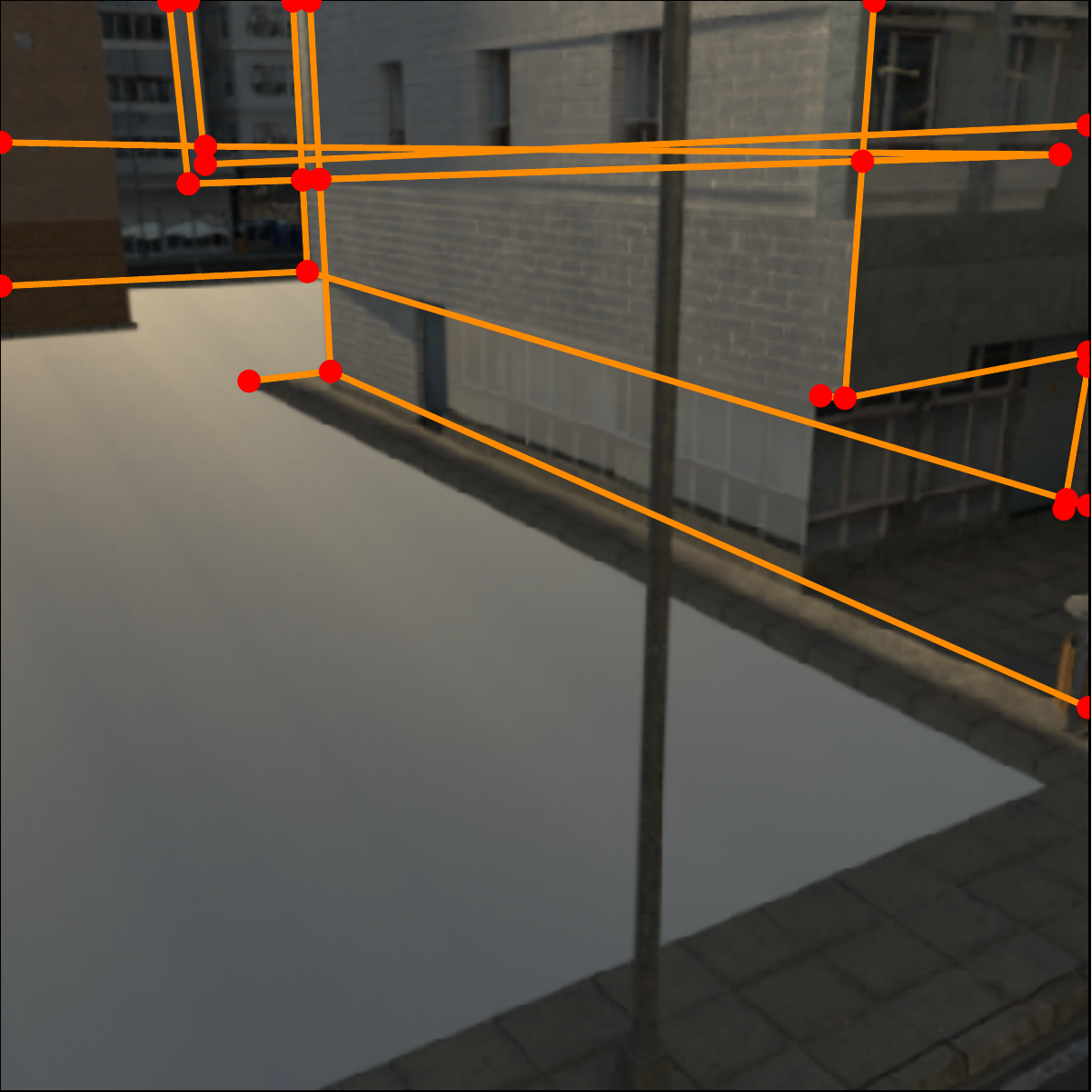}}\vspace{0.5mm}
    \frame{\includegraphics[ width=\linewidth]{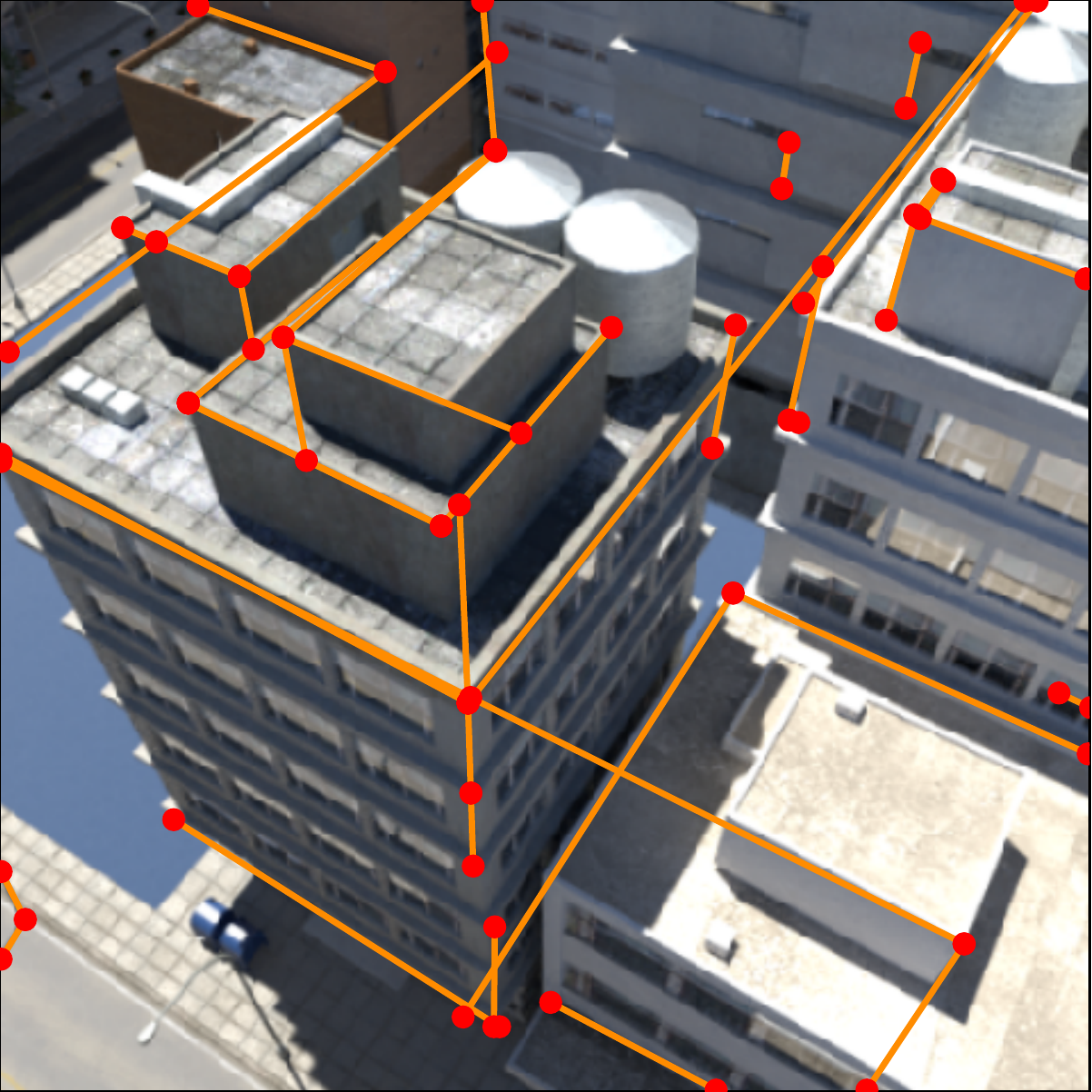}}
    \centering
    \small \cite{Huang:2018:LPW} Inferred 2D \nothing{  Wireframe}
    \end{minipage}
    \hfill
    \begin{minipage}[t]{0.135\linewidth}
    \frame{\includegraphics[width=\linewidth]{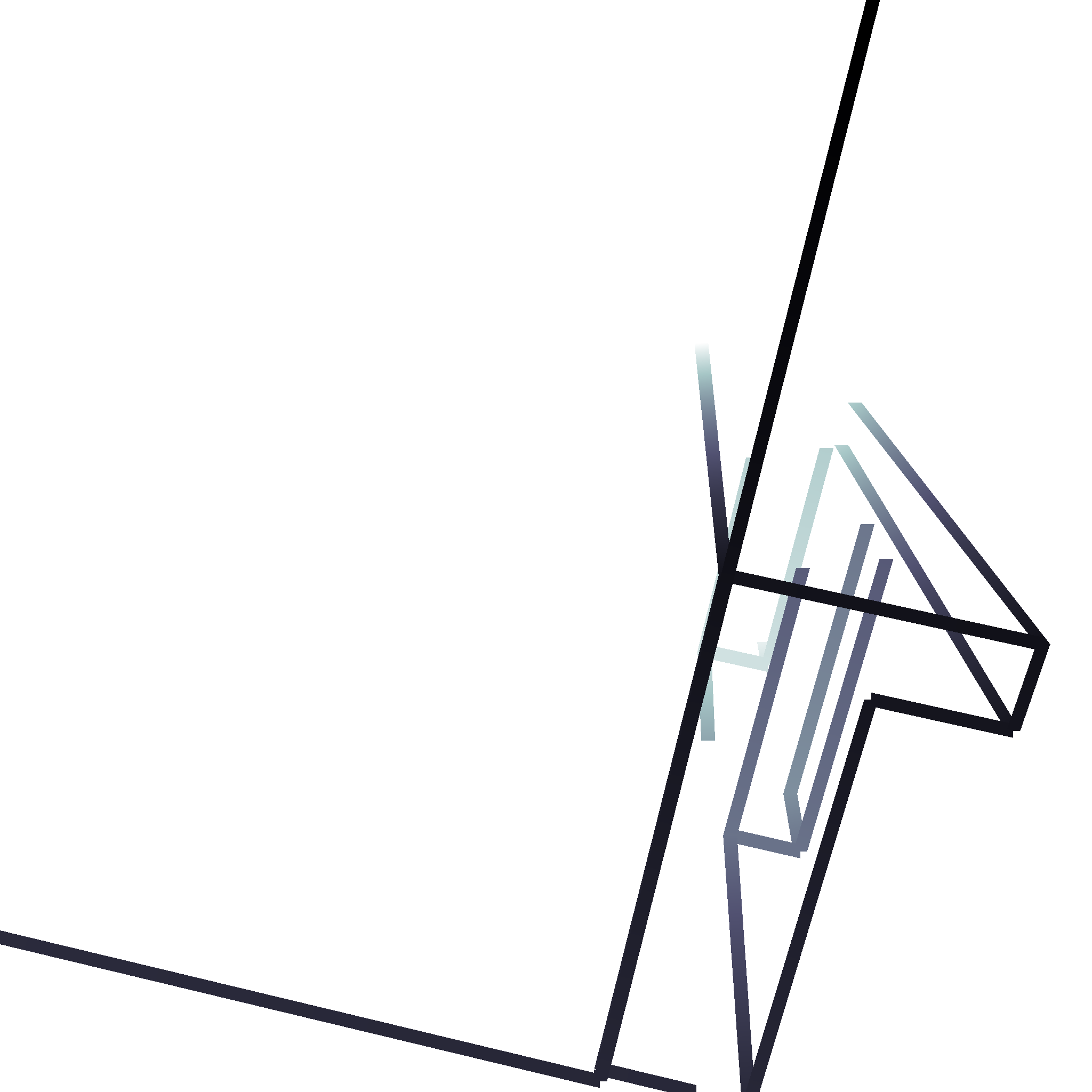}}\vspace{0.5mm}
    \frame{\includegraphics[width=\linewidth]{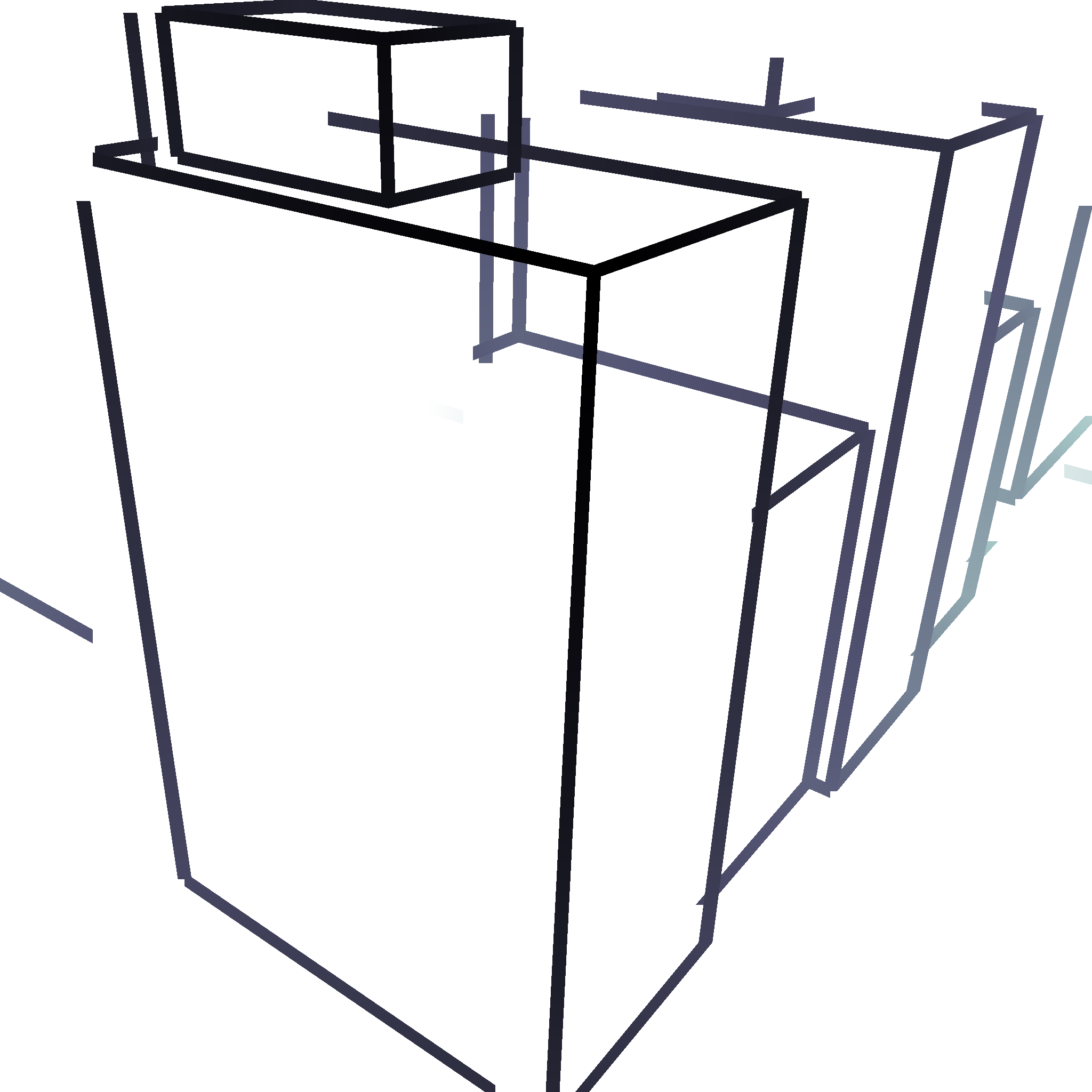}}\vspace{0.5mm}
    \frame{\includegraphics[width=\linewidth]{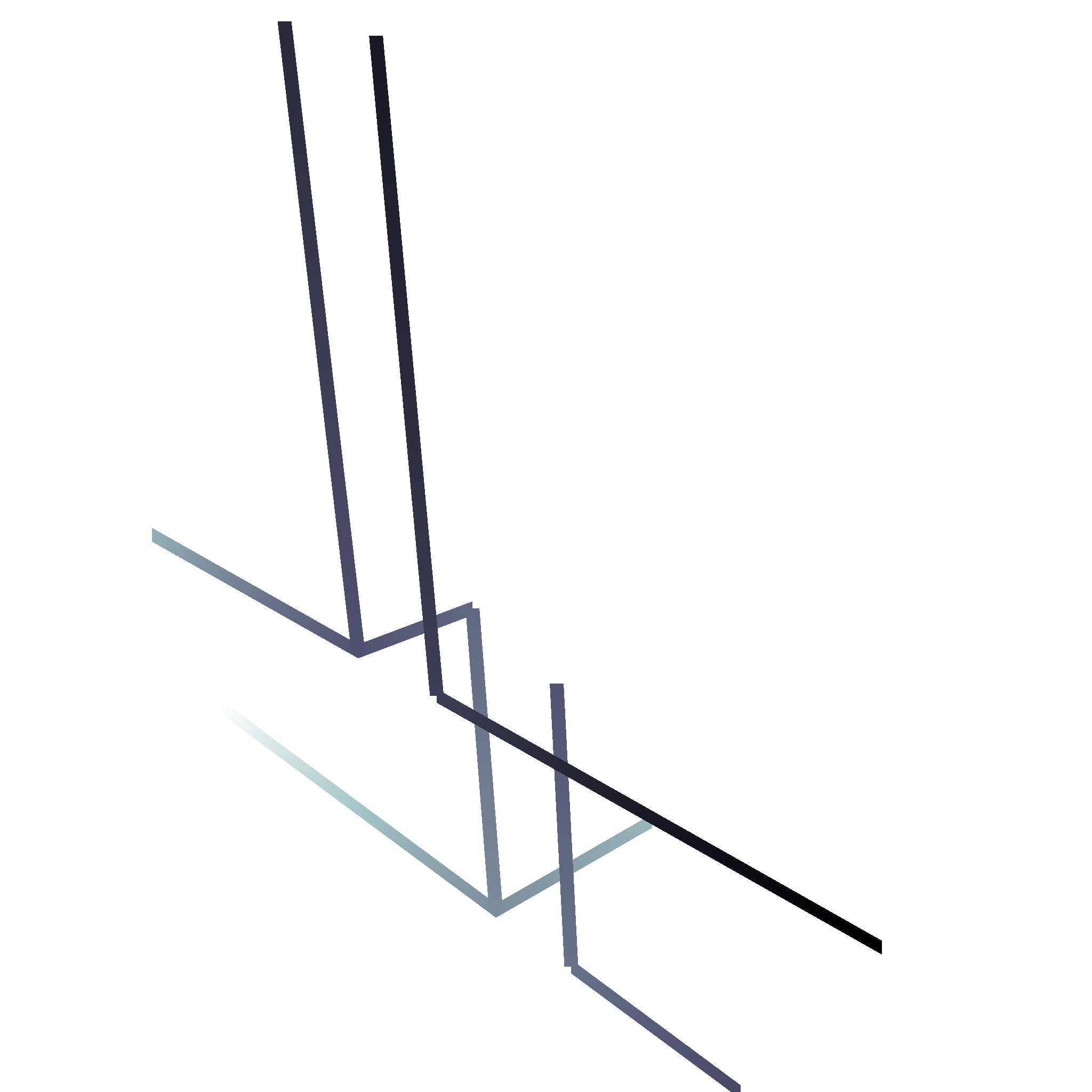}}\vspace{0.5mm}
    \frame{\includegraphics[width=\linewidth]{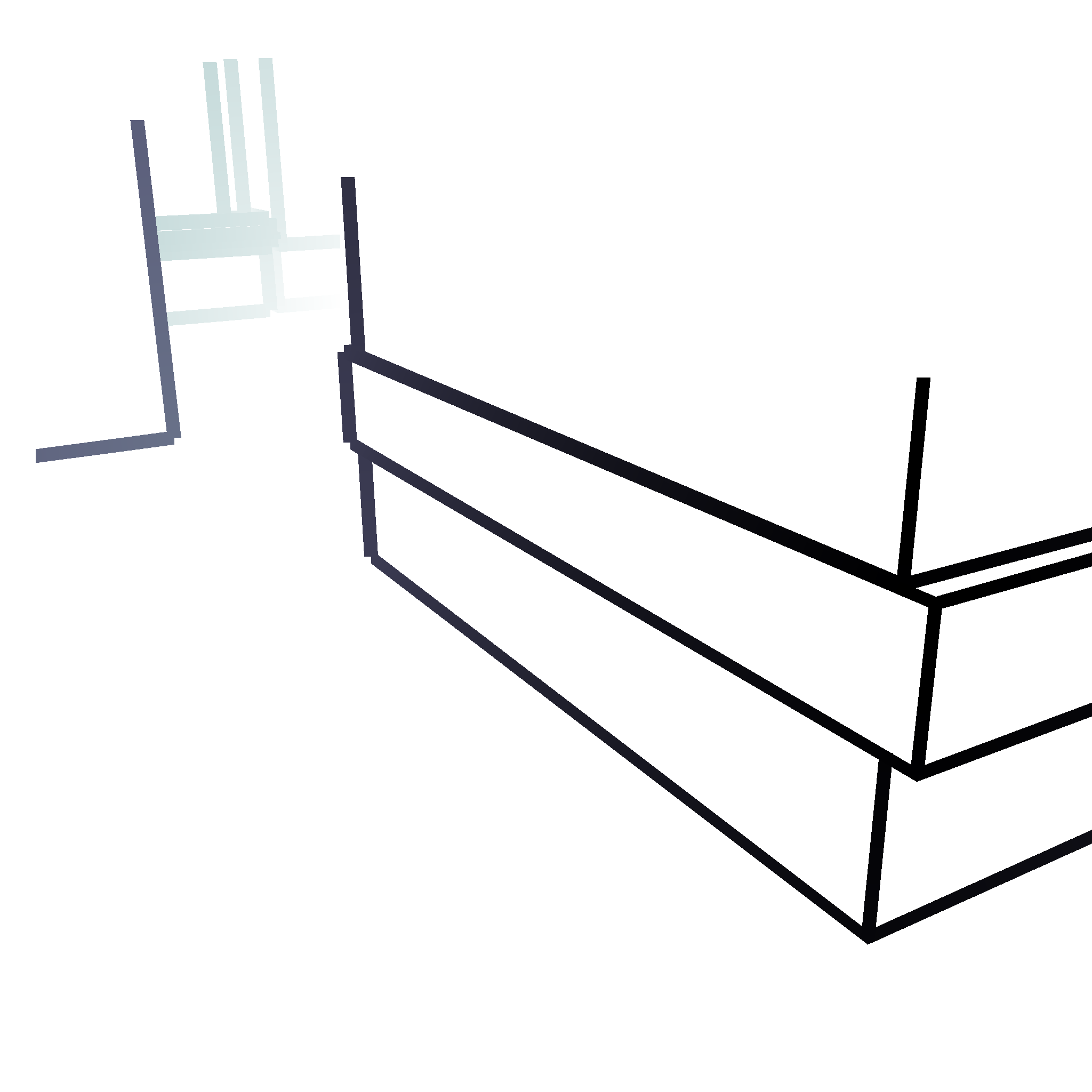}}\vspace{0.5mm}
    \frame{\includegraphics[width=\linewidth]{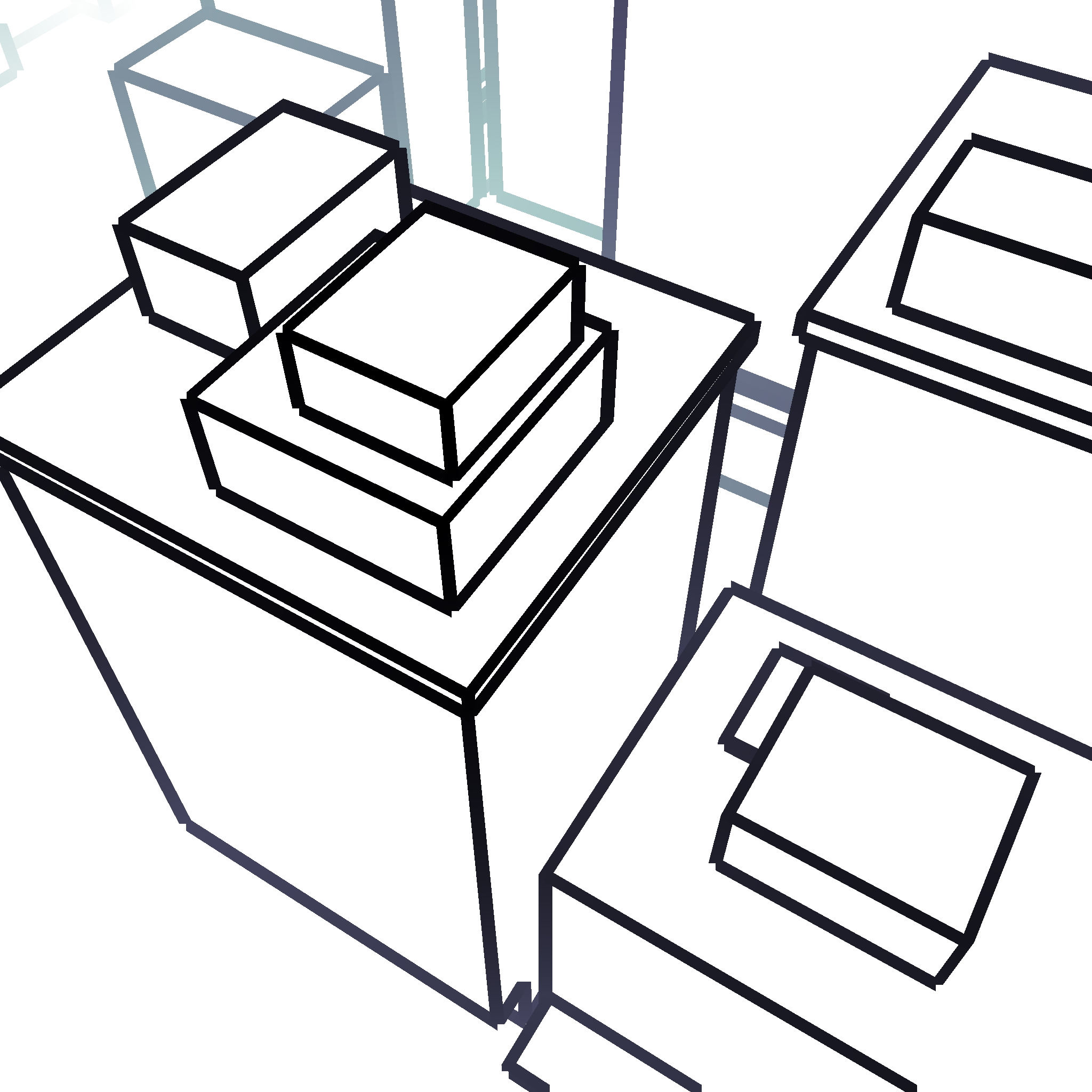}}
    \centering
    \small Ground truth 3D\nothing{  wireframe}
    \end{minipage}
    \begin{minipage}[t]{0.135\linewidth}
    \frame{\includegraphics[ width=\linewidth]{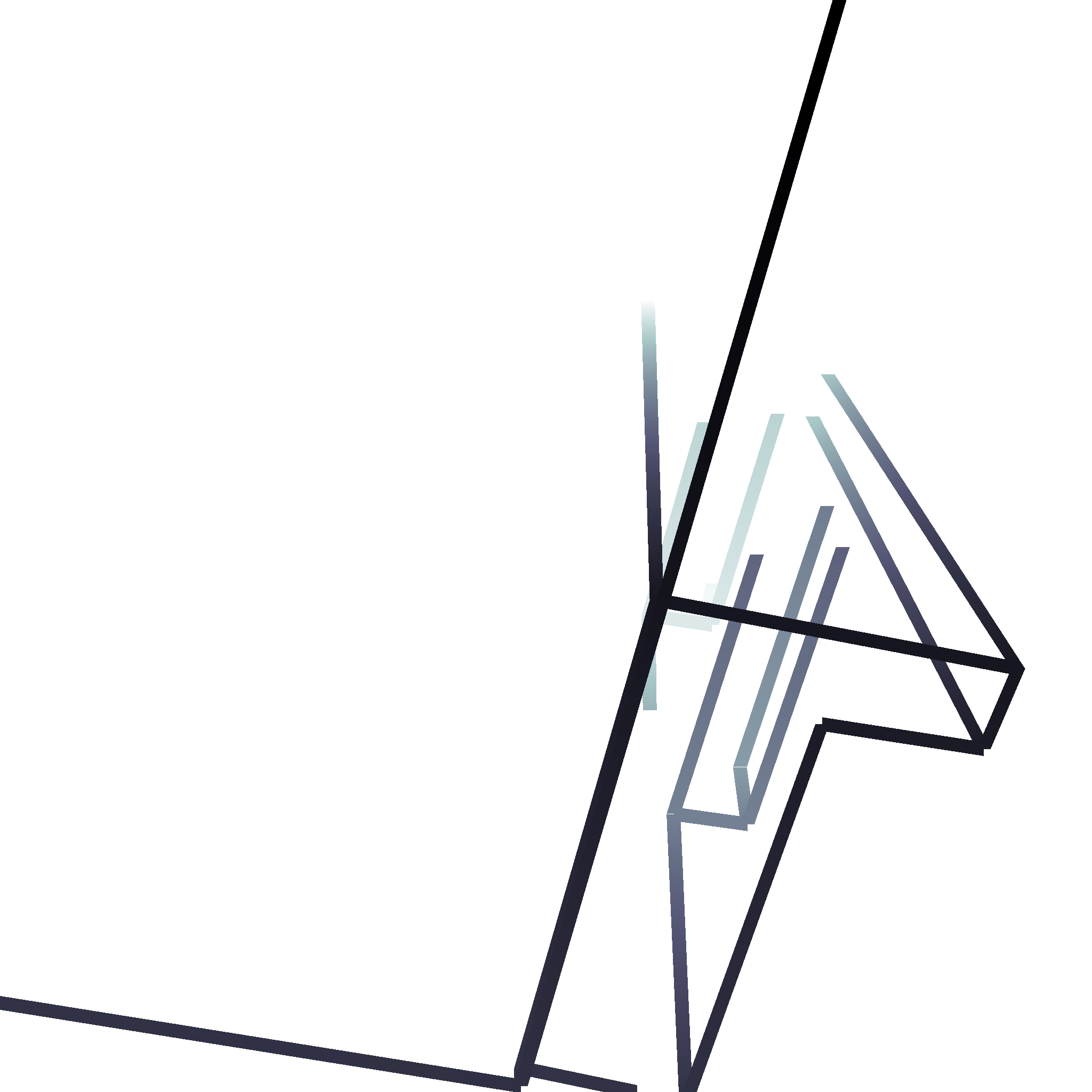}}\vspace{0.5mm}
    \frame{\includegraphics[ width=\linewidth]{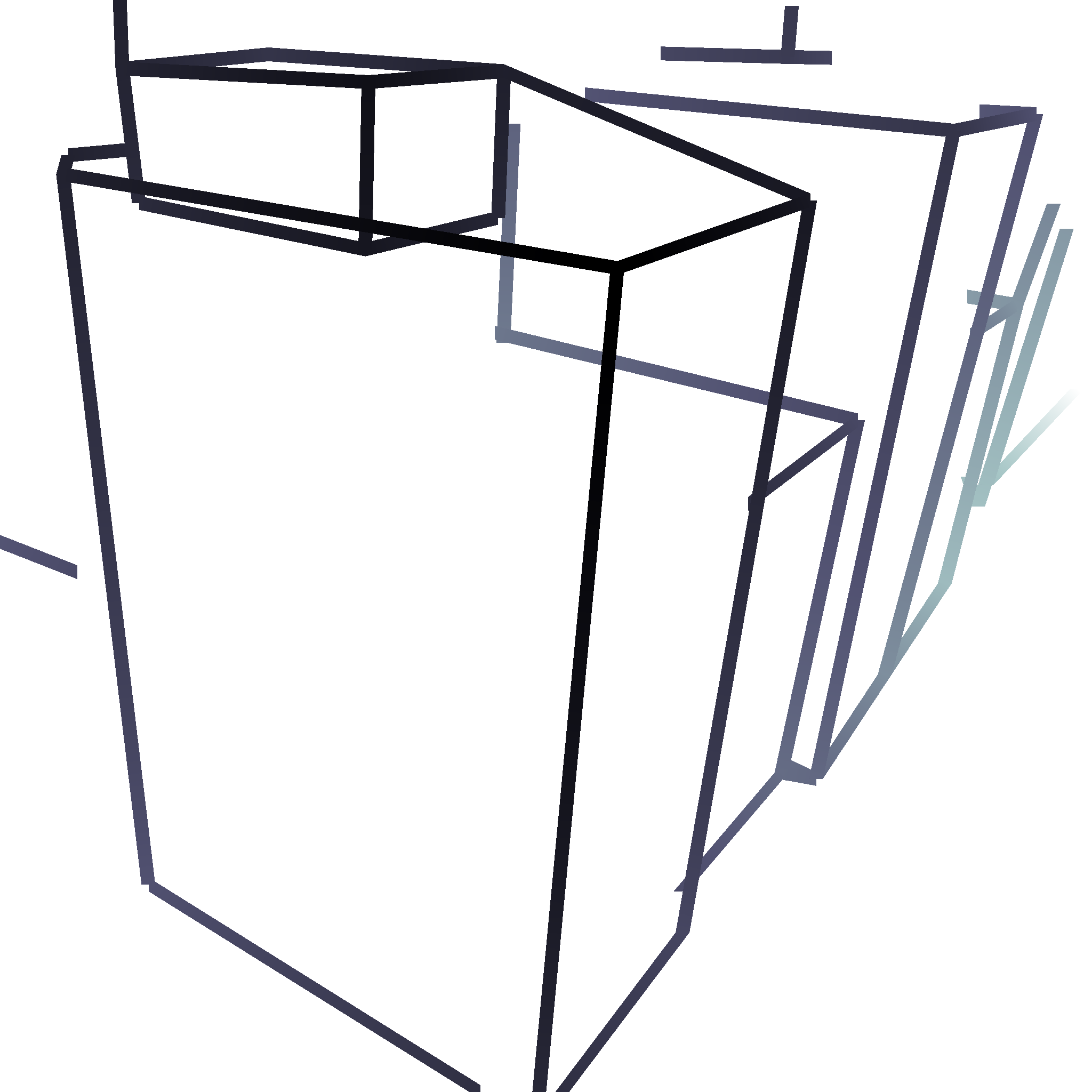}}\vspace{0.5mm}
    \frame{\includegraphics[ width=\linewidth]{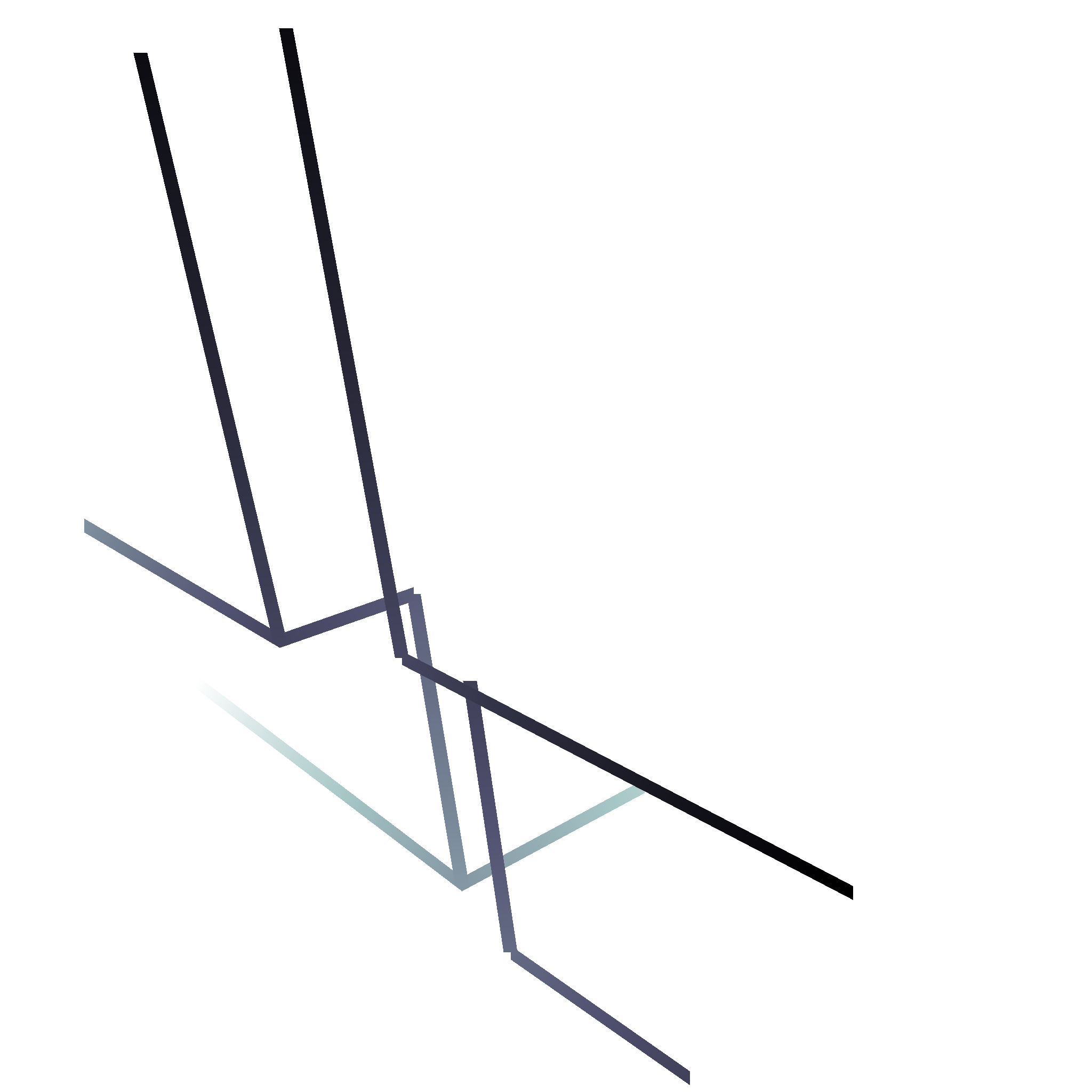}}\vspace{0.5mm}
    \frame{\includegraphics[ width=\linewidth]{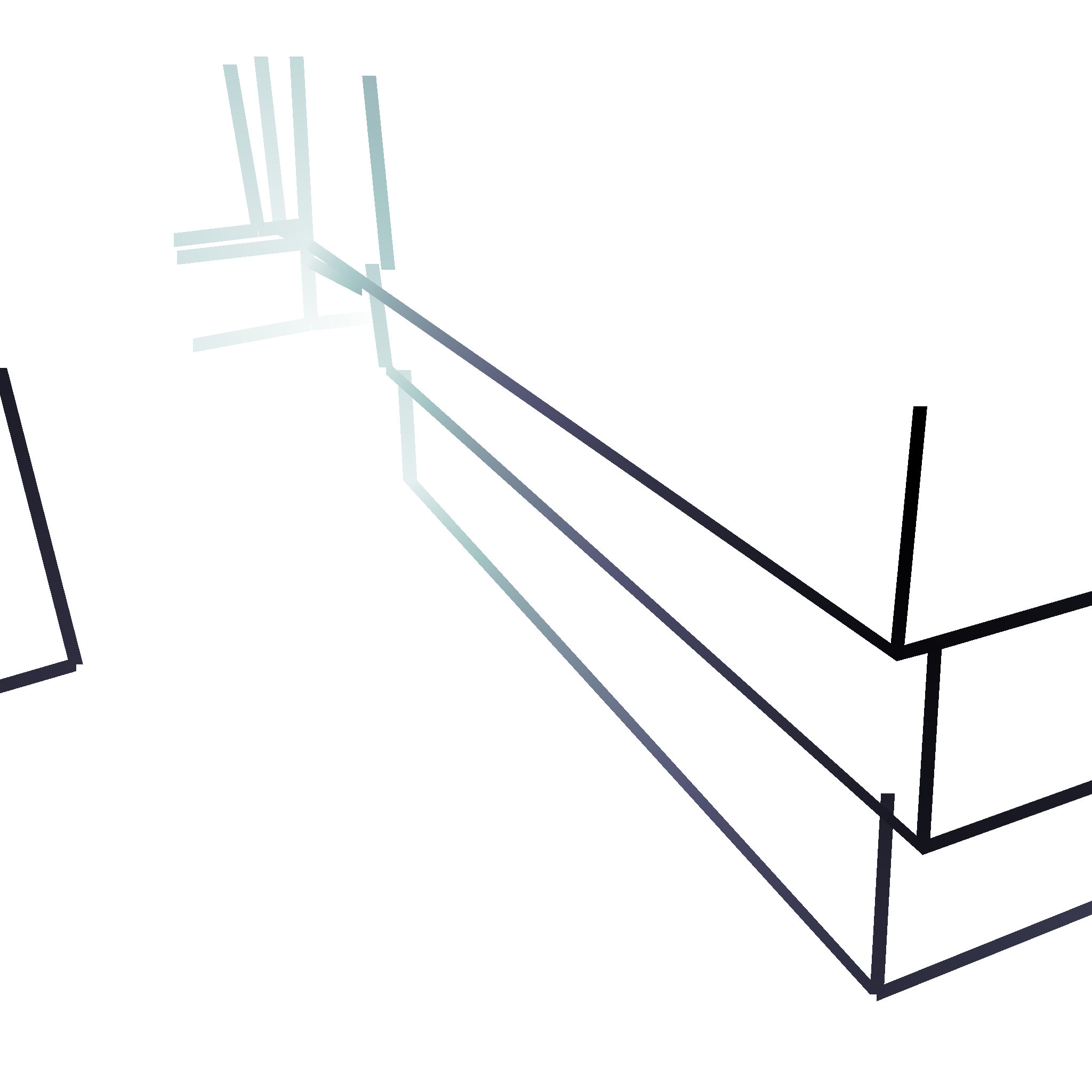}}\vspace{0.5mm}
    \frame{\includegraphics[ width=\linewidth]{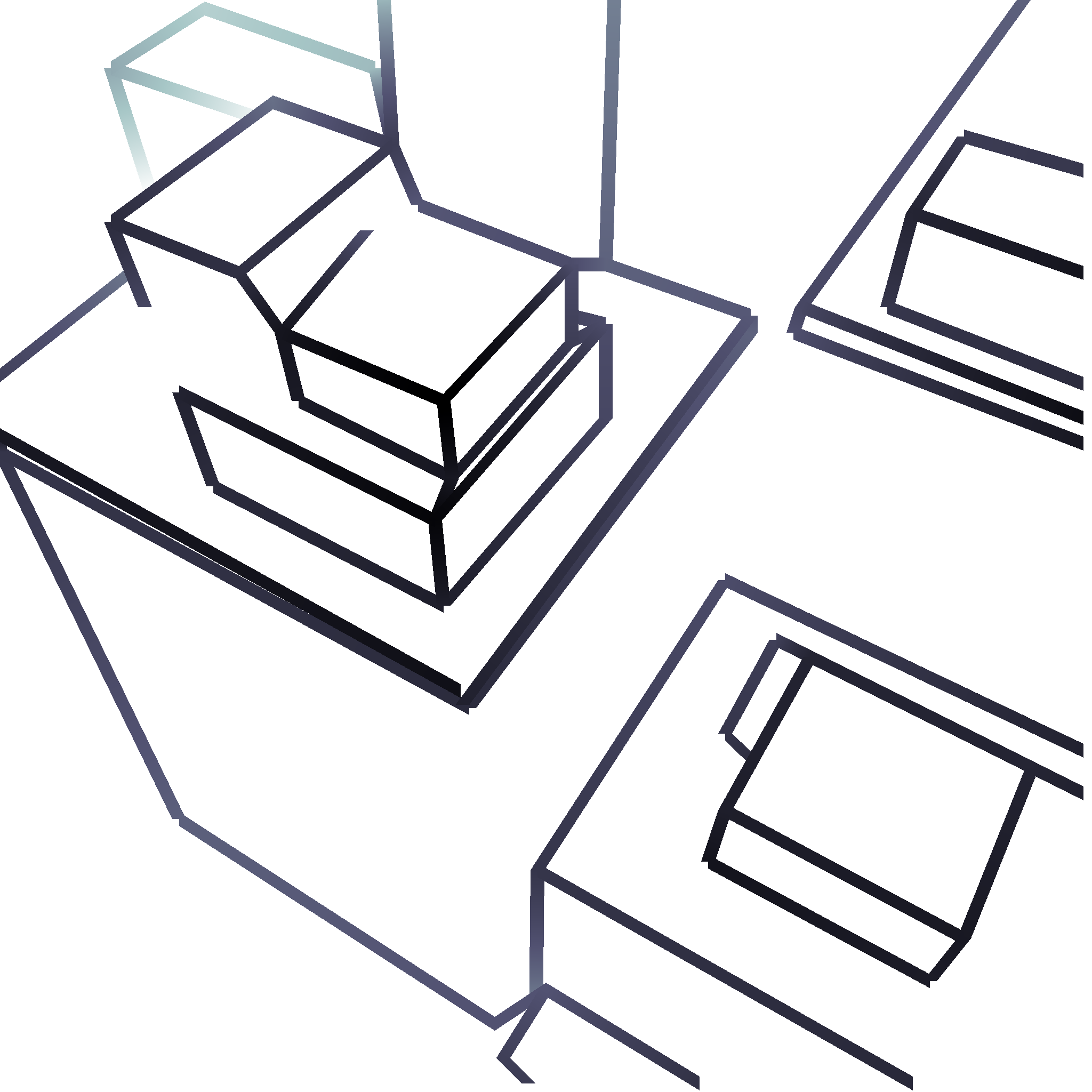}}
    \centering
    \small Inferred 3D\nothing{  wireframe}
    \end{minipage}
    \hfill
    \begin{minipage}[t]{0.135\linewidth}
    \frame{\includegraphics[width=\linewidth]{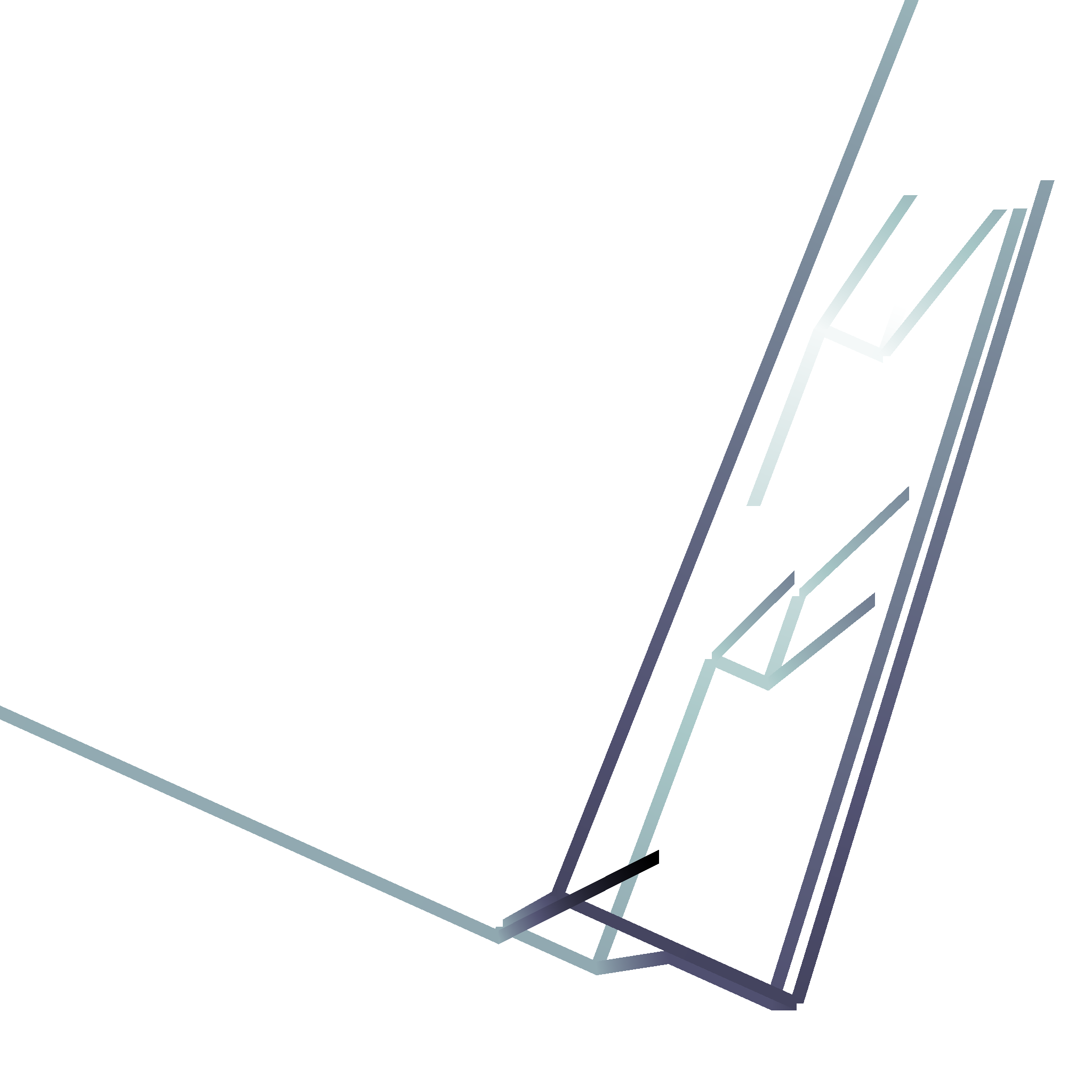}}\vspace{0.5mm}
    \frame{\includegraphics[width=\linewidth]{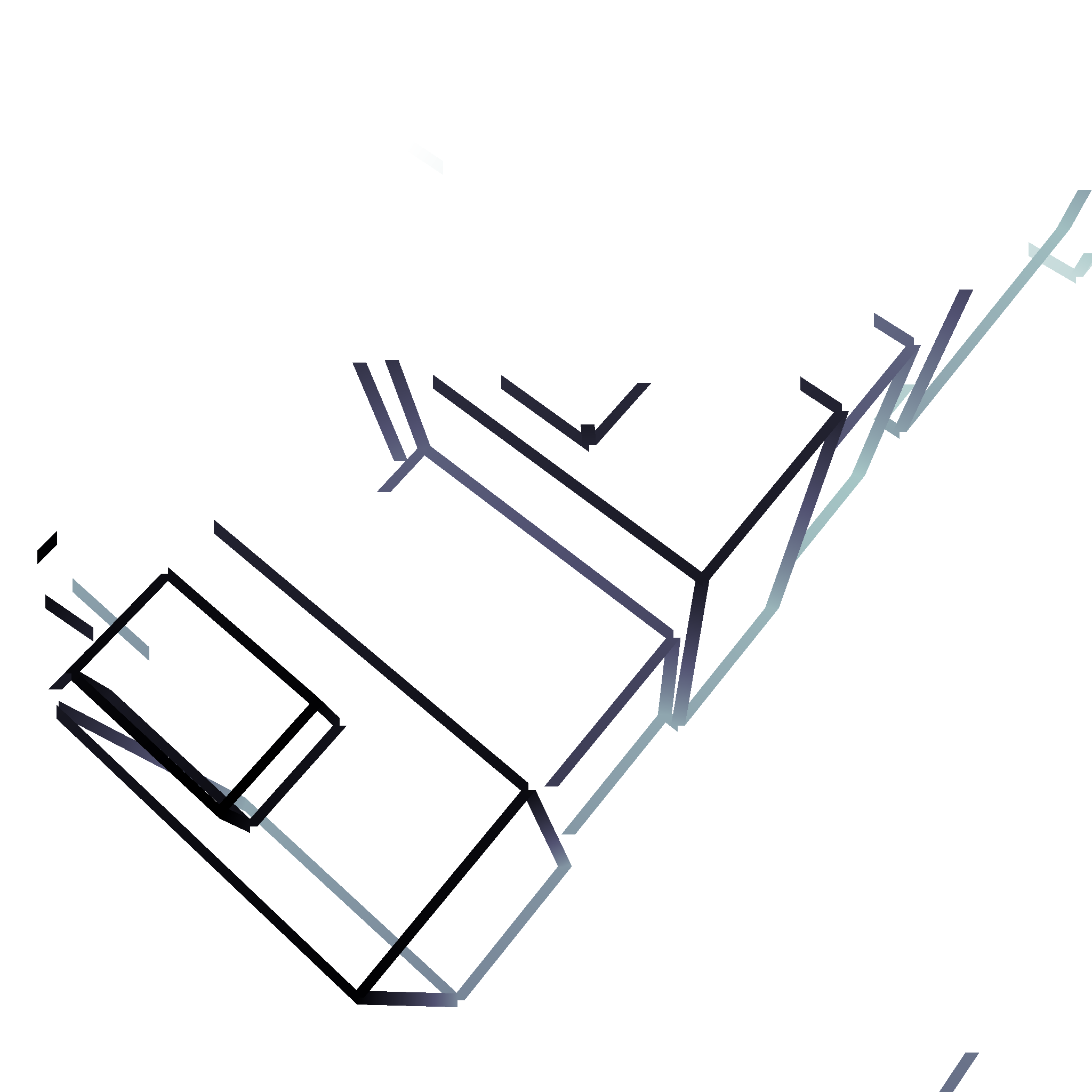}}\vspace{0.5mm}
    \frame{\includegraphics[width=\linewidth]{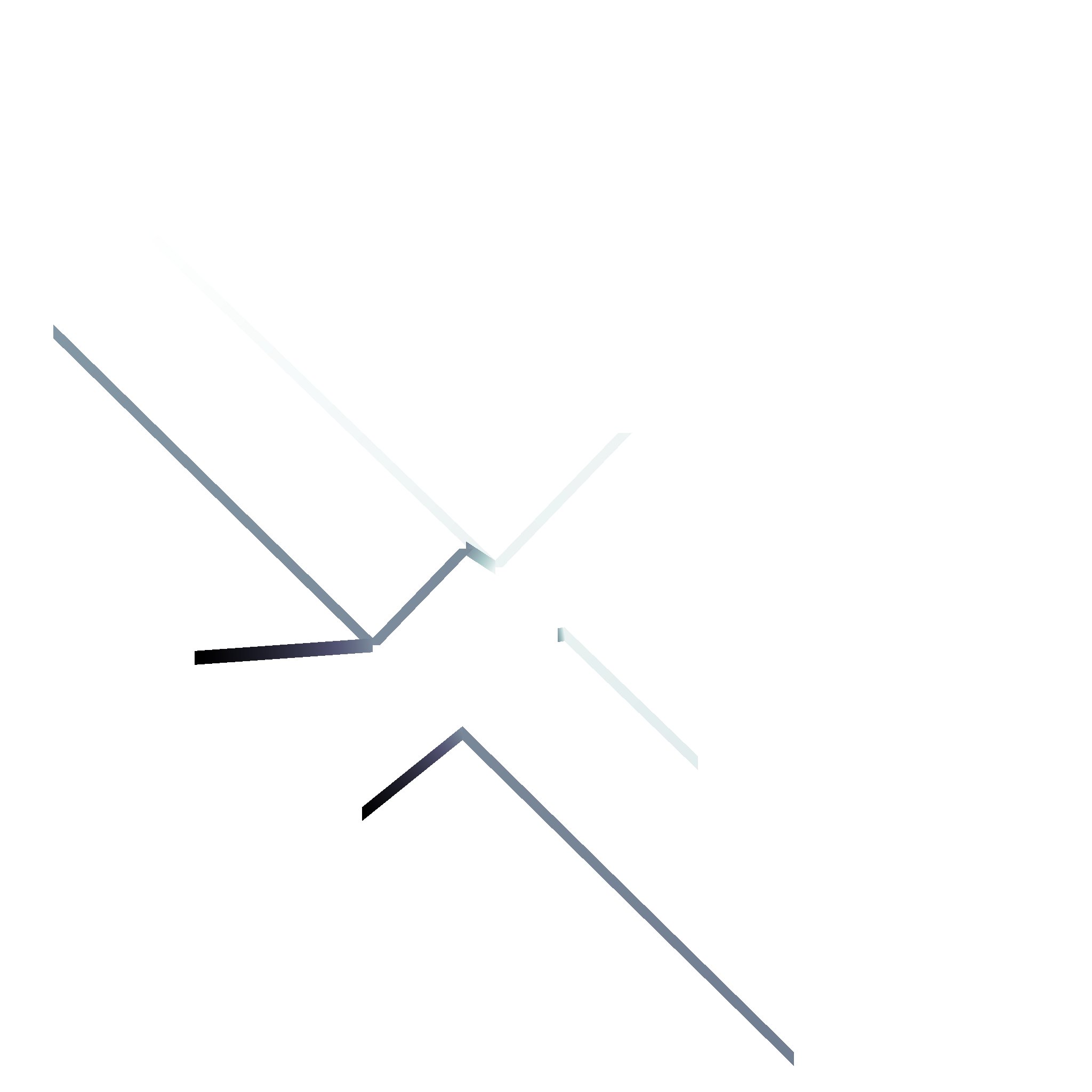}}\vspace{0.5mm}
    \frame{\includegraphics[width=\linewidth]{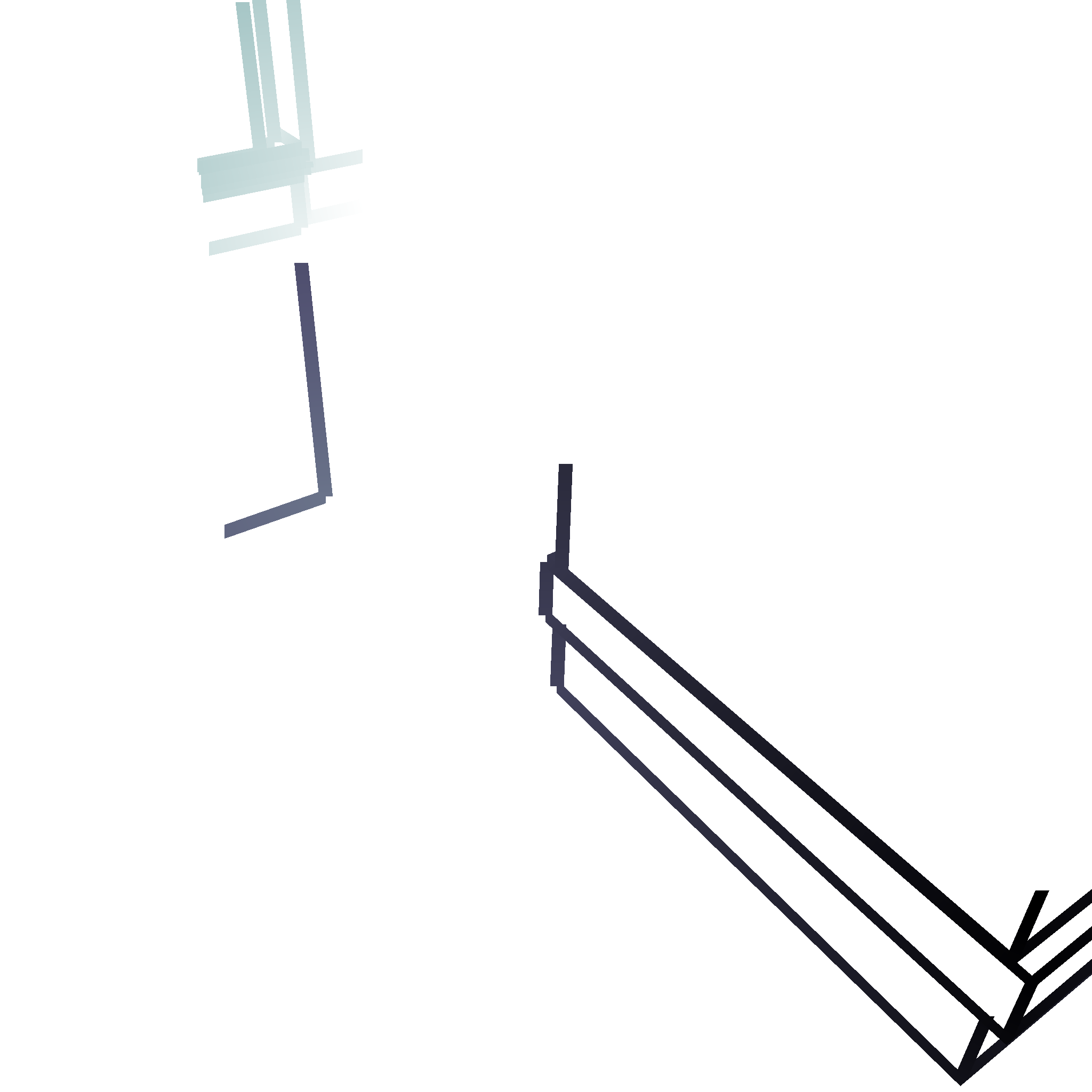}}\vspace{0.5mm}
    \frame{\includegraphics[width=\linewidth]{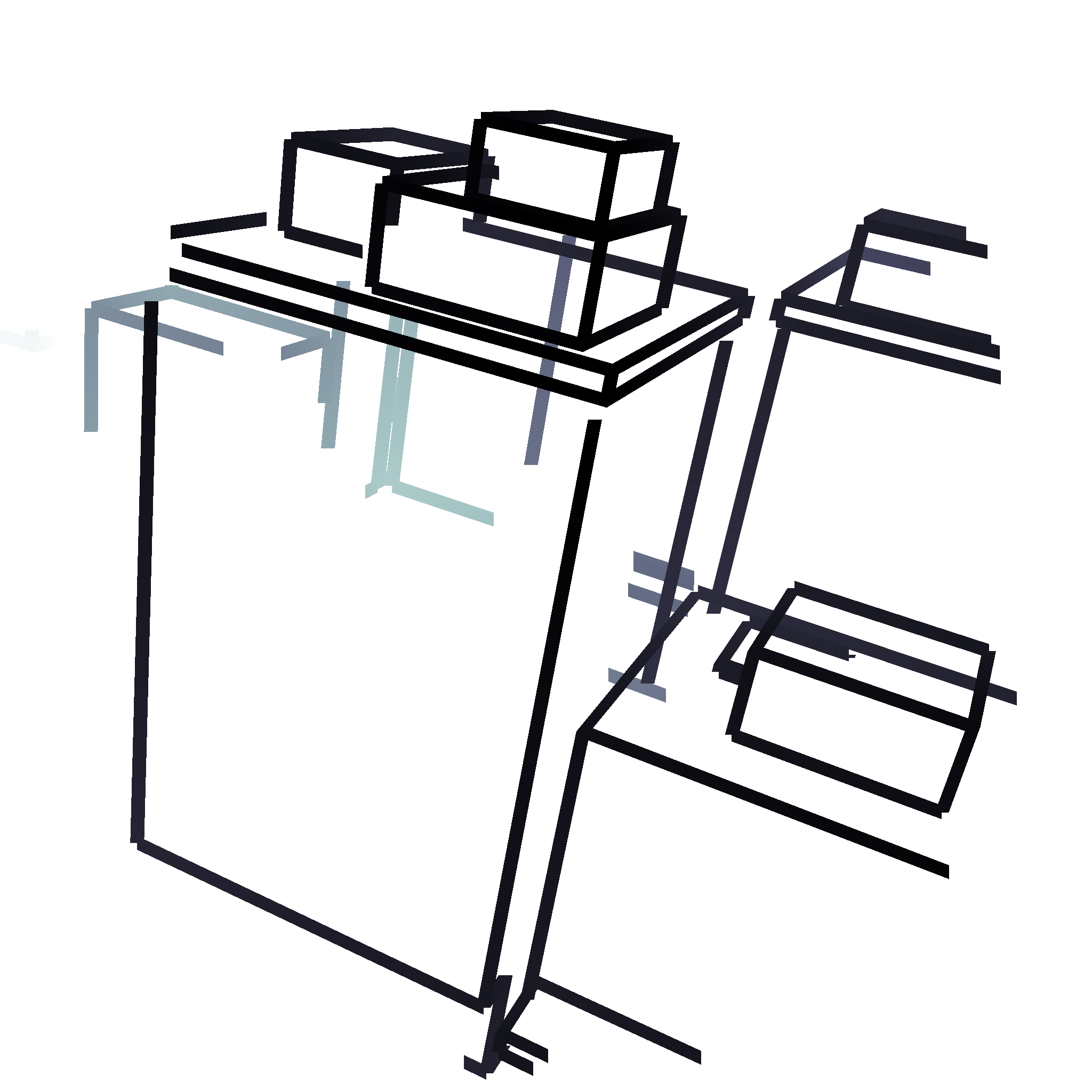}}
    \centering
    \small Ground truth 3D\nothing{  wireframe}
    \end{minipage}
    \begin{minipage}[t]{0.135\linewidth}
    \frame{\includegraphics[width=\linewidth]{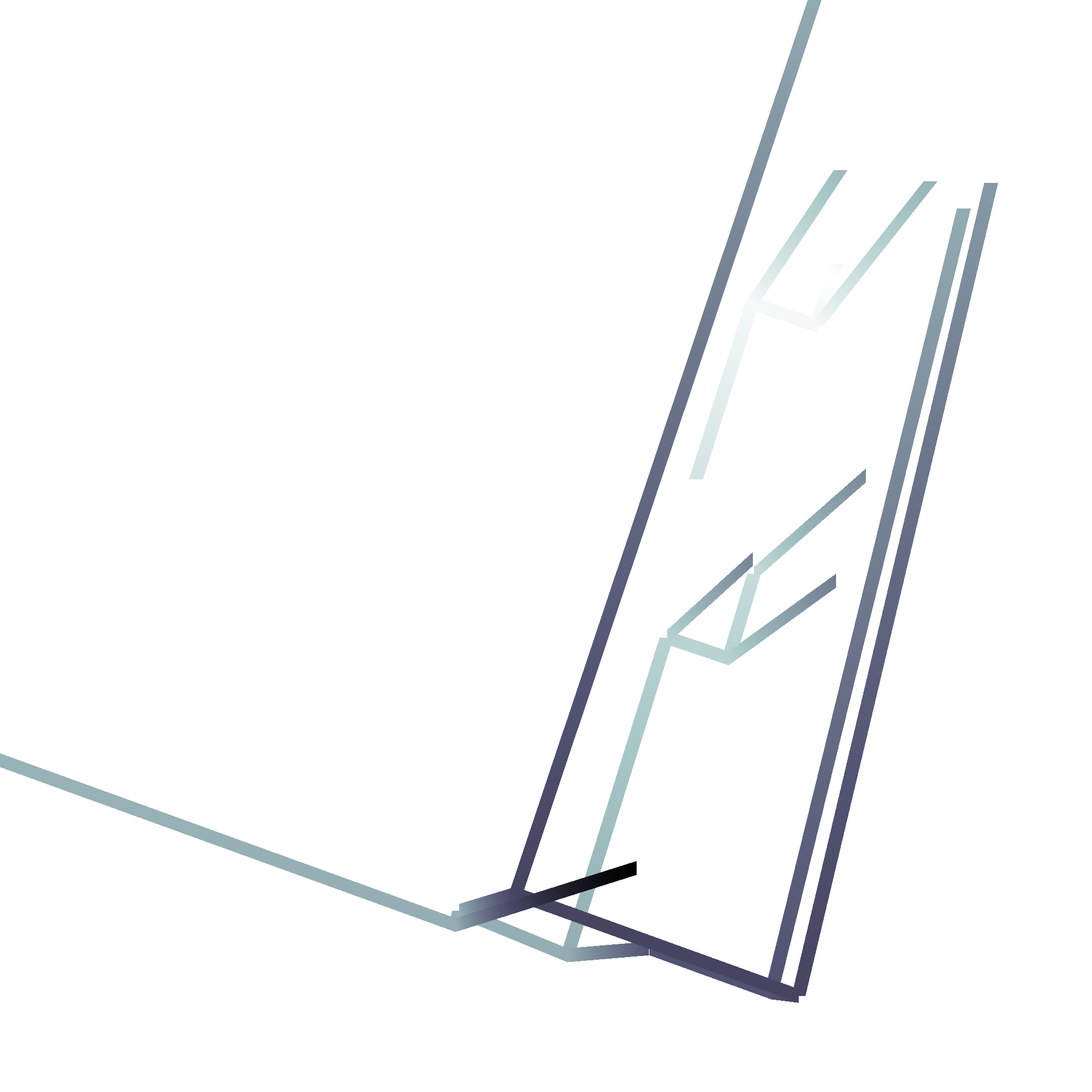}}\vspace{0.5mm}
    \frame{\includegraphics[width=\linewidth]{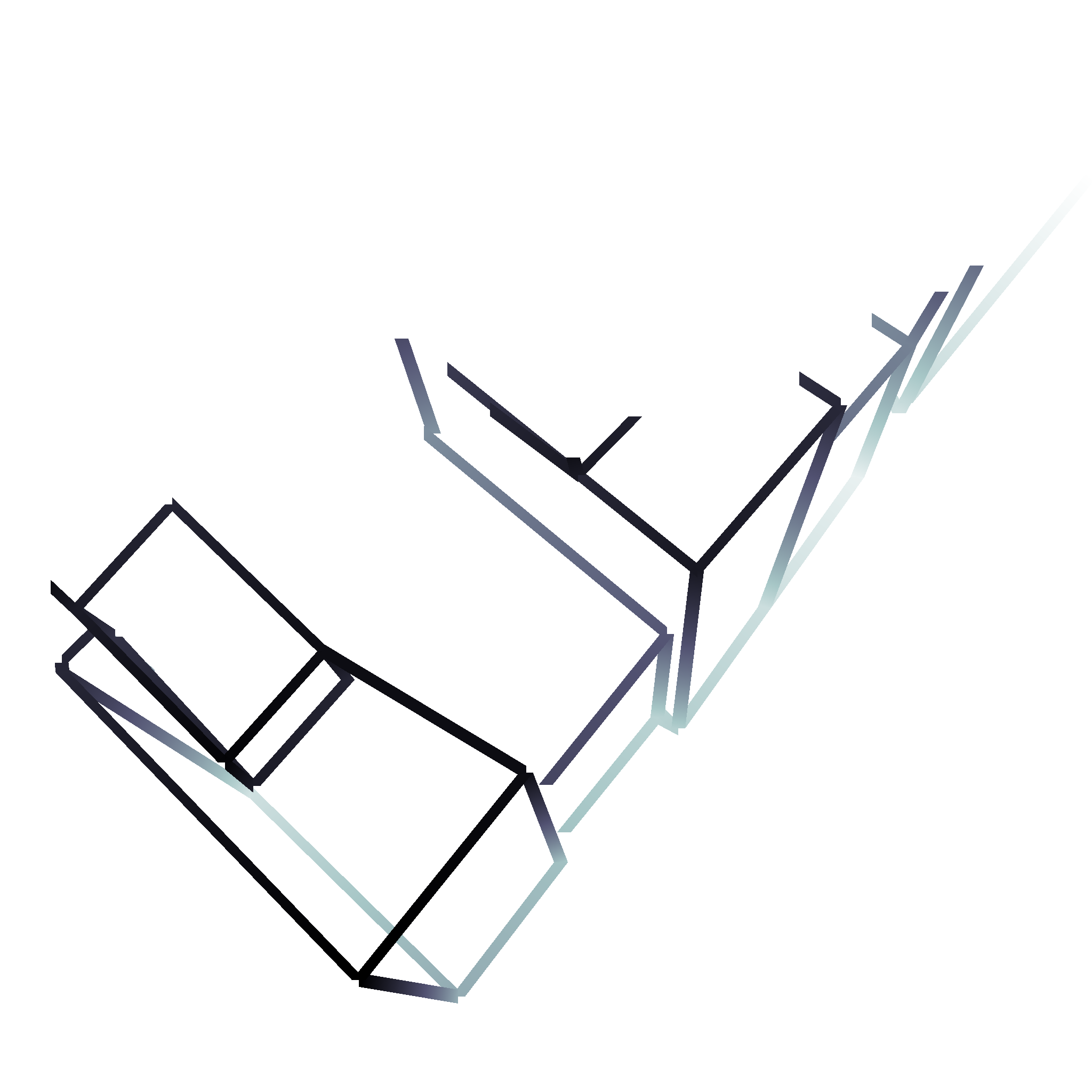}}\vspace{0.5mm}
    \frame{\includegraphics[width=\linewidth]{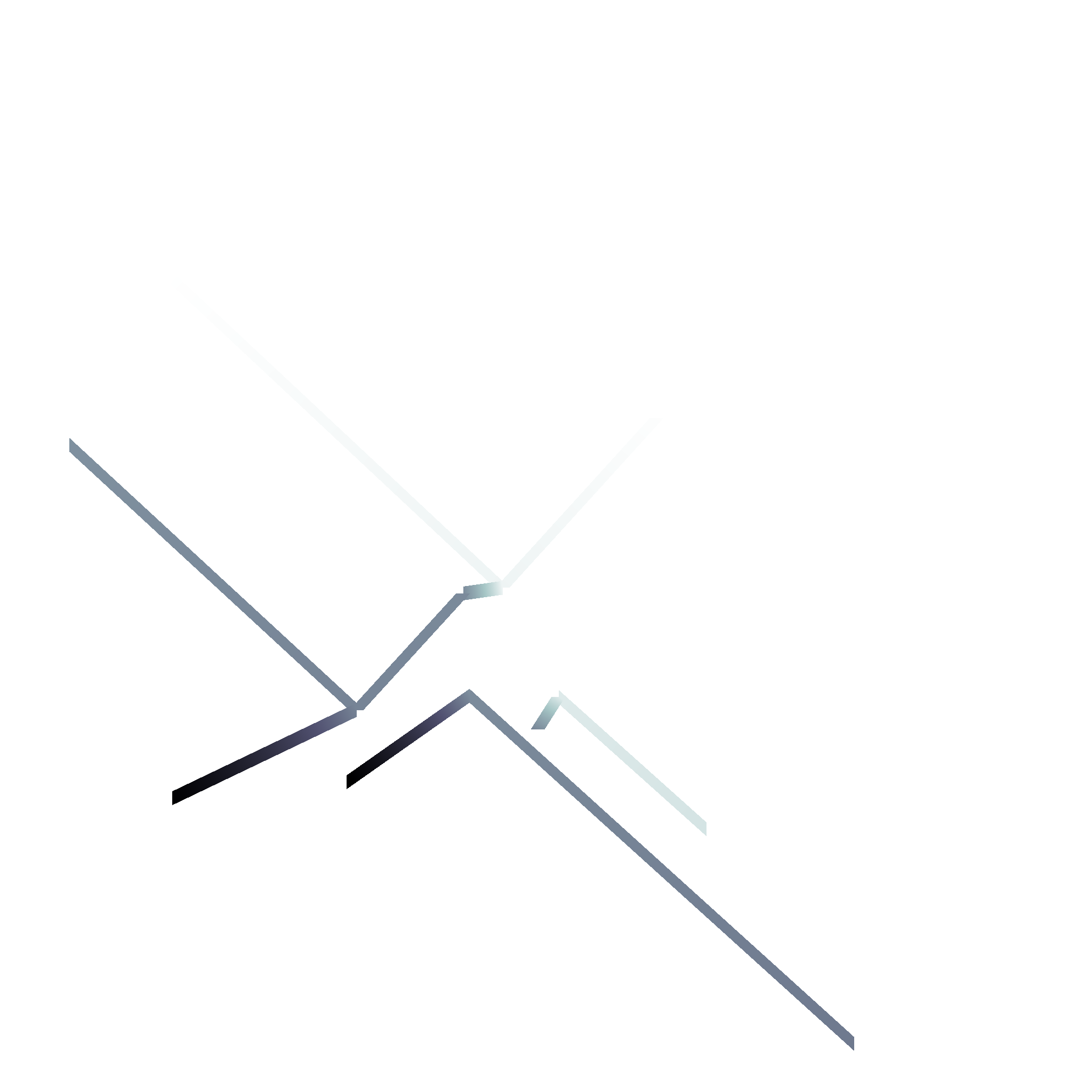}}\vspace{0.5mm}
    \frame{\includegraphics[width=\linewidth]{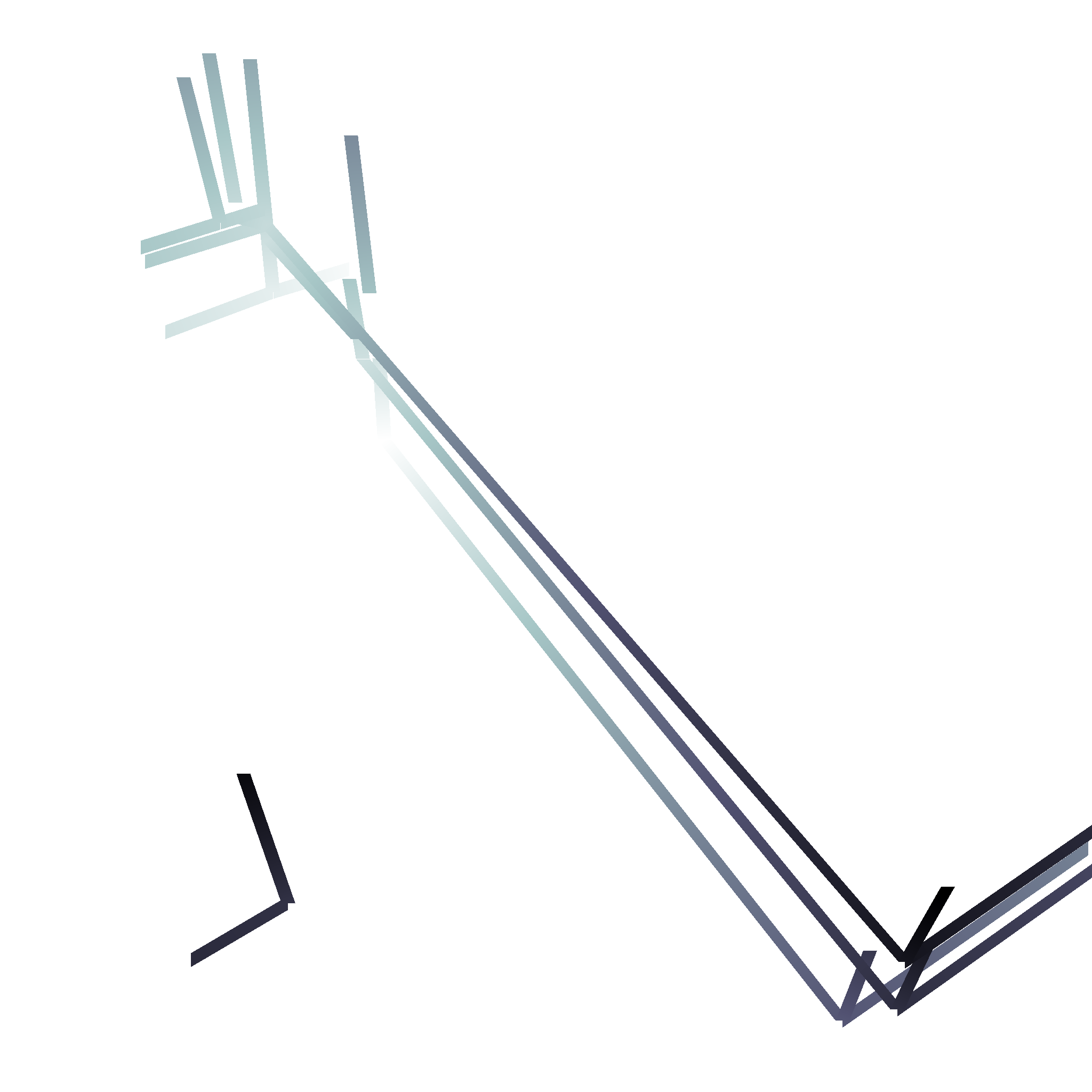}}\vspace{0.5mm}
    \frame{\includegraphics[width=\linewidth]{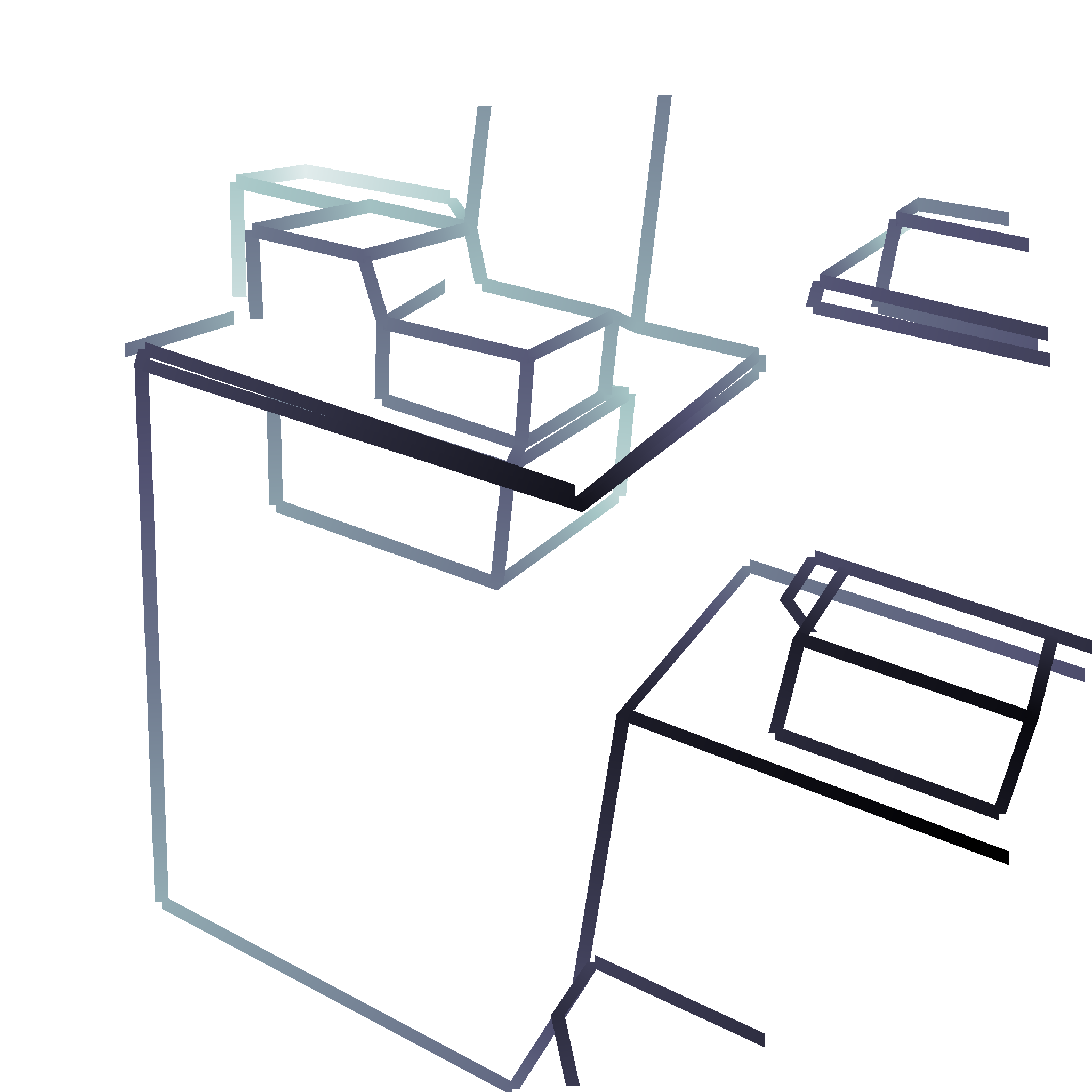}}
    \small Inferred 3D\nothing{  wireframe}
    \centering
    \end{minipage}
\caption[\small Test results on the synthetic SceneCity image dataset.] 
{%
\small Left group: comparison of 2D results between the ground truth (column 1), our predictions (column 2), and \N{the results from wireframe parser \cite{Huang:2018:LPW} (column 3)}.
Middle (columns 4-5) and right groups (columns 6-7): novel views of the ground truths and our reconstructions to demonstrate the 3D representation of the scene.
The color of the wireframes visualizes depth.
}
\label{fig:wireframe-synthetic}
\end{figure*}

\begin{figure*}[p]
\centering
\begin{minipage}[t]{0.155\linewidth}
    \centering
    \frame{\includegraphics[width=\linewidth]{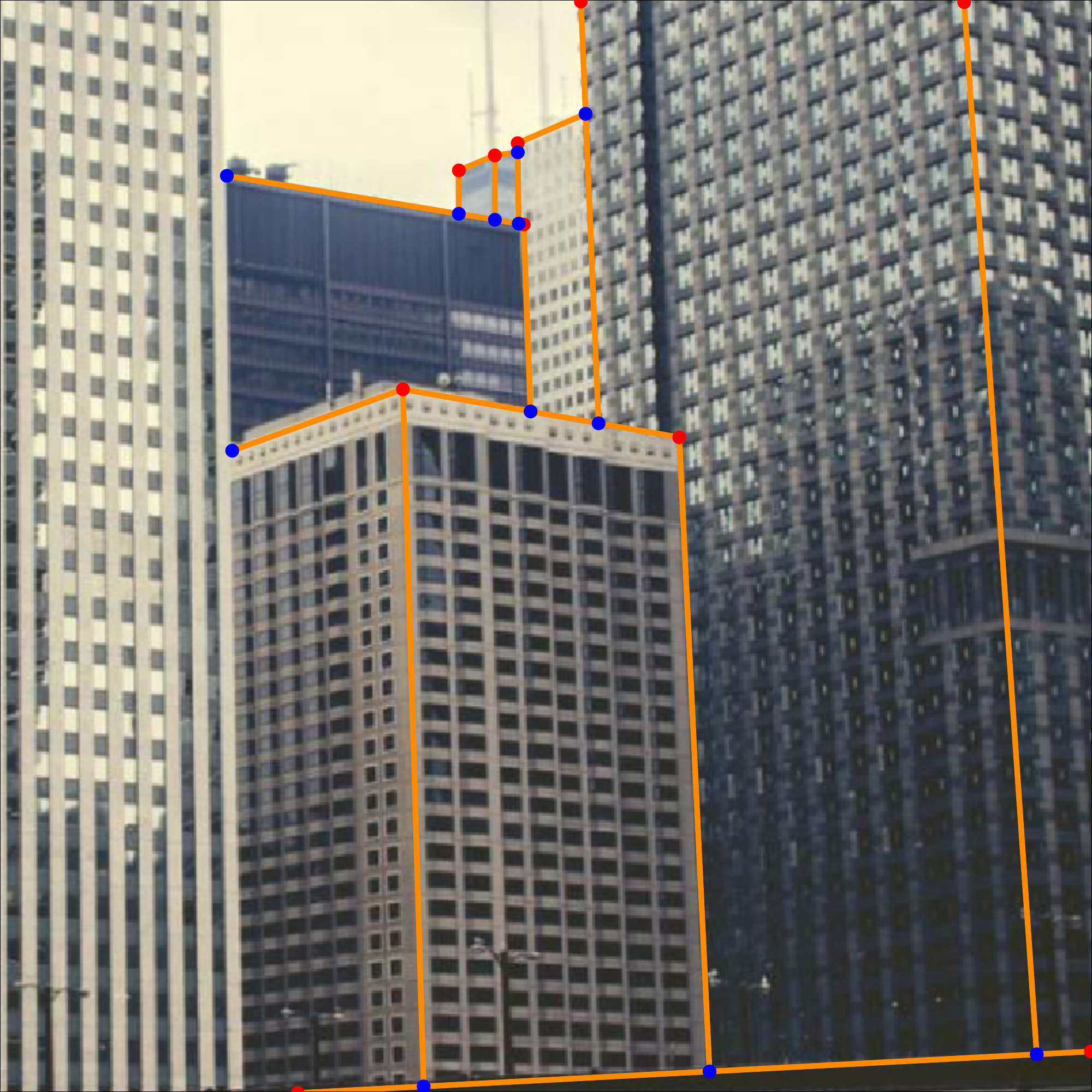}}\vspace{0.5mm}
    \frame{\includegraphics[width=\linewidth]{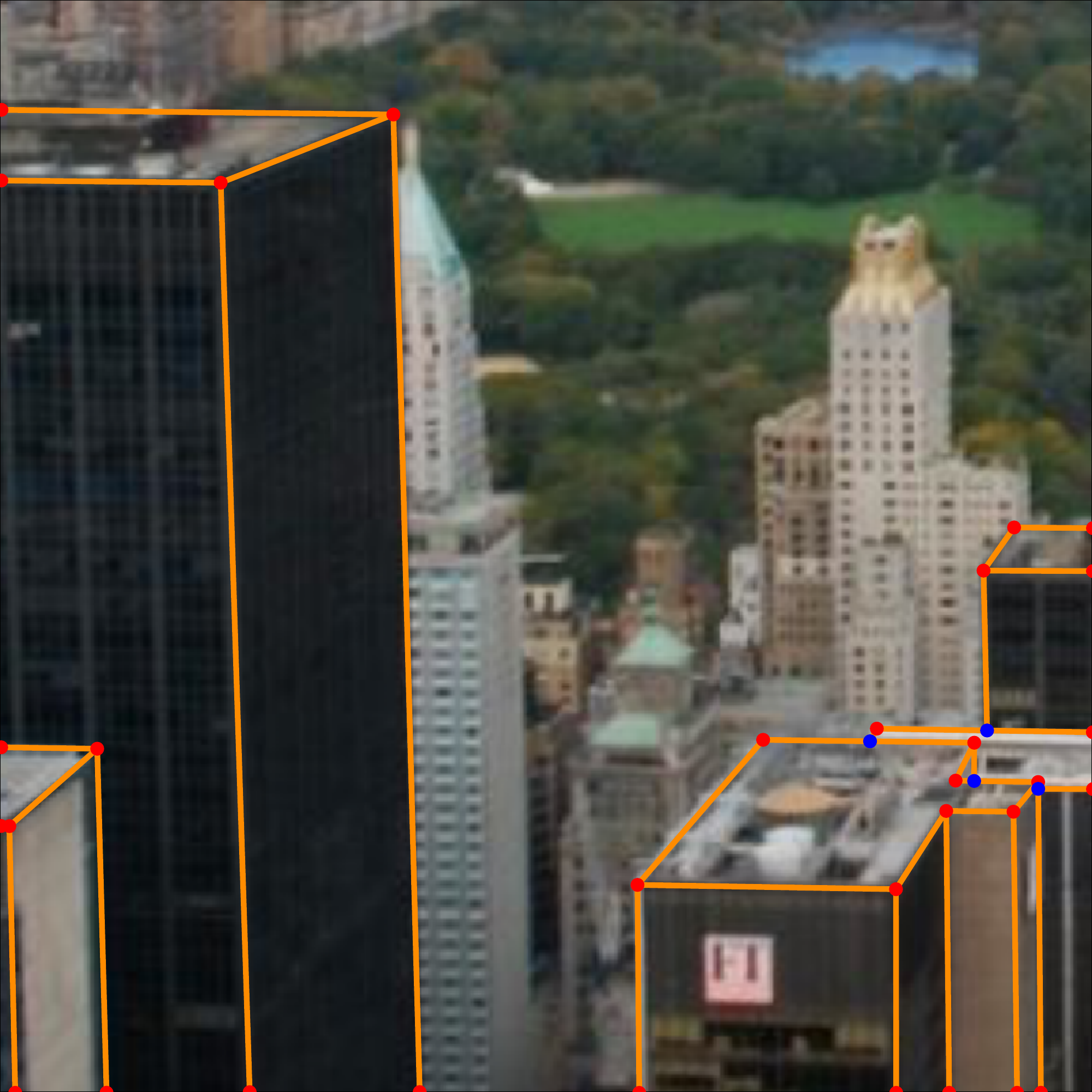}}
    \small Ground truth
    \end{minipage}
    \begin{minipage}[t]{0.155\linewidth}
    \frame{\includegraphics[width=\linewidth]{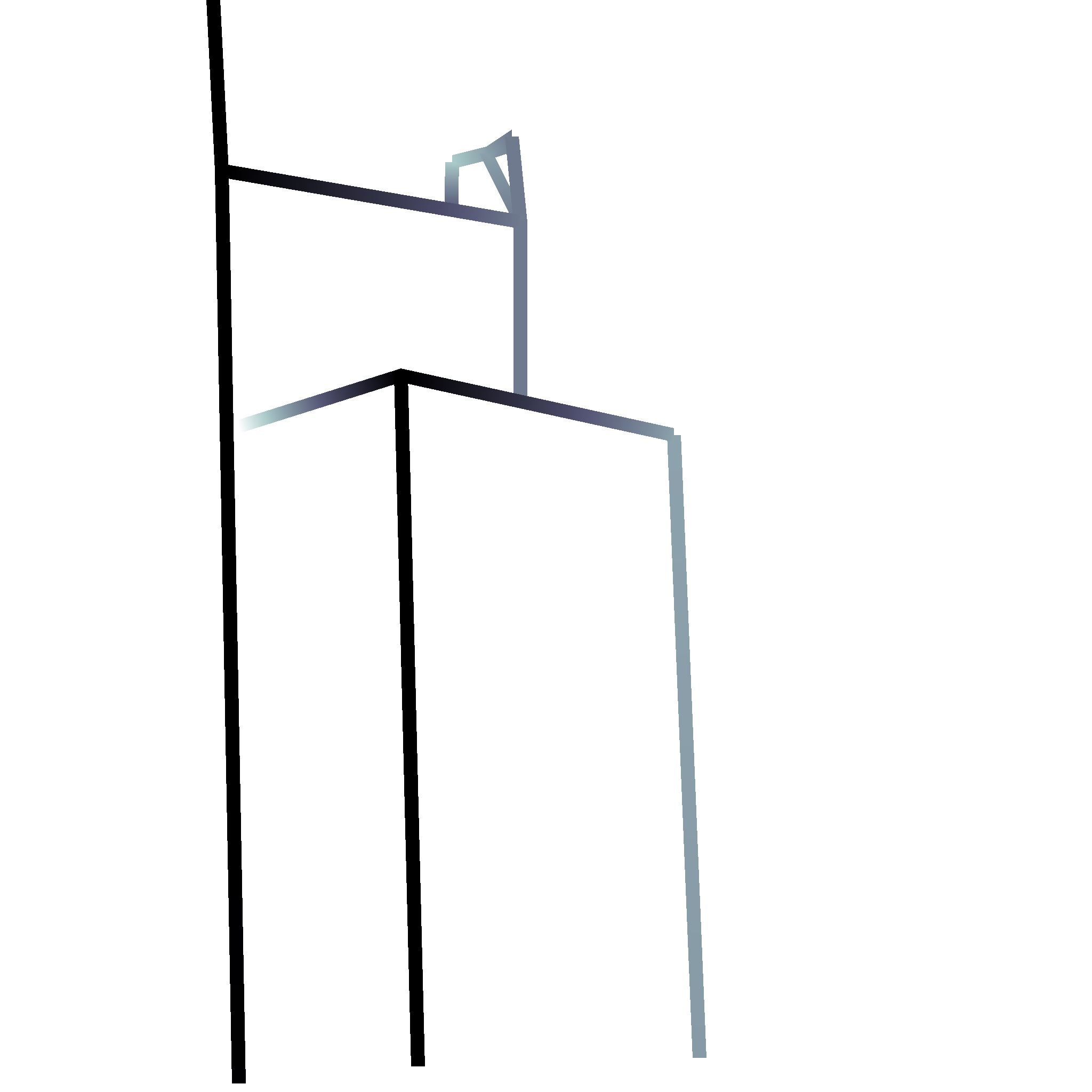}}\vspace{0.5mm}
    \frame{\includegraphics[width=\linewidth]{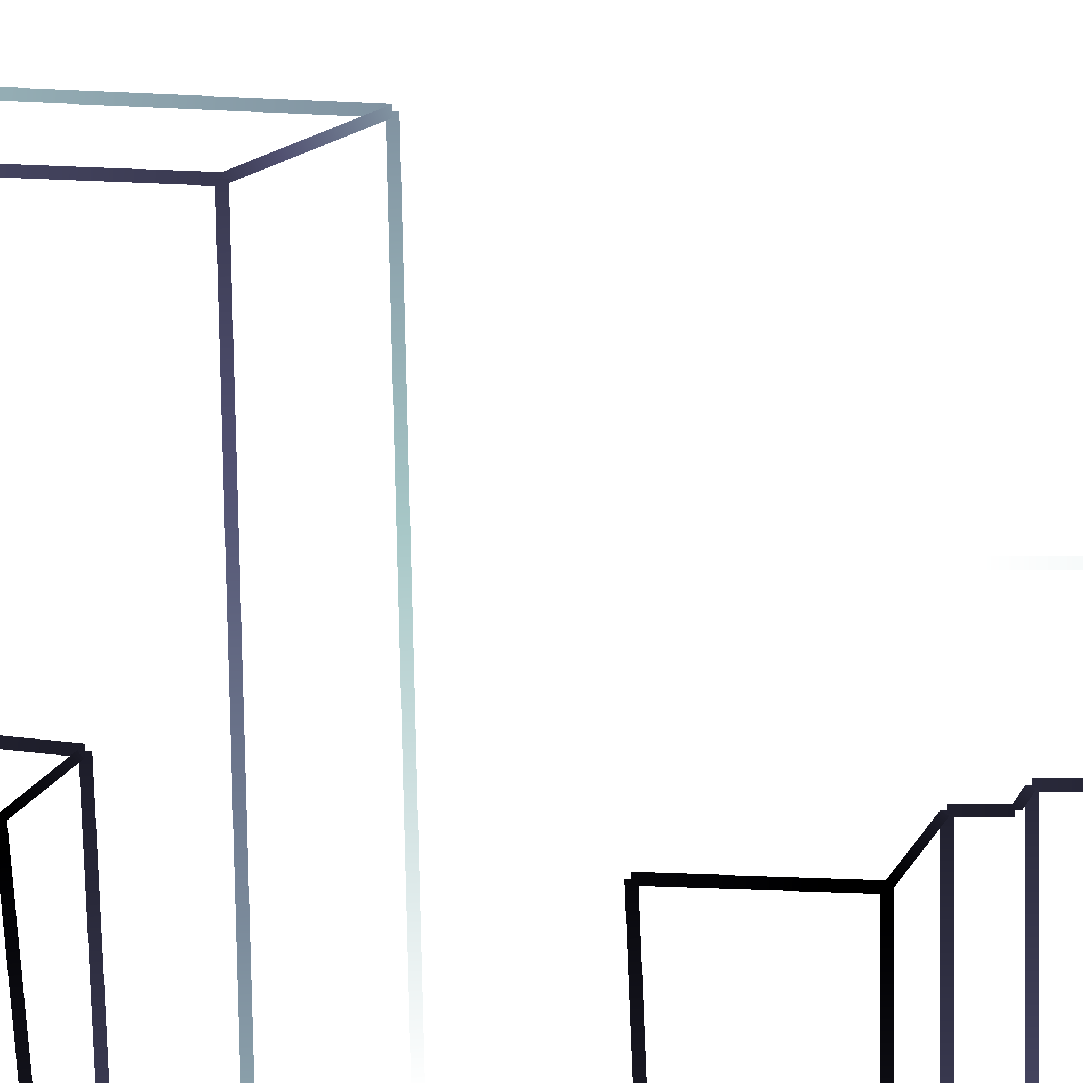}}
    \centering
    \small Inferred
    \end{minipage}
    \begin{minipage}[t]{0.155\linewidth}
    \centering
    \frame{\includegraphics[width=\linewidth]{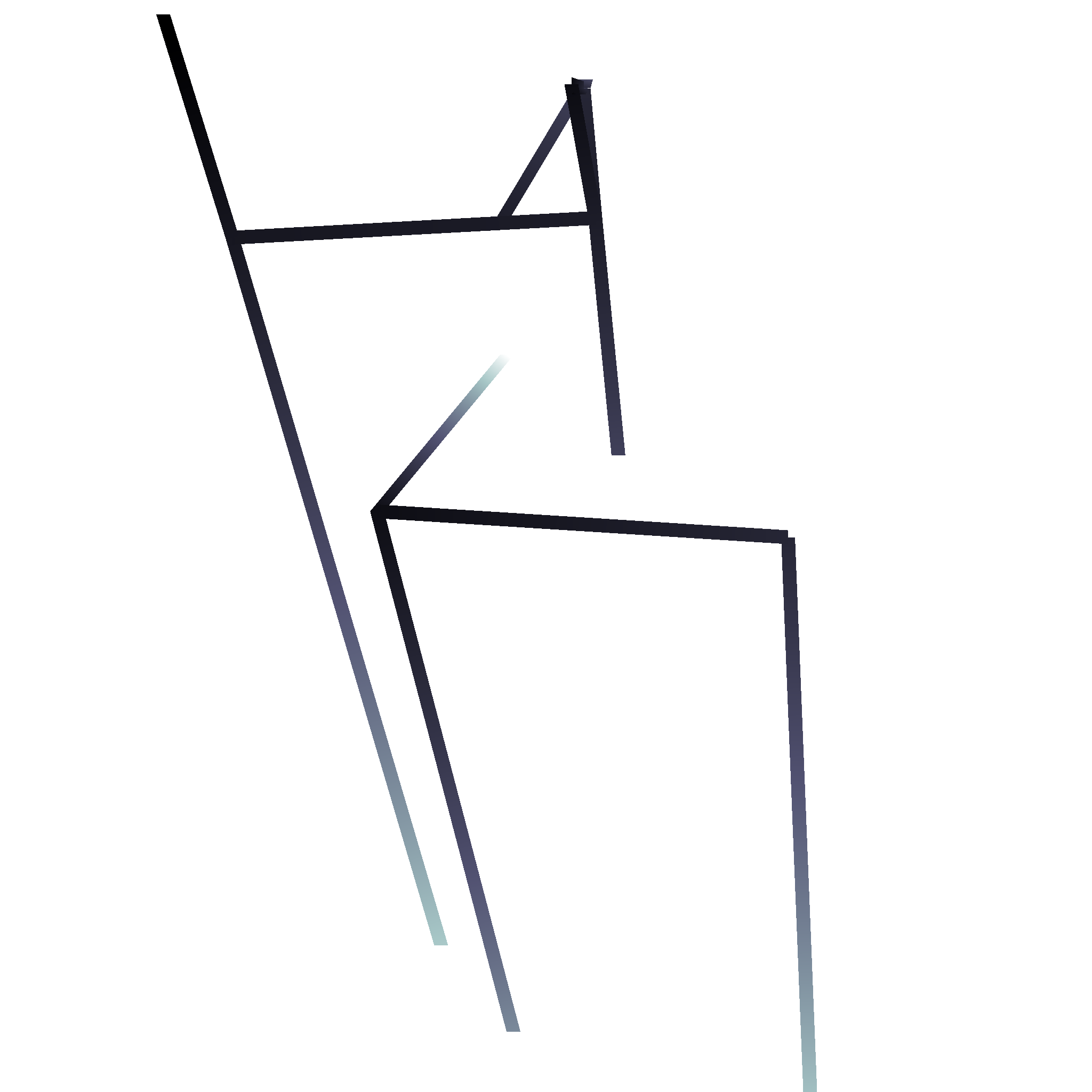}}\vspace{0.5mm}
    \frame{\includegraphics[width=\linewidth]{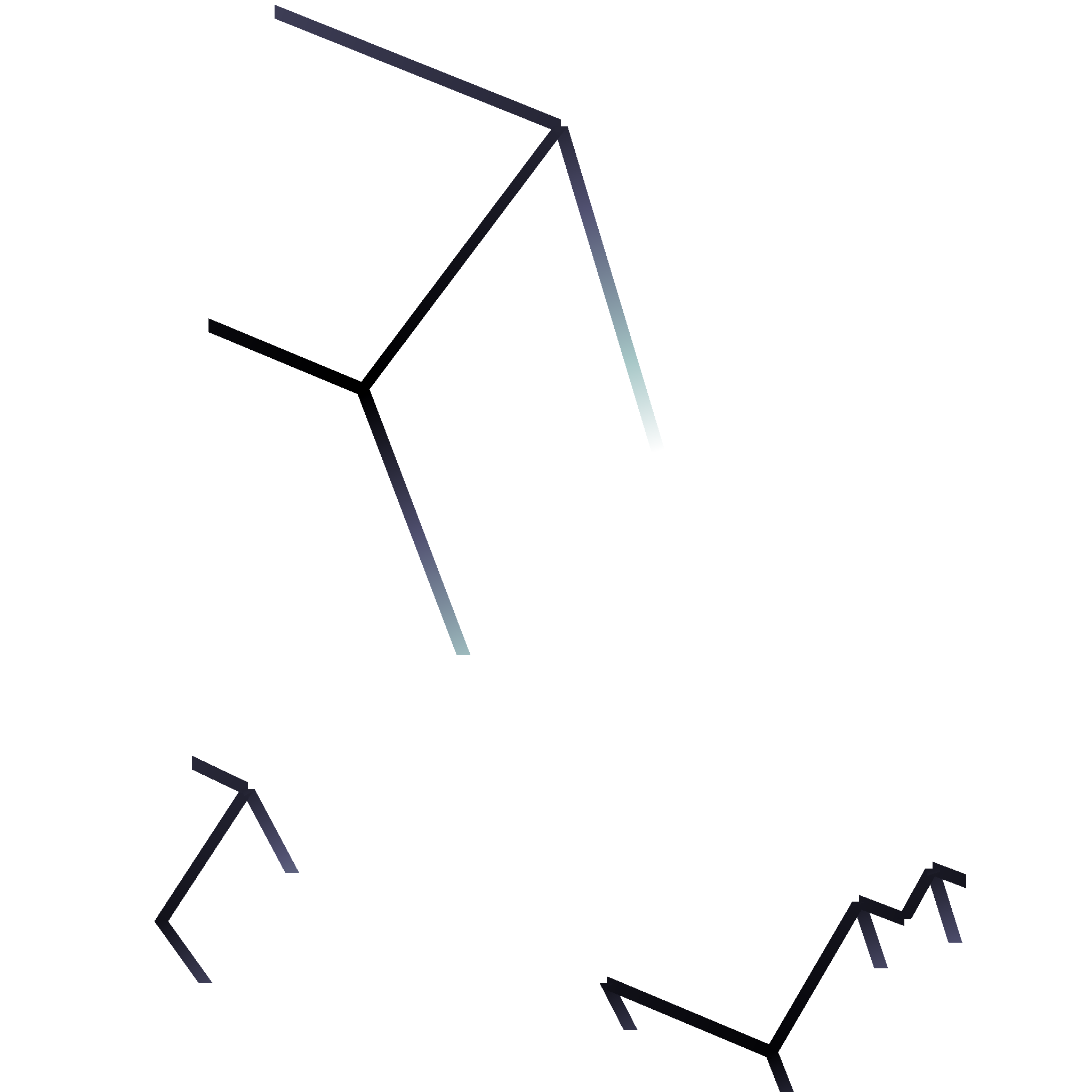}}
    \small Novel views
    \end{minipage}
    \hfill
    \begin{minipage}[t]{0.155\linewidth}
    \centering
    \frame{\includegraphics[width=\linewidth]{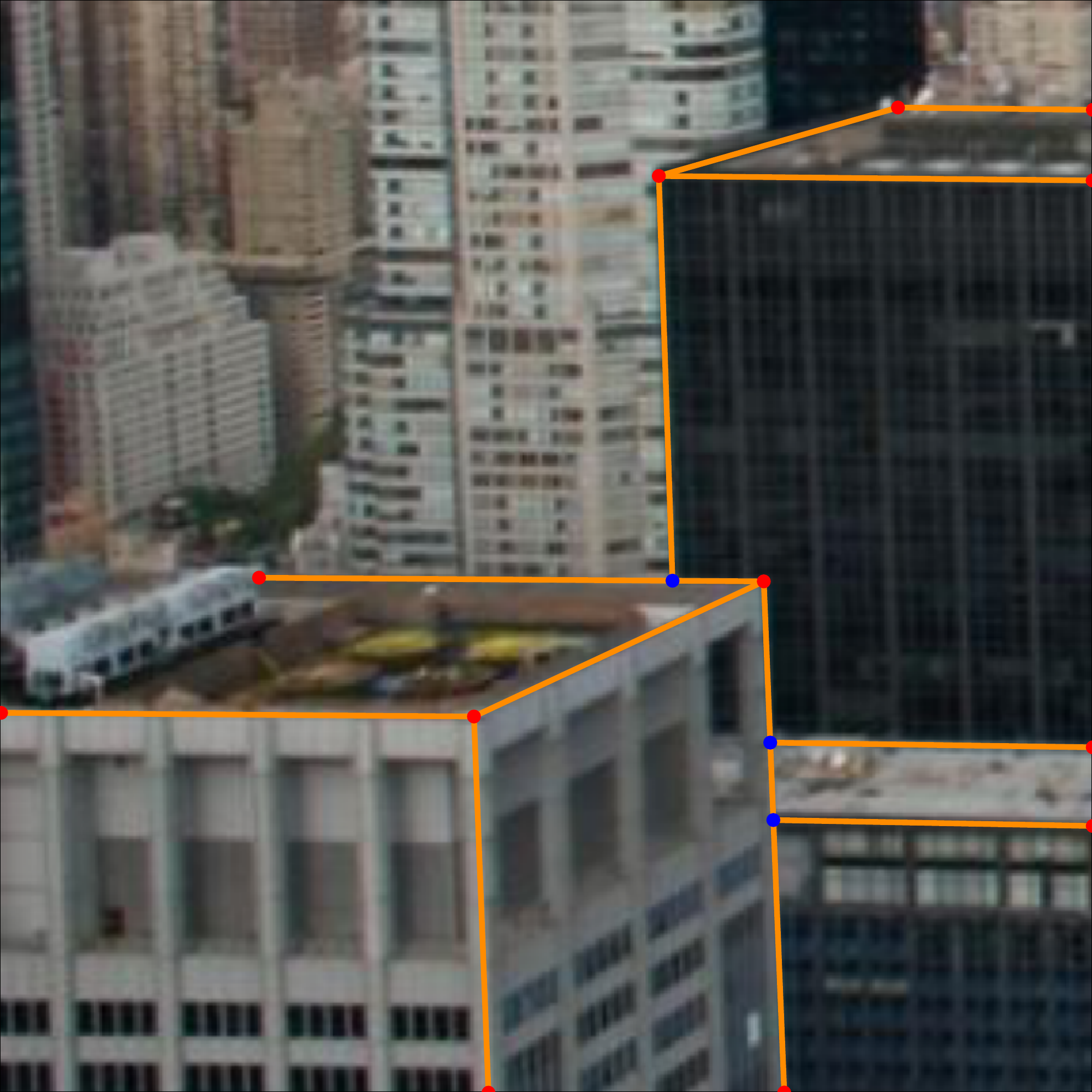}}\vspace{0.5mm}
    \frame{\includegraphics[width=\linewidth]{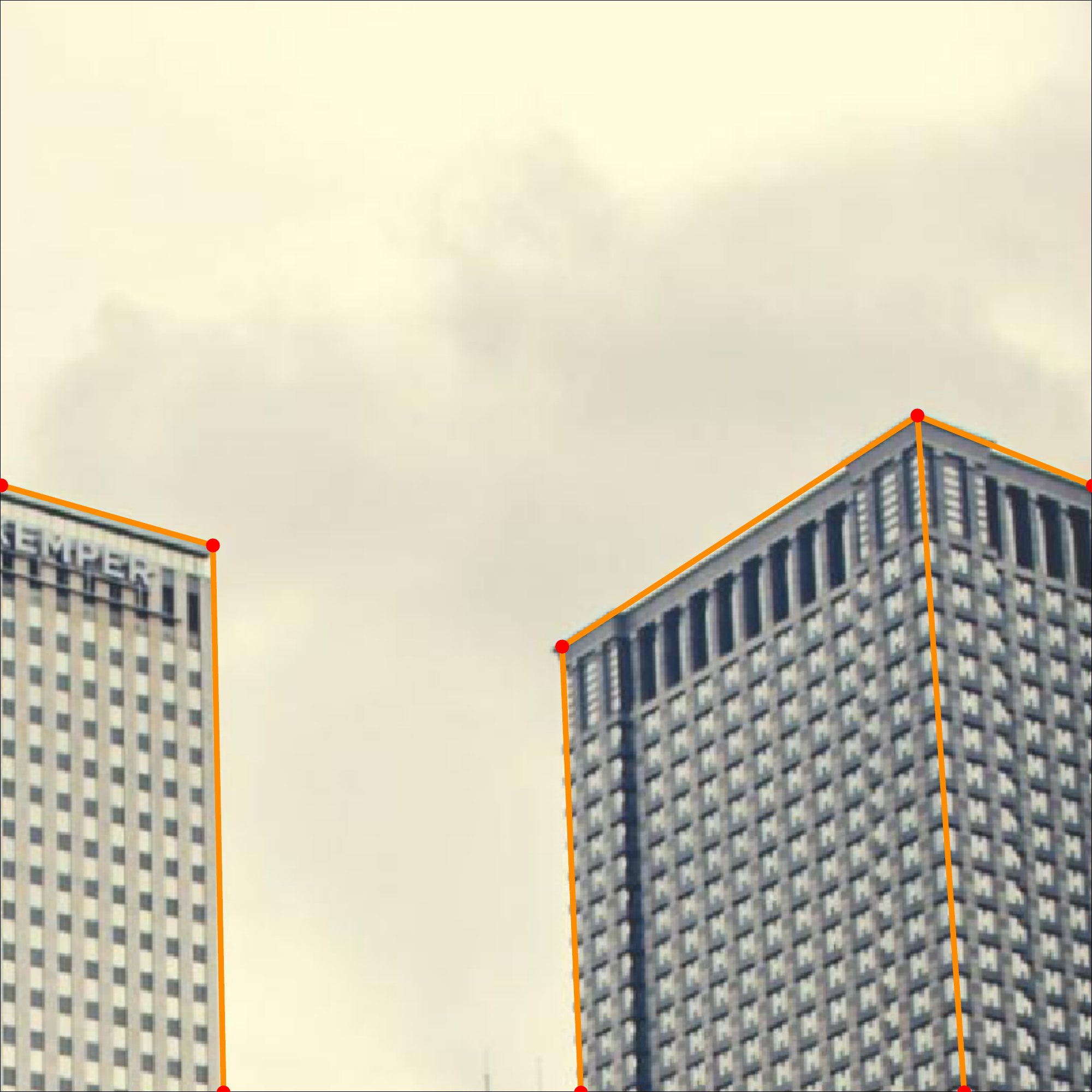}}
    \small Ground truth
    \end{minipage}
    \begin{minipage}[t]{0.155\linewidth}
    \frame{\includegraphics[width=\linewidth]{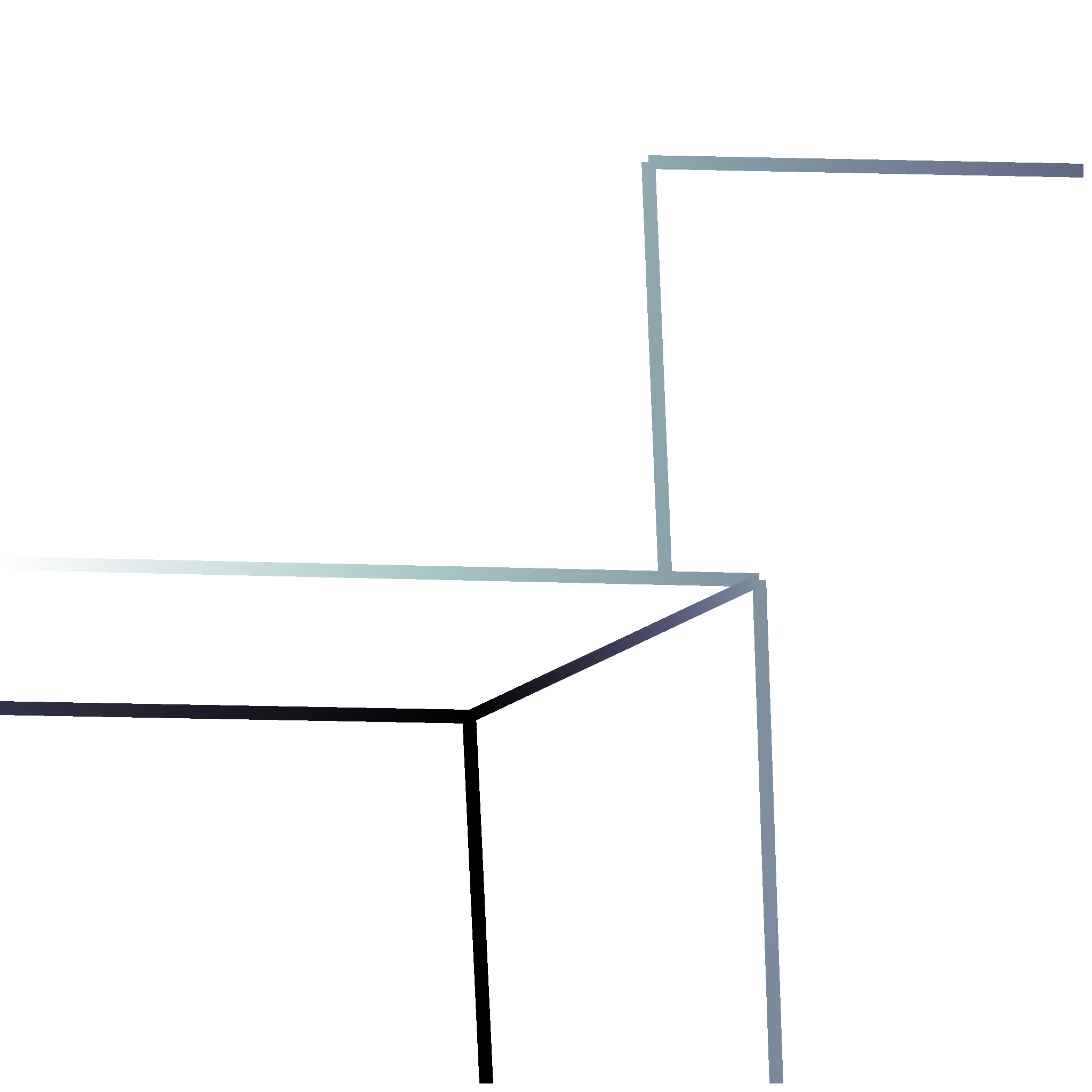}}\vspace{0.5mm}
    \frame{\includegraphics[width=\linewidth]{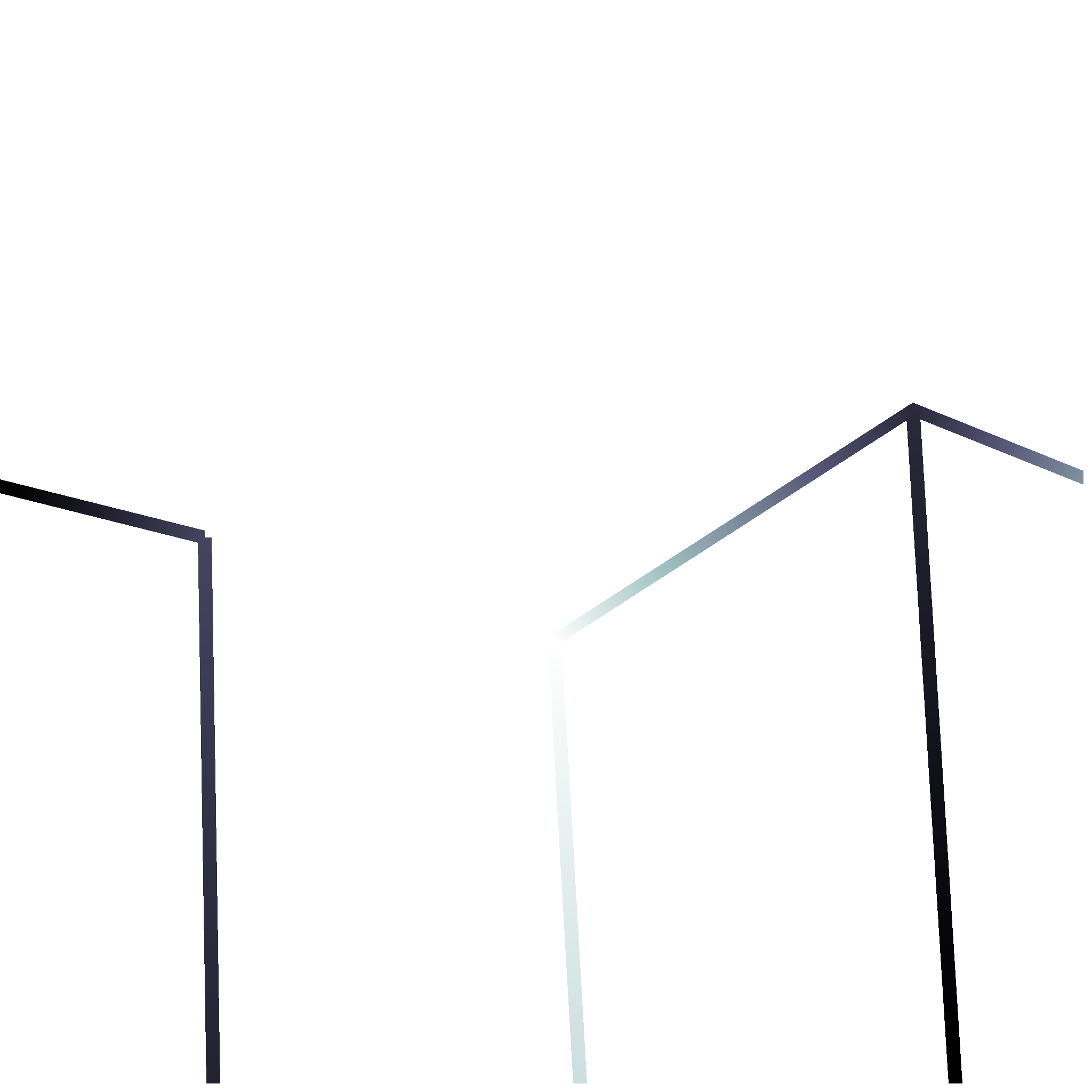}}
    \centering
    \small Inferred
    \end{minipage}
    \begin{minipage}[t]{0.155\linewidth}
    \centering
    \frame{\includegraphics[width=\linewidth]{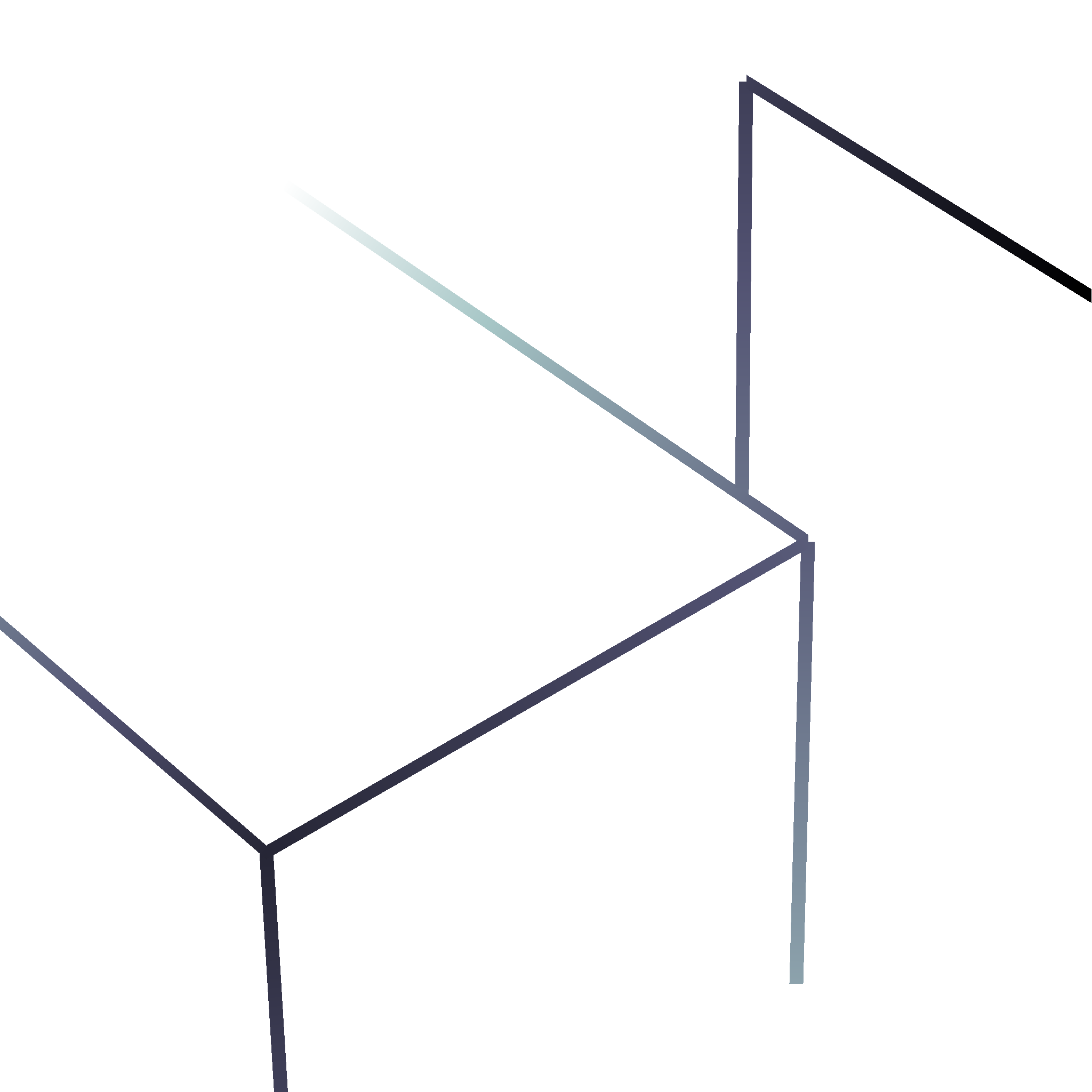}}\vspace{0.5mm}
    \frame{\includegraphics[width=\linewidth]{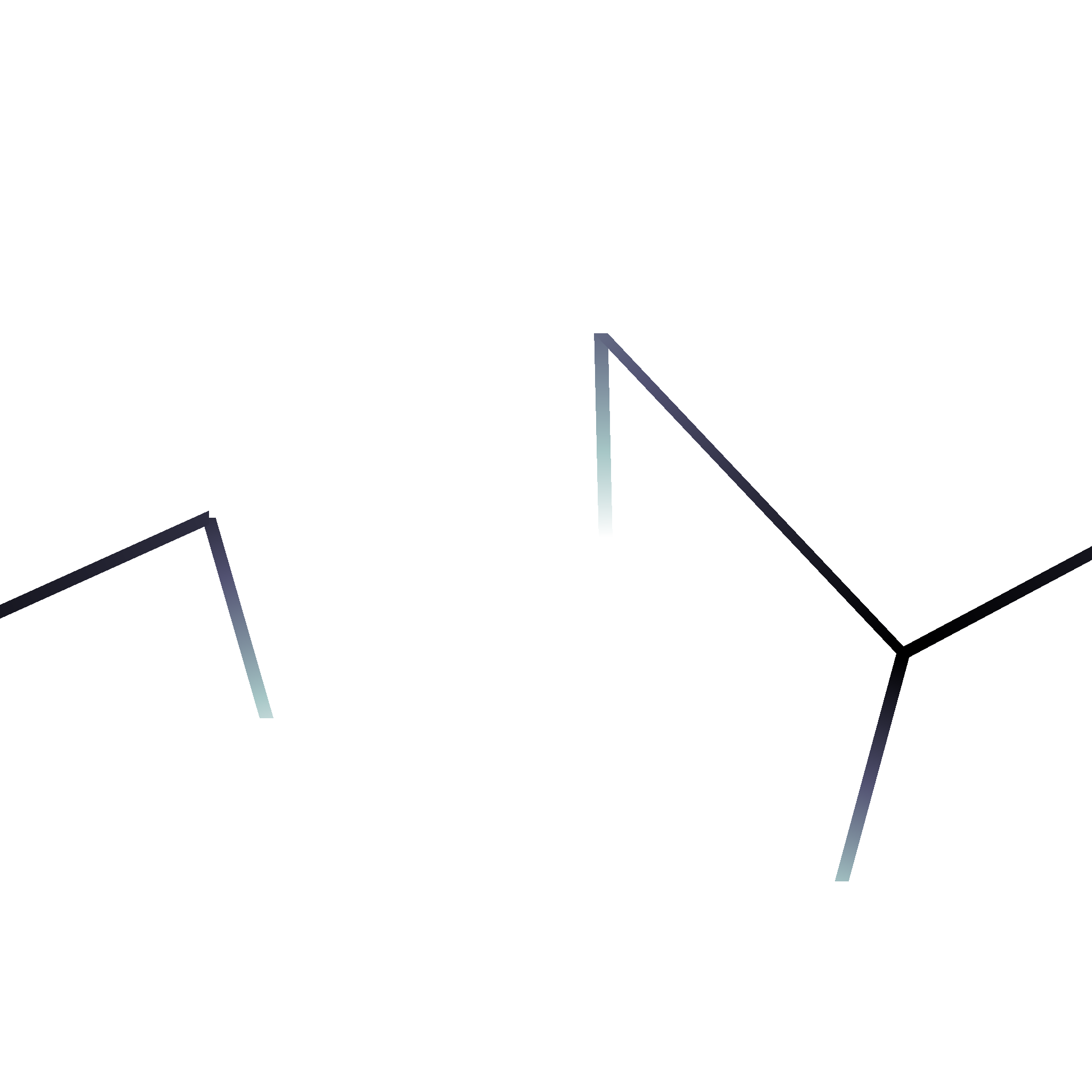}}
    \small Novel views
    \end{minipage}
    \vspace{-1mm}
    \caption{\small Test results on real images from MegaDepth.}
    \vspace{-1mm}
    \label{fig:wireframe-real}
\end{figure*}

We test our 3D wireframe reconstruction method on both the synthetic dataset and the real images.
Examples illustrating the visual quality of the final reconstruction are shown in \Cref{fig:wireframe-synthetic,fig:wireframe-real}.  A video demonstration can be found in \url{http://y2u.be/l3sUtddPJPY}.
We do not show the ground truth 3D wireframes for the real landmark dataset due to its incomplete depth maps.

\section*{Acknowledgement}

This work is partially supported by Sony US Research Center, Adobe Research, Berkeley BAIR, and Bytedance Research Lab.

\small{
\bibliographystyle{ieee_fullname}
\bibliography{paper,misc}

\begin{thebibliography}{10}\itemsep=-1pt

\bibitem{Bertamini:2013:VSP}
Marco Bertamini, Mai Helmy, and Daniel Bates.
\newblock The visual system prioritizes locations near corners of surfaces (not
  just locations near a corner).
\newblock {\em Attention, Perception, {\&} Psychophysics}, 75(8):1748--1760,
  Nov 2013.

\bibitem{Coughlan:1999:MWC}
James~M Coughlan and Alan~L Yuille.
\newblock Manhattan world: Compass direction from a single image by bayesian
  inference.
\newblock In {\em ICCV}, volume~2, pages 941--947, 1999.

\bibitem{Eigen:2014:Depth}
David Eigen, Christian Puhrsch, and Rob Fergus.
\newblock Depth map prediction from a single image using a multi-scale deep
  network.
\newblock In {\em NIPS}, 2014.

\bibitem{Fleet:2014:Lsd}
Jakob Engel, Thomas Sch{\"o}ps, and Daniel Cremers.
\newblock {LSD-SLAM}: Large-scale direct monocular slam.
\newblock In {\em ECCV}. 2014.

\bibitem{Eslami:2018:NSR}
S.~M.~Ali Eslami, Danilo Jimenez~Rezende, Frederic Besse, Fabio Viola, Ari~S.
  Morcos, Marta Garnelo, Avraham Ruderman, Andrei~A. Rusu, Ivo Danihelka, Karol
  Gregor, David~P. Reichert, Lars Buesing, Theophane Weber, Oriol Vinyals, Dan
  Rosenbaum, Neil Rabinowitz, Helen King, Chloe Hillier, Matt Botvinick, Daan
  Wierstra, Koray Kavukcuoglu, and Demis Hassabis.
\newblock Neural scene representation and rendering.
\newblock {\em Science}, 2018.

\bibitem{everingham2010pascal}
Mark Everingham, Luc Van~Gool, Christopher~KI Williams, John Winn, and Andrew
  Zisserman.
\newblock The {Pascal} visual object classes ({VOC}) challenge.
\newblock {\em International journal of computer vision}, 88(2):303--338, 2010.

\bibitem{Furukawa:2009:MWS}
Yasutaka Furukawa, Brian Curless, Steven~M Seitz, and Richard Szeliski.
\newblock {Manhattan-world stereo}.
\newblock In {\em CVPR}, 2009.

\bibitem{geiger2013vision}
Andreas Geiger, Philip Lenz, Christoph Stiller, and Raquel Urtasun.
\newblock Vision meets robotics: The {KITTI} dataset.
\newblock {\em The International Journal of Robotics Research}, 2013.

\bibitem{Groueix:2018:AN}
Thibault Groueix, Matthew Fisher, Vladimir~G. Kim, Bryan Russell, and Mathieu
  Aubry.
\newblock {AtlasNet}: A papier-m\^ach\'e approach to learning {3D} surface
  generation.
\newblock In {\em CVPR}, 2018.

\bibitem{Hofer:2017:Efficient}
Manuel Hofer, Michael Maurer, and Horst Bischof.
\newblock Efficient {3D} scene abstraction using line segments.
\newblock {\em Computer Vision and Image Understanding}, Apr. 2017.

\bibitem{Huang:2017:TCS}
Jingwei Huang, Angela Dai, Leonidas Guibas, and Matthias Niessner.
\newblock {3Dlite}: Towards commodity {3D} scanning for content creation.
\newblock {\em ACM Trans. Graph.}, 2017.

\bibitem{Huang:2018:LPW}
Kun Huang, Yifan Wang, Zihan Zhou, Tianjiao Ding, Shenghua Gao, and Yi Ma.
\newblock Learning to parse wireframes in images of man-made environments.
\newblock In {\em CVPR}, 2018.

\bibitem{izadi2011kinectfusion}
Shahram Izadi, David Kim, Otmar Hilliges, David Molyneaux, Richard Newcombe,
  Pushmeet Kohli, Jamie Shotton, Steve Hodges, Dustin Freeman, Andrew Davison,
  et~al.
\newblock {KinectFusion}: real-time {3D} reconstruction and interaction using a
  moving depth camera.
\newblock In {\em Proceedings of the 24th annual ACM symposium on User
  interface software and technology}, pages 559--568. ACM, 2011.

\bibitem{kazhdan2006poisson}
Michael Kazhdan, Matthew Bolitho, and Hugues Hoppe.
\newblock Poisson surface reconstruction.
\newblock In {\em Proceedings of the fourth Eurographics symposium on Geometry
  processing}, volume~7, 2006.

\bibitem{Kingma:2014:Adam}
Diederik~P Kingma and Jimmy Ba.
\newblock Adam: A method for stochastic optimization.
\newblock {\em arXiv preprint arXiv:1412.6980}, 2014.

\bibitem{Zhengqi:2018:MegaDepth}
Zhengqi Li and Noah Snavely.
\newblock {MegaDepth}: Learning single-view depth prediction from internet
  photos.
\newblock In {\em Computer Vision and Pattern Recognition (CVPR)}, 2018.

\bibitem{Li:2018:MDL}
Zhengqi Li and Noah Snavely.
\newblock {MegaDepth}: Learning single-view depth prediction from internet
  photos.
\newblock In {\em CVPR}, 2018.

\bibitem{Liu:2018:Planenet}
Chen Liu, Jimei Yang, Duygu Ceylan, Ersin Yumer, and Yasutaka Furukawa.
\newblock {PlaneNet}: Piece-wise planar reconstruction from a single {RGB}
  image.
\newblock In {\em CVPR}, 2018.

\bibitem{lorensen1987marching}
William~E Lorensen and Harvey~E Cline.
\newblock Marching cubes: A high resolution {3D} surface construction
  algorithm.
\newblock In {\em ACM siggraph computer graphics}, volume~21, pages 163--169.
  ACM, 1987.

\bibitem{Ma:2003:IVI}
Yi Ma, Stefano Soatto, Jana Kosecka, and S.~Shankar Sastry.
\newblock {\em An Invitation to {3D} Vision: From Images to Geometric Models}.
\newblock SpringerVerlag, 2003.

\bibitem{Mur-Artal-2015}
Ra{\'u}l Mur-Artal, JMM Montiel, and Juan~D Tard{\'o}s.
\newblock {ORB-SLAM}: A versatile and accurate monocular {SLAM} system.
\newblock {\em IEEE Transactions on Robotics}, 2015.

\bibitem{Newell:2016:Stacked}
Alejandro Newell, Kaiyu Yang, and Jia Deng.
\newblock Stacked hourglass networks for human pose estimation.
\newblock In {\em ECCV}, 2016.

\bibitem{Ramalingam:2013:lifting}
Srikumar Ramalingam and Matthew Brand.
\newblock Lifting {3D} manhattan lines from a single image.
\newblock In {\em Proceedings of the IEEE International Conference on Computer
  Vision}, pages 497--504, 2013.

\bibitem{Song:2017:Semantic}
Shuran Song, Fisher Yu, Andy Zeng, Angel~X. Chang, Manolis Savva, and Thomas
  Funkhouser.
\newblock Semantic scene completion from a single depth image.
\newblock In {\em CVPR}, 2017.

\bibitem{tardif2009non}
Jean-Philippe Tardif.
\newblock Non-iterative approach for fast and accurate vanishing point
  detection.
\newblock In {\em 2009 IEEE 12th International Conference on Computer Vision},
  pages 1250--1257. IEEE, 2009.

\bibitem{toldo2008robust}
Roberto Toldo and Andrea Fusiello.
\newblock Robust multiple structures estimation with {J}-linkage.
\newblock In {\em European conference on computer vision}, pages 537--547.
  Springer, 2008.

\bibitem{wang2017point}
Xinggang Wang, Kaibing Chen, Zilong Huang, Cong Yao, and Wenyu Liu.
\newblock Point linking network for object detection.
\newblock {\em arXiv}, 2017.

\bibitem{wu2016single}
Jiajun Wu, Tianfan Xue, Joseph~J Lim, Yuandong Tian, Joshua~B Tenenbaum,
  Antonio Torralba, and William~T Freeman.
\newblock Single image {3D} interpreter network.
\newblock In {\em European Conference on Computer Vision}, pages 365--382.
  Springer, 2016.

\bibitem{Yang:2018:ECCV}
Fengting Yang and Zihan Zhou.
\newblock Recovering {3D} planes from a single image via convolutional neural
  networks.
\newblock In {\em ECCV}, 2018.

\bibitem{zhang2016joint}
Kaipeng Zhang, Zhanpeng Zhang, Zhifeng Li, and Yu Qiao.
\newblock Joint face detection and alignment using multitask cascaded
  convolutional networks.
\newblock {\em IEEE Signal Processing Letters}, 23(10):1499--1503, 2016.

\bibitem{Zhang:2017:PRI}
Yinda Zhang, Shuran Song, Ersin Yumer, Manolis Savva, Joon-Young Lee, Hailin
  Jin, and Thomas Funkhouser.
\newblock Physically-based rendering for indoor scene understanding using
  convolutional neural networks.
\newblock In {\em CVPR}, 2017.

\bibitem{zhang2012tilt}
Zhengdong Zhang, Arvind Ganesh, Xiao Liang, and Yi Ma.
\newblock Tilt: Transform invariant low-rank textures.
\newblock {\em International journal of computer vision}, 99(1):1--24, 2012.

\bibitem{Zingl:2012:Rasterizing}
Alois Zingl.
\newblock A rasterizing algorithm for drawing curves, 2012.

\bibitem{Zou:2018:LNR}
Chuhang Zou, Alex Colburn, Qi Shan, and Derek Hoiem.
\newblock {LayoutNet}: Reconstructing the {3D} room layout from a single {RGB}
  image.
\newblock In {\em CVPR}, 2018.

\end{thebibliography}
}

\nothing{
\appendix
\normalsize
\section{Supplementary Materials}

\subsection{Pseudocode for Line Vectorization}

\Cref{alg:edge-detection} gives a more detailed description for line vectorization.  The algorithm takes the C-junction set $\mathsf{V}_C$ and
T-junction set $\mathsf{V}_T$ as the input and outputs a vectorized wireframe $(\vertexSet, \edgeSet)$. In the first stage (Lines \ref{line:vec-c-start}-\ref{line:vec-c-end}), we
find the lines among C-junctions according to the line confidence function.  The procedure
\textsc{Prune-Lines} greedily removes the lines with the lowest confidence that either intersect with other lines (\Cref{line:intersect}) or are too close to
other lines in term of the polar angle (\Cref{line:angle}).    In the second stage (Lines \ref{line:vec-t-start}-\ref{line:vec-t-end}),
we add the T-junctions into the wireframe.  From Lines \ref{line:vec-t-detect-start}-\ref{line:vec-t-detect-end}, we find the T-junctions that are on the existing wireframe.
We first adjust the positions of those T-junctions by projecting them onto the line (\Cref{line:project}) and then add them to the candidate T-junction set $\vertexSet'$ (\Cref{line:add}). 
Because the degree of a T-junction is always one, we try to find the connection with the highest confidence for those candidates T-junctions (Lines \ref{line:vec-t-find-best}-\ref{line:vec-t-end}).  We repeat the this process until $\vertexSet$, $\vertexSet'$, and $\edgeSet$ remain the same in the last iteration.

\begin{algorithm}[htbp]
  \caption{Edge Vectorization Algorithm} \label{alg:edge-detection}
  \begin{algorithmic}[1]
    \Require Candidate C-junction set $\mathsf{V}_C$, T-junction set $\mathsf{V}_T$.
    \Require Hyper-parameters $\eta_c$ and $\eta_\circ$.
    \Ensure Wireframe $(\vertexSet, \edgeSet)$.
    \Procedure{Vectorize}{$\mathsf{V}_C$, $\mathsf{V}_T$}
    
      \State $\vertexSet \gets \mathsf{V}_C$ \label{line:vec-c-start}
      \State $\edgeSet \gets \Call{Prune-Lines}{\{(\vertexA, \vertexB) | \vertexA, \vertexB \in \vertexSet, c(\vertexA, \vertexB) > \eta_c\}}$    \label{line:vec-c-end}
      
      \State $\vertexSet' \gets \varnothing$ \label{line:vec-t-start}
      \While{$\vertexSet$, $\vertexSet'$, or $\edgeSet$ change in the last iteration}
        \ForAll{$\vertex \in \mathsf{V}_T$} \label{line:vec-t-detect-start}
          \ForAll{$\edge = (\vertexA, \vertexB) \in \edgeSet$}
            \If{$\vertex$ is near $\edge$}
              \State project $\vertex$ to the line $\edge$ \label{line:project}
              \State $\vertexSet' \gets \vertexSet' \cup \{\vertex\}$  \label{line:add}
              \State \textbf{break}
            \EndIf
          \EndFor
        \EndFor \label{line:vec-t-detect-end}
        
        \ForAll{$\vertexA \in \vertexSet'$} \label{line:vec-t-find-best}
          \State $\vertexB \gets \argmax_{\vertexB \in \vertexSet \cup \vertexSet'} c(\vertexA, \vertexB)$
          \If{$c(\vertexA, \vertexB) \ge \eta_c$}
            \State $\vertexSet' \gets \vertexSet' \backslash \{\vertexA\}$
            \State $\vertexSet \gets \vertexSet \cup \{\vertexA\}$
            \State $\edgeSet \gets \edgeSet \cup \{(\vertexA, \vertexB)\}$
          \EndIf
        \EndFor
        \State $\mathsf{V}_T \gets \mathsf{V}_T \backslash (\vertexSet \cup \vertexSet')$
      \EndWhile
      \State $\edgeSet \gets \Call{Prune-Lines}{\edgeSet}$  \label{line:vec-t-end}
      \State \Return $(\vertexSet, \edgeSet)$
    \EndProcedure

    \Procedure{Prune-Lines}{$\edgeSet$}
      \State sort $\edgeSet$ w.r.t confidence values in descending order
      \State $\edgeSet' \gets \varnothing$
      \ForAll{$\edge \in \edgeSet$}
        \If{$\exists \edge' \in \edgeSet':$ $\edge$ intersects with $\edge'$} \label{line:intersect}
          \State \textbf{continue}
        \EndIf
        \If{$\exists \edge' \in \edgeSet':$ $\edge' \cap \edge \ne \varnothing$ \textbf{and}  $\angle(\edge, \edge') < \eta_\circ$} \label{line:angle}
          \State \textbf{continue}
        \EndIf
        \State $\edgeSet' \gets \edgeSet' \cup \{\edge\}$
      \EndFor
      \State \Return $\edgeSet'$
    \EndProcedure
    
  \end{algorithmic}
\end{algorithm}

\subsection{Line Assignments for Vanishing Points}
In \Cref{eq:lift-objective}, we need to find the set of lines $\assignment_i \subseteq \edgeSet$ corresponding to the vanishing point $i$.  Mathematically, we define the objective function
\begin{equation*} 
    \min_{\assignment} \sum_{i}^3 \sum_{(\vertexA, \vertexB) \in \assignment_i} 
    \left\| (\vertexA - \vanishPoint_i) \times (\vertexA - \vertexB) \right\|_2,
    \label{eq:vp}
\end{equation*}
where $\|(\cdot)\times(\cdot)\|_2$ can be understood as the parallelogram area formed by two vectors.  Since each line in this equation is mutually independent, we can solve this optimization problem by greedily assigning each line to the best vanishing point $i$ to minimize the objective function.

\subsection{Sampled Failure Cases and Discussions}   \label{sec:failure}
\begin{figure}[htbp]
    \begin{center}
        \begin{minipage}[t]{0.32\linewidth}
            \includegraphics[width=\linewidth]{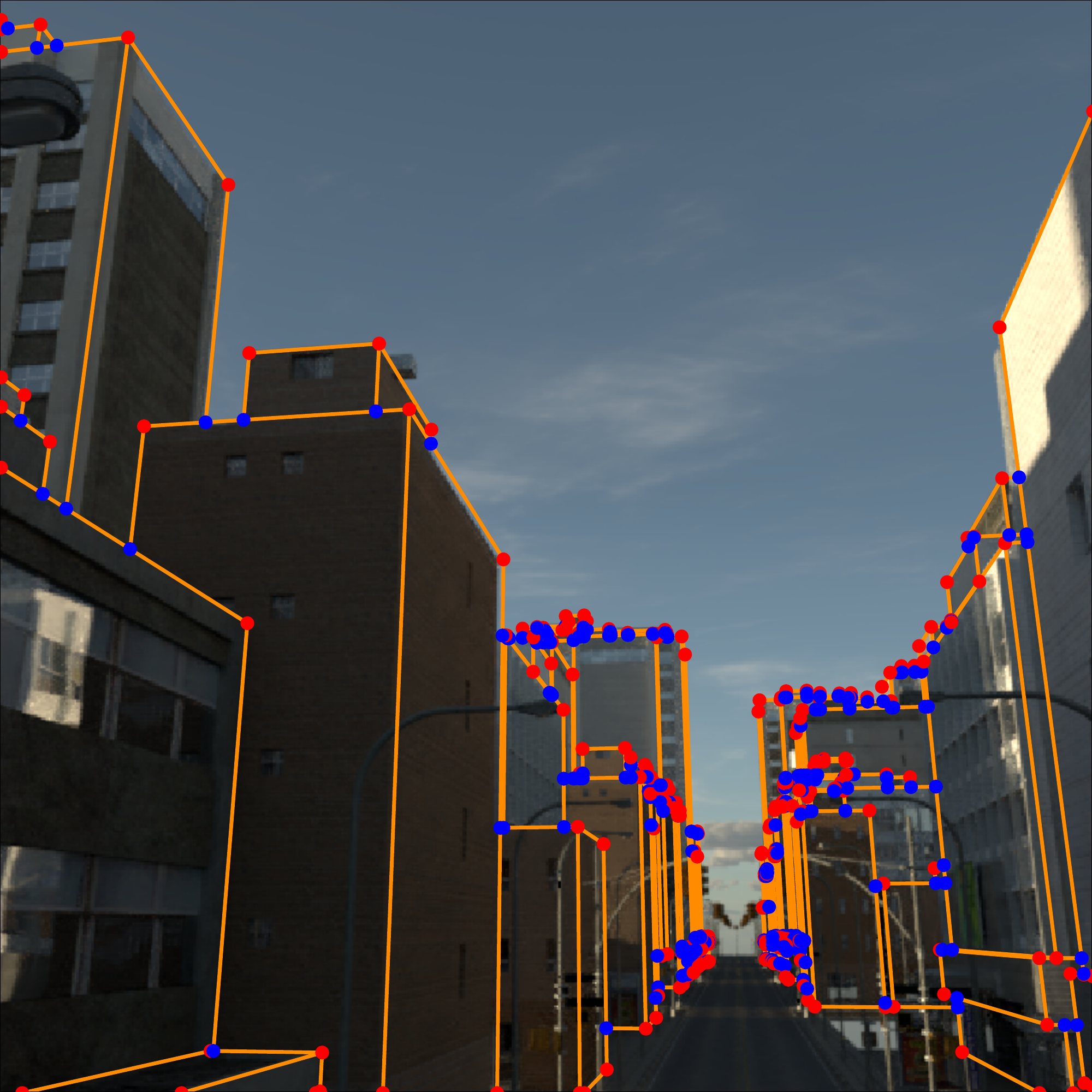} 
            
            \vspace{1mm}
            
            \includegraphics[width=\linewidth]{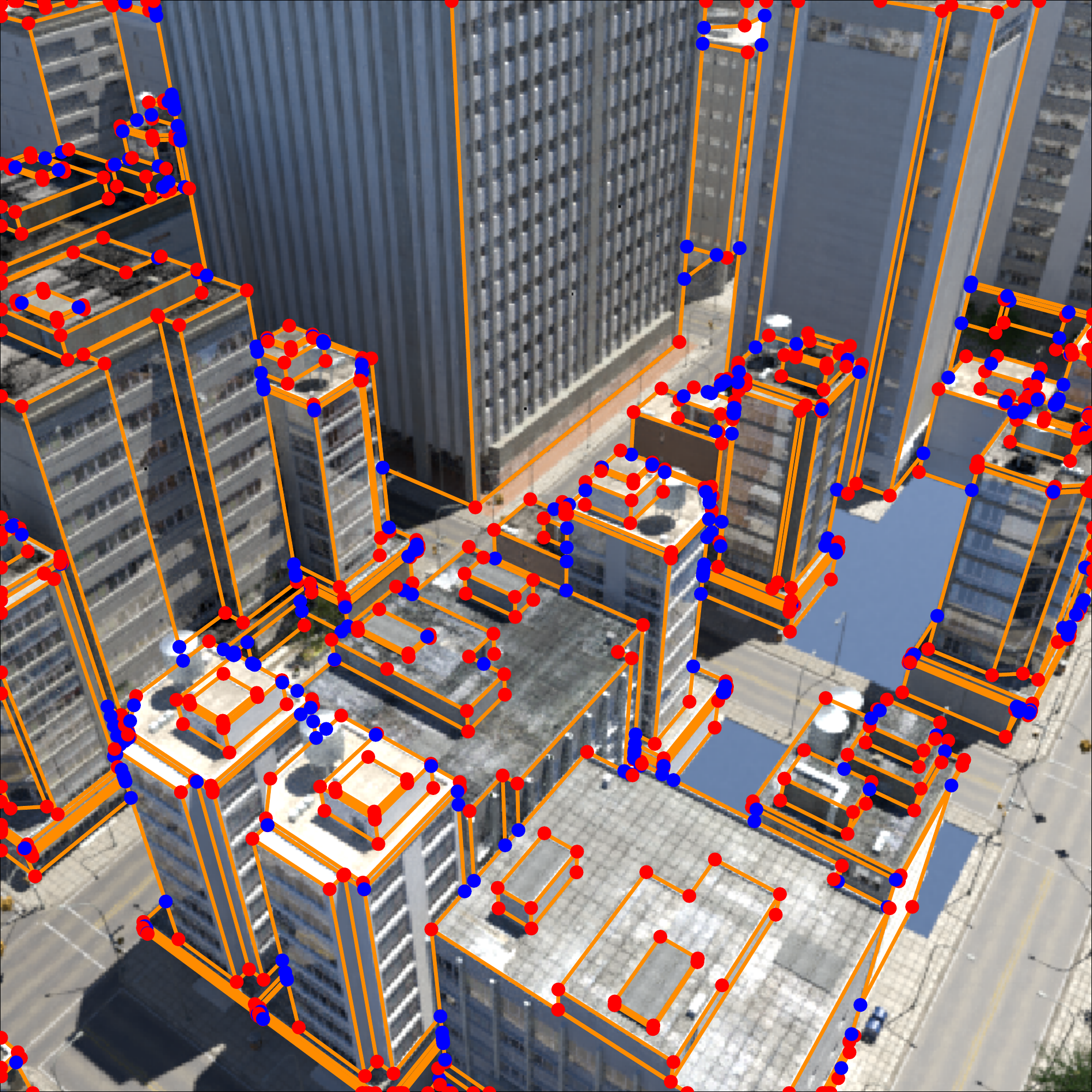} 
            \centering
            \small Ground truth
        \end{minipage}\hfill
        \begin{minipage}[t]{0.32\linewidth}
            \frame{\includegraphics[width=\linewidth]{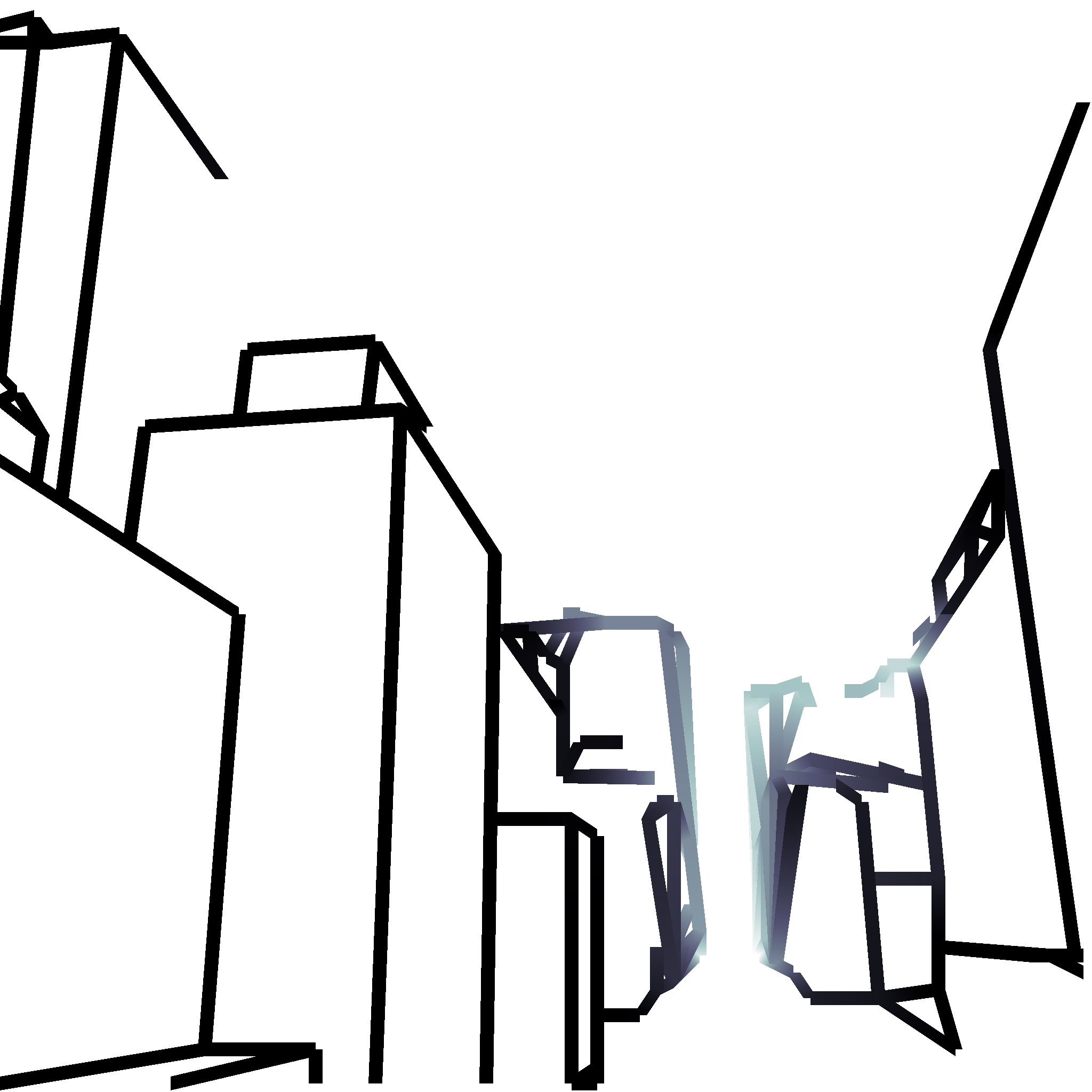}}
            
            \vspace{1mm}
            
            \frame{\includegraphics[width=\linewidth]{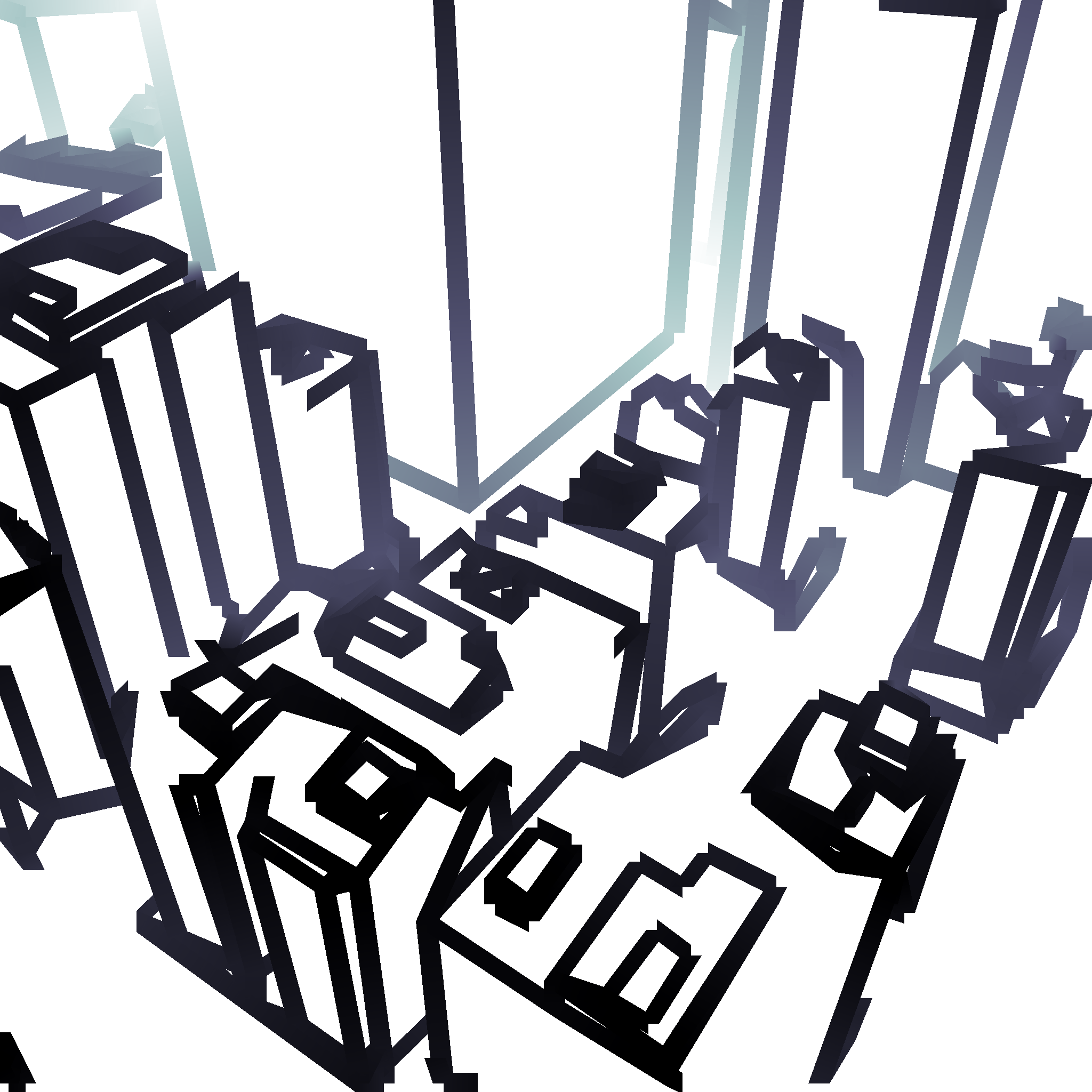}}
            \centering
            \small Inferred 3D
        \end{minipage}\hfill
        \begin{minipage}[t]{0.32\linewidth}
            \frame{\includegraphics[width=\linewidth]{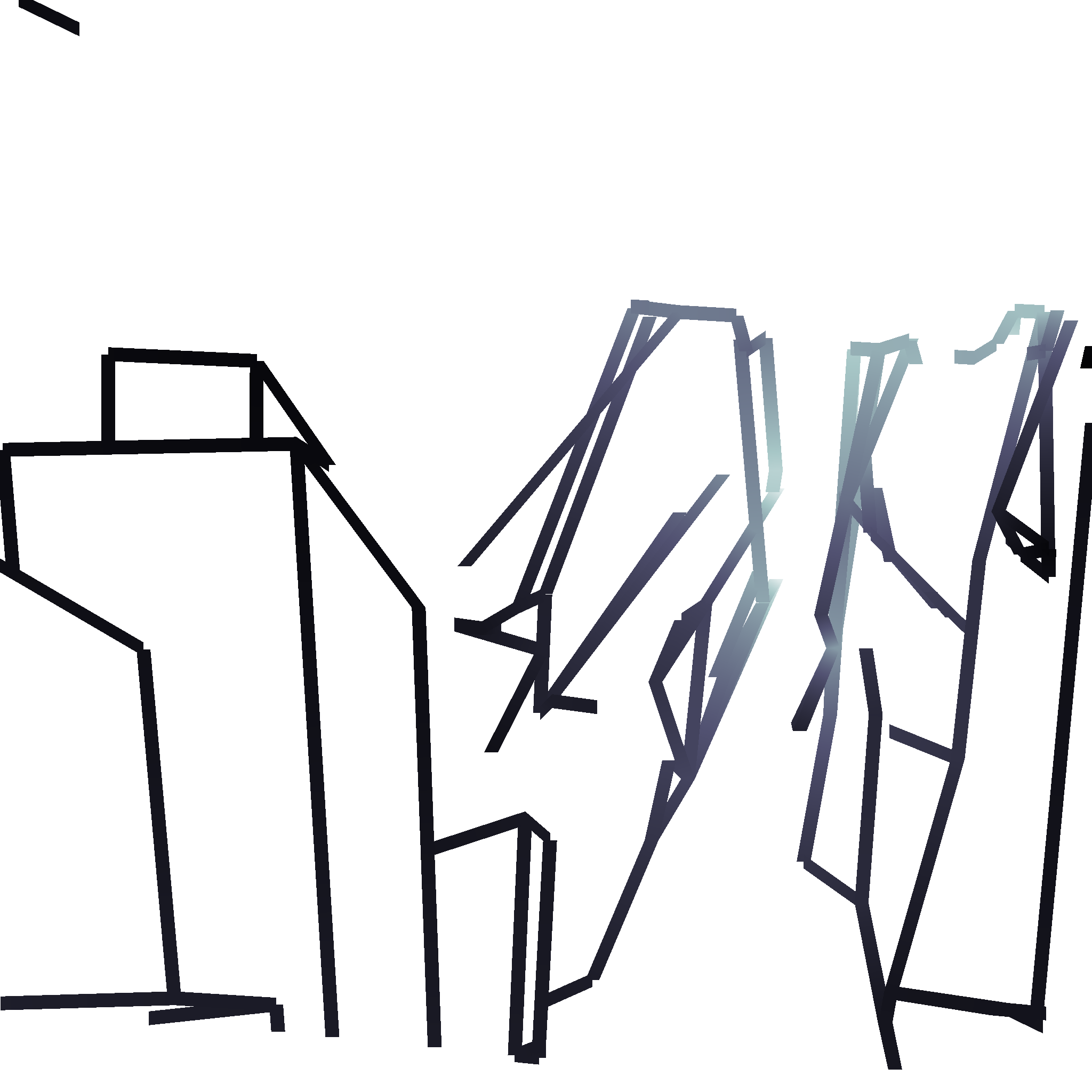}}
            
            \vspace{1mm}
            
            \frame{\includegraphics[width=\linewidth]{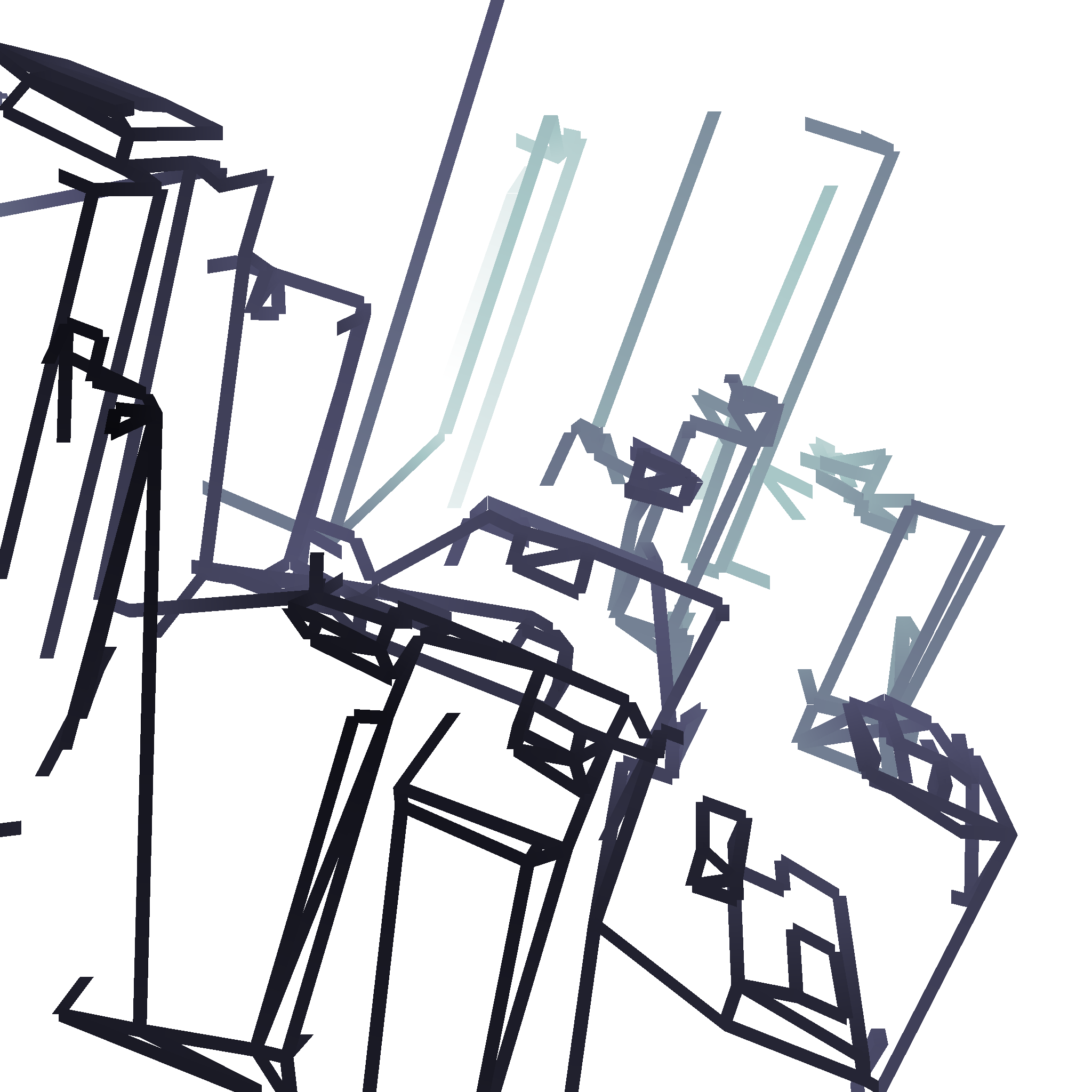}}
            \centering
            \small Novel views
        \end{minipage}
    \end{center}
    \caption{Failure cases on the SceneCity dataset.} \label{fig:failure}
\end{figure}

\Cref{fig:failure} demonstrates some failure cases in our SceneCity dataset.  We found that our pipeline might not work well on the scenes in which there are many lines and junctions that are close to each other.  This is because the resolution of the output heat map is $128 \times 128$, so any detail whose size is below two or three pixels might get lost during the vectorization stage.  Therefore, one of our future work is to explore the possibility of using high-resolution input and output images.  There are also issues in the 3D depth refinement stage.  When the scene is complex, finding the assignment $\assignment_i$ for each line can be hard, due to the error in the junction position and line direction.   In addition, the term contributed by erroneous lines in \Cref{eq:lift-objective} can make the depth of some junctions inaccurate.  Such problem might potentially be alleviated by increasing the resolution of the input and output images, using a more data-driven method, designing a better objective function, or employing a RANSIC approach in those two stages.

\subsection{Network Comparison}

In this section, we discuss the difference between our network and the one in \cite{Huang:2018:LPW}. Our network is similar as their line detection network with the difference in the following perspectives:
\begin{enumerate}
    \item In each hourglass module, they use two consecutive residual modules (RM) at each spatial resolution while we only use one RM, resulting less parameters in each hourglass module. Note that our design is the same as the original hourglass paper \cite{Newell:2016:Stacked}, which enables us to use more RM in each hourglass module and reduce the computational complexity since more computation is allocated in lower resolution stages. We adopt such design since we find using two RM gives negligible gains to the performance compared with only one.
    \item We apply the intermediate supervision to the stacked hourglass network.  For each hourglass modules, the loss term associated with the predicted heat maps is added to the final loss.   \cite{Huang:2018:LPW} does not such intermediate supervision in their method.  We find such intermediate supervision vital in both our synthetic dataset as well as their 2D dataset in terms of both accuracy and robustness.
    \item We observe that using 2 stacked hourglass modules gives a similar performance as the 5 stacked hourglass modules in \cite{Huang:2018:LPW}. On the other hand, using 2 stacks consumes less memory, which gives us more design flexibility. For example, we are able to utilize a larger batch size to make the gradients more stable.
\end{enumerate}

}%

\ifthenelse{\equal{\final}{0}}
{
\clearpage
\pagenumbering{roman}
\input{todo}
\input{narrative}
\section{Blog}

\begin{description}
\item[November 23, 2018]
{
\yichao{
Hi all,

One chair mail to me that our paper has a lot of domain conflicts
\begin{quote}
Please make sure that each of the authors on your paper has only listed institutions where they have worked in the last 12 months as their domain conflicts.  At last check, your paper had conflicts with nearly all AC panels, but it is possible that this is correct due to the long author list.
\end{quote}

This will likely result in an assignment to AC based primarily on conflicts, rather than expertise.
Please check and reply with your domain conflicts setting in the CMT3 system (https://cmt3.research.microsoft.com/CVPR2019/User/ConflictDomains).  Mine is "berkeley.edu;adobe.com".
}%
}
\item[November 22, 2018: TTS]
{
\yichao{
For TTS, I suggest that we can use https://cloud.google.com/text-to-speech/docs/basics.  I have tested it and it offers much better sound quality.
}%
}
\item[September 11, 2018: quick meeting note]
{
\liyi{
Focus on having a working pipeline for synthetic datasets.

Annotating real datasets will take longer and might belong to another project.

Yi Ma is optimistic about having only synthetic datasets.
We can run the trained networks on real photographs, and visualize how the results look like, e.g. some rendering or popup effects \cite{Hoiem:2005:APP}.

The 3D lifting part is orthogonal to the 2D wireframe vertex and edge detection part, and can run on single or multi-view inputs.
}%
}
\item[July 13, 2018: quick meeting summary]
{
\liyi{

\paragraph{Scope}
What happens if we apply scenes with curved objects into the current network?
Seeing the failure cases can provide inspirations on the next steps.

Yichao believes that the scope should be objects with clear contours, while Zhili wonders if we can also handle objects not entirely defined by curve networks such as cylinders and balls.
The 2D network should still be able to detect silhouettes, so it is a question on how to properly do the 3D lifting step.

\paragraph{Applications}

We need to figure out something atop the basic algorithm for patent.
}%
}
\item[May 25, 2018: meeting note]
{
\liyi{
Hi Yichao and Qi:

I encourage you to summarize the key points of the meeting, so that we will not forget what has been verbally discussed and we can follow up with all main points.
Due to the nature of human memory, the earlier we write down the notes, the less we will forget.

\paragraph{Scope}

We construct wireframe models from a photograph, video, with or without depth information.
We can start with the simplest case of a single color photograph.
We should be able to handle straight lines and curves for general CAD models, as straight lines only can be too limited (only box-like shapes).

\paragraph{Method}

I would like to hear more from Yichao and Yi decide on this part as they know more on the vision side.
Possibilities include machine learning, and traditional computer vision and/or geometry methods (Niloy Mitra, Alla Sheffer).
Personally I would recommend try the latter first, and if that does not work well (based on prior literature and our own experiences), switch to the former.
Black-box machine learning can be hard to debug, and even if it works we should be able to explain why it works, which I believe is a key missing from many machine learning papers.

\paragraph{Literature}

Qi and I will help with literature survey (via \Cref{sec:prior}), including both academic papers and industry products.
This exercise is important to understand competitions (e.g. we do not want to compete with Apple or Google on AR infrastructure) and define the goal and scope of this project.

\paragraph{Data}
Ideally, we want clean CAD datasets already annotated with semantic contours.
There are many geometry scenes and good renderers we can use, and embed such CAD models.
Our method should be renderer agnostic, e.g. works with both realistic and NPR styles.
We can start with simple, sketch/line drawings to test and train our methods.
}%

\yichao{
Hi all,

Sorry for the miscommunication so that Yi was not able to attend today's weekly sync meeting. I confirmed with Yi and we will still meet at 11:00am Friday next week.

To summarize this meeting, in following week I will (sorted by priority)
\begin{enumerate}
\item	
build up a toy 3D scene and rendering pipeline so it is able to generate images and proper annotations for edges and junctions;
\item
implement the paper "Lifting 3D Manhattan Lines from a Single Image" so that we can use the ground truth data from 1. to infer its 3d structure;
\item
play with the data photo realistic dataset Amazon Lumberyard Bistro \url{https://developer.nvidia.com/orca/amazon-lumberyard-bistro}.
\end{enumerate}

Meanwhile, Li-Yi Wei and Qi Sun will do some literature survey from the perspective of computer graphics to provide some insight for methods and check the novelty.

Best
}%
}
\item[May 22, 2018]
{
\liyi{
Hi guys:

I saw Oliver Wang's offsite poster today and found out he is also working on 3D reconstruction of proxies for image and video editing with 2 interns.
He also mentioned \cite{Groueix:2018:AN} to me.
We should sync up together after the retreat.
}%

\liyi{
I talked with Jimei Yang today (his poster is next right to mine), and he told me that he has been working on related (but not the same) 3D reconstruction stuff.
Specifically, he has a project reconstructing bounding boxes of indoor objects for semantic understanding and motion planning applications.
This is another sync up opportunity.
}%
}
\item[May 19, 2018]
{
[continuation from the previous post]

\liyi{
Hi Yichao:

Good progress!

Before diving into technical implementation details, allow me to ask the reason for reproducing \cite{Huang:2018:LPW}.
For example, do you intend to extend their method for reconstructing 3D geometry with regular structures?
It can help to describe your plan (e.g. via \Cref{sec:introduction,sec:method}) first before doing anything else.

\paragraph{Latex}

Are you using the built-in text editor for MikTex or something, as I noticed strangely wrapped lines.
(See my updates and comments under the abstract.)

\paragraph{Git repo}

Strangely, I did not receive any email notifications from the bitbucket repo.
I will watch this repo more closely from now on.
}%
}
\item[May 18, 2018]
  \yichao{
    The training is finished.  I will start to evaluate the performance of wireframe 2D.

    OK. The result is a little bit disappointing.  Because of the method they used, wireframe can
    either be missing, several pixels off, or false positives.  I am not sure whether those result are
    usable now.   I am making a document on this.

    The method they use is to regress the junction point and the heat map separately. The joints
    might be off or not detected, and it may not match with the heat map.  I am thinking about a
    method that can combine the edge detection and line transform.  The paper Holistically Nested
    Edge Detection seems to be helpful.
  }
\item[May 17, 2018]
  \yichao{
    Today I tried to reproduce the experiment from CVPR 2018's paper ``Learning to Parse Wireframes
    in Images of Man-Made Environments''.  Training the junction network is blasting fast, which only
    takes about 1 hour.  Training the line network takes more than 24 hours.  I don't think it will
    finish today.  We need to wait until tomorrow for evaluating the result.

    Today I tried to review the code of that CVPR's paper.  I will say that the code quality is
    so so.  It uses 4 different library for progress bar, and there are some strange code that I
    may never know their usage.  Nevertheless, with some minor modification and create some
    directories, the code runs.  I might need spend more time on reading the code for junction
    detection tomorrow.

    I also spend some time on reading the dataset.  It does not provide a utility script to
    visualize its training data.  I wrote my own in jupyter.  First, the data are stored in
    python's pickle format so I turn them into json format for human readibility.  The origin
    pickle file contains a lot of entries.  I still do not understand some entires such as
    \texttt{pointlines} due to the lack of document.  But I think they are just some redundant
    format.  The only important labeling is the position of junctions and the line between them.

    The dataset quality is not as good as I think, probably due to the difficulty of labeling.  I
    notice the following problems:
    \begin{enumerate}
      \item The positions of junctions are not pixel-accurate;  My estimation of average error can
        be as large as 2-5 pixels or so;
      \item The types of junctions are not labeled, e.g., T, L, etc;
      \item The definition of a joint is not clear.  Sometimes labeler will mark some random points
        in non-regular objects such as sofa as joints.  I don't think prediction of joints will be
        reliable for non-regular objects.
    \end{enumerate}
  }%
\item[May 16, 2018]
  \yichao{
    Hi Everyone,

    Last two days I am dealing with the things such as setting up PCs and finishing various
    training.  I will forcus on my project from today.  Thanks liyi for sending such a nice
    template. One of my question is what is the usage of Meta directory?  What is non-project
    specific stuff?  I will start to update the introduction section based on my understanding of
    the project and try to reproduce the result from literature.

    I just switch the video card and display driver from Nvidia to Intel so there is no annoy
    display lag during training.

    Now I start to train the 2D version of wireframe from literature:
    \url{https://github.com/zhou13/wireframe}.  I probably should move the code into this
    repository.  If everything go well, I can test its performance tomorrow.

    Best.
  }%
\item[May 15, 2018]
{
\liyi{
Hi Yichao:

You can use this paper draft as a place to exercise our thinking.
One possibility is to start updating \Cref{sec:introduction} based on your understanding and planning of the project.
Other collaborators, such as Qi, Zhili, Yi, and I, can also participate in the paper draft.
You can compile the pdf files (look at {Makefile}) which also contain explanations of why I advocate the use of paper drafts to manage research projects.

Concurrently, we can also have weekly meetings to sync up.
I have sent out a meeting invitation to everyone.

Feel free to let me know if you have any questions.
You can use something like \yichao{I am awesome!}.

I am traveling for I3D this week, and Zhili and I will be in Adobe research offsite next week, so you will not see me in person until Friday May 25 2018.
But I am reachable asynchronous (I have written most of my SIGGRAPH papers without any meetings), and Qi can visit the office after he relocates.

Best.
}
}
\item[May 9, 2018]
{
\liyi{
There are capture systems that are very easy to use such as \cite{Li:2015:PS,Zhou:2018:FU} for faces.
We aim to do something analogous for shapes, starting from interiors.
}%
}
\item[May 8, 2018: Yichao's project]
{
\liyi{
Hi Qi, I can see connections of their global reconstruction to AR redirected walking + touching (your adobe intern project) and even your first siggraph submission on reconstructing repetitive geometry.
If you like, we can let them talk first and you show your video demo of the AR project. We can also show the slides of ar sculpting if time allows.

\paragraph{After meeting with Yi Ma and Yichao}

It is quite clear that Yi Ma has strong motivation to pursue database-driven reconstruction from casual inputs.
Thus, we can keep this as a constraint, and think more on the UI and application sides.

The overall direction is still quite vague with many free variables we can design:
\begin{description}
\item[Input]
Anything that does not require a lot of user workloads, such as taking a few photos from a cell phone, sketching a few contours, etc.

\item[Method]
I mentioned 3 possibilities: traditional geometry fitting, machine learning, and inverse procedural modeling.
The Berkeley guys did not specify any preference as far as I can see.

\item[Applications]
Use photographs to bootstrap a good initial model, from which users can edit and customize.
Similar to pix2pix, once the system/network works for real photographs, we can try manual sketches to see if the method can scale to novel designs.

\item[Scope]
Yi Ma suggested that we start from indoor environments so that we can restrict the scope to furniture, walls, and boxes.
\end{description}

Yichao did not say anything during the meeting, so it is not clear to me what he is really thinking.
But I guess he will follow what Yi Ma says.
It is possible to align their global reconstruction with our VR/AR content creation, for potential long-term collaboration.

It might help for you and Yichao to chat in person (just the two of you) so that you can have some sense of his true interests and passions.
I can also talk to him when he is officially onboard, but then we will have lost a few days for ideation.

\paragraph{Patent}

BTW, I wonder if you want to let them know about the Adobe patent policy.
The legals usually do not file patents jointly with schools (and thus the school advisers).
So if Yichao would like to have a patent, the policy requires us not to discuss with Yi Ma during the internship, even though we can loop him back for publications after the end of the internship.
I am open either way, so I will let you and they decide.
}%
}

\end{description}

\input{cemetery}
}
{}

\end{document}